\documentclass{preprint}

\title{Maximizing Efficiency of Dataset Compression for Machine Learning Potentials With Information Theory}

\author
{Benjamin Yu,$^{1}$
Vincenzo Lordi,$^{2}$
Daniel Schwalbe-Koda$^{1}$\footnote{E-mail: dskoda@ucla.edu}\\
\vspace{1em} 
\normalfont{\small $^{1}$Department of Materials Science and Engineering, University of California, Los Angeles, CA, United States\\
$^{2}$Materials Science Division, Lawrence Livermore National Laboratory, CA, United States
}
}

\usepackage{float}
\usepackage{hyperref}
\usepackage[nameinlink,noabbrev]{cleveref}
\usepackage{chngcntr}
\usepackage{doi}
\usepackage{url}

\usepackage{caption}
\usepackage{algorithm}
\usepackage{algorithmic}
\usepackage{longtable}
\floatname{algorithm}{Algorithm}                             
\renewcommand{\thealgorithm}{\arabic{algorithm}}

\usepackage{dsk}
\DeclareMathOperator*{\argmax}{arg\,max}

\usepackage[resetlabels]{multibib}
\newcites{Supp}{Supplementary References}
\newcites{Main}{References}

\makeatletter
\renewcommand{\thealgorithm}{\arabic{algorithm}}           
\renewcommand{\fnum@algorithm}{\textbf{Algorithm~\thealgorithm}} 
\makeatother
\begin{document}

\maketitle
\thispagestyle{firstpagestyle}

\begin{abstract}
Machine learning interatomic potentials (MLIPs) balance high accuracy and lower costs compared to density functional theory calculations, but their performance often depends on the size and diversity of training datasets. Large datasets improve model accuracy and generalization but are computationally expensive to produce and train on, while smaller datasets risk discarding rare but important atomic environments and compromising MLIP accuracy/reliability. Here, we develop an information-theoretical framework to quantify the efficiency of dataset compression methods and propose an algorithm that maximizes this efficiency. By framing atomistic dataset compression as an instance of the minimum set cover (MSC) problem over atom-centered environments, our method identifies the smallest subset of structures that contains as much information as possible from the original dataset while pruning redundant information. The approach is extensively demonstrated on the GAP-20 and TM23 datasets, and validated on 64 varied datasets from the ColabFit repository. Across all cases, MSC consistently retains outliers, preserves dataset diversity, and reproduces the long-tail distributions of forces even at high compression rates, outperforming other subsampling methods. Furthermore, MLIPs trained on MSC-compressed datasets exhibit reduced error for out-of-distribution data even in low-data regimes. We explain these results using an outlier analysis and show that such quantitative conclusions could not be achieved with conventional dimensionality reduction methods. The algorithm is implemented in the open-source QUESTS package and can be used for several tasks in atomistic modeling, from data subsampling, outlier detection, and training improved MLIPs at a lower cost.
\end{abstract}

\section{Introduction}

Machine learning (ML) interatomic potentials (IPs) show excellent ability to approximate potential energy surfaces (PES) with lower cost compared to density functional theory (DFT) calculations.\cite{Behler2007,Bartok2010,thompson2015spectral,Chmiela2017,zhang2018deep,Mueller2020Machine,Manzhos2020a,Unke2021a,Batzner2022,Batatia2022,Park2024seven}
Particularly with neural network (NN) models, MLIPs reach impressively high accuracy across spaces of structures and compositions, but are often governed by power laws that relate dataset sizes to model performance.\cite{Batzner2022,merchant2023scaling}
In general, larger datasets tend to substantially improve the accuracy of models, though at the expense of higher costs for generating larger datasets and training models on them.
Multiple works have proposed ``foundation-like'' approaches to MLIPs, where NNs are trained on very large datasets of energy minimization trajectories, well-sampled PESes, and others.\cite{chen2022universal,merchant2023scaling,batatia2023foundation,deng2023chgnet,yang2024mattersim}
Nevertheless, training models on increasingly diverse and well-sampled PESes requires ensuring that atomistic datasets are not becoming redundant, and are instead well-representative of increasingly larger phase spaces.\cite{Karabin2020EntropyMaximization,Smith2021,schwalbekoda2021differentiable,MontesdeOcaZapiain2022,allen2022optimal,kulichenko2023uncertainty,barroso2024open,kaplan2025foundational,schwalbekoda2025information,tan2025enhanced}
Thus, there is a need for tools that analyze and subsample datasets to remove redundancies, maximizing the diversity of the data while also minimizing dataset sizes, which reduces training cost and complexity.\cite{Shapeev2016,imbalzano2018automatic,kulichenko2024data,sun2025meagraph}

In MLIPs, dataset subsampling methods often rely on partitioning schemes that sample structures (i.e., collections of atoms in space, either with or without boundary conditions) that are mostly distinct from each other based on a given representation.\cite{Bartok2017,rowe2020accurate,qi2024robust}
Often, to encode structures in a size-invariant manner, representations are created by taking the average of either learned or fixed per-atom representations across environments in a molecule or unit cell.
However, this averaging scheme often leads to information losses or introduction of redundancies in a dataset.
For instance, average representations are unable to distinguish between a supercell and a primitive cell of a ground state structure, which carry the same amount of information, yet contain different number of atoms.
A dataset built using the supercell has a larger memory footprint at training even though the additional atoms do not contribute additional information to the dataset and indeed can overweight those redundant atomic environments.
On the other hand, information losses also occur when interesting outlier behavior --- e.g., rare events, metastability, and more --- are underrepresented in subsampled datasets due to their sparse nature.
In many subsampling methods, there is no guarantee that rare events will be reliably captured in the downsampled datasets despite their importance in atomistic modeling.
The redundancy-outlier tradeoff is exacerbated when outlier environments in large cells are overlooked due to the use of average, per-structure representations.

In this work, we use information theory and combinatorics to minimize information losses when subsampling atomistic datasets.
Specifically, we created an algorithm that compresses datasets by solving the minimum set cover (MSC) problem over the space of atomic environments.
The MSC analysis attempts to find a minimum number of structures that spans the entire distribution of environments in the original dataset, thus maximizing the retained information on a per-dataset basis.
Our algorithm is demonstrated to preserve outliers, the long tail of force distributions, and the diversity of the datasets, outperforming baseline dataset compression methods, including farthest point sampling.
Our approach is exemplified with case studies on the GAP-20\cite{rowe2020accurate} and TM23\cite{owen2024complexity} datasets, where outlier analysis and model training demonstrate the efficiency of our compression algorithm compared to the baseline approaches.
We also benchmark the algorithm  on 64 different datasets obtained from the ColabFit repository\cite{vita2023colabfit} and show our approach outperforms the baseline algorithms in the vast majority of cases.
This work provides an information-theoretical strategy to automate the compression of atomistic datasets while minimizing loss of outliers and information.
The algorithms and implementation are publicly available in the QUESTS package at \url{https://github.com/dskoda/quests}.

\section{Methods}
\customlabel{sec:methods}{Methods}

\subsection{Information entropy and QUESTS method} \label{EntropyMethod}

Central to the idea of measuring efficiency of data compression is the theory of information from Shannon.\cite{shannon1948mathematical}
Here, we employ an information theoretical perspective to atomistic simulations\cite{schwalbekoda2025information} to propose algorithms that compress atomistic datasets, as well as figures of merit to evaluate their performance.
Below, we describe the representation and relevant information-theoretical quantities used in the manuscript.

\noindent\textbf{Representation:} The representation of atomic environments was computed as described in our previous work.\cite{schwalbekoda2025information}
In short, an atom-centered descriptor is obtained by sorting the distances from a central atom $i$ to its $k$ nearest neighbors (considering periodic boundary conditions), forming a vector $\X_i^{(1)}$ of length $k$ as

\begin{equation}\label{eq:x1-short}
\X_i^{(1)} =
\begin{bmatrix}
\frac{w(r_{i1})}{r_{i1}} & \ldots & \frac{w(r_{ik})}{r_{ik}}
\end{bmatrix}^T,
\quad
r_{ij} \le r_{i(j + 1)},
\end{equation}

\noindent where $1 \le j \le k$ and $w$ is a smooth cutoff function,

\begin{align}\label{eq:cutoff}
w(r) =
\begin{cases}
\left[1 - \left(\frac{r}{r_c}\right)^2\right]^2, & 0 \le r \le r_c, \\
0, & r > r_c.
\end{cases}
\end{align}

\noindent As $\X_i^{(1)}$ accounts only for two-body terms, we proposed a second term $\X_i^{(2)}$ that accounts for three-body terms by measuring the distances between neighbors of atom $i$:

\begin{equation}\label{eq:x2-short}
X_{in}^{(2)} =
\left\langle
\frac{\sqrt{w(r_{ij}) w(r_{il})}}{r_{jl}}
\right\rangle_n,
\quad
j, l \in \mathcal{N}(i),~
X_{in} \ge X_{i(n + 1)},
\end{equation}

\noindent where $j$ and $l$ are atoms in the neighborhood $\mathcal{N}(i)$ of $i$, $\langle . \rangle_n$ denotes the average over the $n$-th terms, and $1 \le n \le k - 1$.
The final descriptor $\X_i$ is formed by concatenating $\X_i^{(1)}$ and $\X_i^{(2)}$.
We used the default hyperparameters for the representation, as published in our prior work, namely a number of $k = 32$ neighbors and a cutoff of $r_c = 5$~\AA~ were used to represent each environment.
Extensive details on the representation are provided in the original reference.\cite{schwalbekoda2025information}

\noindent\textbf{Information entropy:}
the information entropy $\Info$ of a set of descriptors $\Xset$ encodes the amount of information contained in the descriptor distribution.\cite{schwalbekoda2025information}
We quantified the information entropy from a set $\Xset$ of $N$ feature vectors $\X_i$ using a kernel density estimate over the empirical distribution of points,

\begin{equation}\label{eq:info}
    \Info (\mathbf{\{X\}}) = -\frac{1}{N}\sum_{i=1}^{N} \log \left[
    \frac{1}{N}\sum_{j=1}^{N} K_h(\X_i, \X_j)
    \right],
\end{equation}

\noindent where our choice of the kernel $K_h$ is the Gaussian kernel,

\begin{equation}
    K_h(\mathbf{X}_i, \mathbf{X}_j) = \text{exp}\Biggl(\frac{-||\mathbf{X}_i - \mathbf{X}_j||^2}{2h^2} \Biggr)
\end{equation}

\noindent and $h$ is the bandwidth of the kernel.
For the bandwidth, we selected the default bandwidth of $h = 0.015$ \AA$^{-1}$ in the QUESTS method, which has proven useful for a range of applications and datasets.\cite{schwalbekoda2025information}
Throughout this work, the natural logarithm was used for computing the information entropy, scaling the units of information to nats.

\noindent\textbf{Differential entropy:}
To determine whether individual data points $\Y$ are contained in the distribution induced by $\Xset$, we compute the relative information provided by $\Y$ given the current knowledge of $\Xset$, defining the so-called differential entropy $\dH$ as:\cite{schwalbekoda2025information}

\begin{equation}\label{eq:dH}
    \dH (\Y | \Xset) = - \log \left[
    \sum_{i=1}^{N}K_h(\Y, \X_i)
    \right].
\end{equation}

\noindent With this definition, $\dH \leq 0$ only if $\Y$ is contained in the distribution induced by $\Xset$.
If $\dH > 0$, then $\Y$ provides additional ``surprise'' to the dataset $\Xset$.
Within Eq. \eqref{eq:dH}, the kernel $K_h$ is chosen to be Gaussian, the bandwidth $h$ is equal to 0.015 \AA$^{-1}$, and the unit of $\dH$ is nats, thus mirroring Eq. \eqref{eq:info}.

\noindent\textbf{Dataset diversity:} The diversity of a given set of descriptors $\Xset$ is defined as:\cite{schwalbekoda2025information}

\begin{equation}\label{eq:diversity}
    \mathcal{D} (\Xset) = \log \left[
    \sum_{i=1}^{N} e^{\delta \mathcal{H}(\X_i | \Xset)}
    \right],
\end{equation}

\noindent which approximates a measure of the support of the distribution.
While the dataset diversity $\mathcal{D}$ is an approximation and has not been mathematically proven to be an exact measure, it is quite useful in defining the size of the phase space.
The entropy $\Info$, on the other hand, takes into account the probability of each data point and may be low even for high-coverage phase spaces.
For instance, a near-degenerate distribution (e.g., nearly all the probability is concentrated in a single point) exhibits low information entropy, even though its support may be vast.
In this case, its diversity will remain high despite the low entropy.

\subsection{Evaluation methods}

\begin{figure}[htb!]
    \centering
    \includegraphics[width=\textwidth]{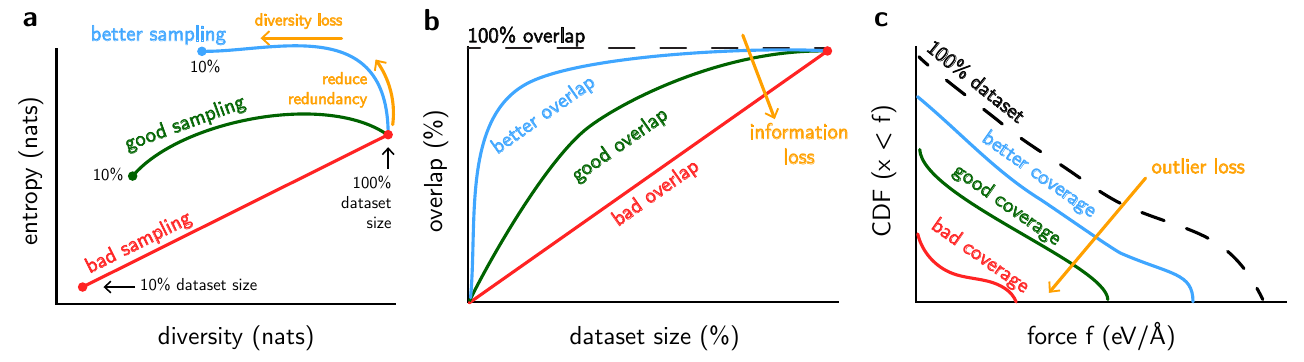}
    \caption{
    \textbf{Figures of merit that quantify the efficiency of a dataset compression algorithm.}
    \textbf{a}, An ideal algorithm should retain as much as the original dataset diversity as possible while removing redundancies from the data. In an entropy vs. diversity plot, this is depicted by curves that increase in entropy without reducing the diversity. At higher levels of compression, however, information loss is inevitable, and dataset diversity naturally start to decrease.
    \textbf{b}, An ideal algorithm should maximize the overlap of a compressed dataset with the original, complete dataset. Overlaps near 100\% indicate a compression with effectively no information loss.
    \textbf{c}, An ideal dataset compression method should retain outlier environments as much as possible. Given that high-force environments tend to be outliers in datasets sampled from molecular dynamics or optimization trajectories, preserving this ``long tail'' of the force distribution is evidence of an algorithm that prioritizes these outlier environments. This long tail is represented by the cumulative distribution function (CDF) of forces.
    }
    \label{fig:evalMethods}
\end{figure}

Evaluating the performance of a dataset subsampling technique based on test errors obtained from models trained on these smaller datasets is the usual method to measure subsampling quality, but it relies on somewhat expensive model training runs and is prone to the variability in the training process and in the test distributions.
Alternatively, our model-free analysis measure the effectiveness of the compression algorithms in preserving the original distribution of the dataset by quantifying dataset completeness and overlap between datasets from atomistic simulations.\cite{schwalbekoda2025information}
In this work, we use all these metrics to compare the efficiency of dataset compression algorithms in subsampling atomistic datasets.
Figure \ref{fig:evalMethods} illustrates three of such assessment techniques, each of which is discussed below.

\noindent\textbf{Entropy and diversity:}
First, we define subsampling performance by using the entropy and diversity metrics defined in Eqs. \eqref{eq:info} and \eqref{eq:diversity}.
A good subsampling algorithm should maximize the entropy of a dataset (i.e., remove the redundancy in the data) while also preserving as much of its original diversity as possible.
To quantify this, we subsample a dataset at multiple fractions of its original size and calculate the entropy and diversity of the sampled dataset for different compression algorithms (Fig. \ref{fig:evalMethods}a).
For a constant dataset size, worse subsampling methods will lead to datasets with smaller entropy and diversity, indicating information losses when subsampling the data.

\noindent\textbf{Overlap:}
Given the importance of preserving the support of the data distribution, a second possible figure of merit is the overlap of a dataset $\Xset$ with respect to a reference dataset $\{\X_{\mathrm{ref}}\}$.
Intuitively, the overlap represents the proportion of environments from $\Xset$ which are contained in $\{\X_{\mathrm{ref}}\}$.
This can be quantified using the definition of $\dH$ in Eq.~\eqref{eq:dH} and the notion of ``contained'' as $\dH \leq 0$, as

\begin{equation}
    \text{Overlap} \left(\mathbf{\{X\}}|\mathbf{\{X_{ref}\}}\right)
    = \frac{\#  \ \text{envs with } \delta \mathcal{H}(\X_i \in \Xset | \{\mathbf{X_{ref}}\}) \leq 0}{\# \text{ envs in } \{\textbf{X}\}}.
\end{equation}

\noindent Given that the overlap cannot be larger than 100\%, a good compression algorithm should maximize the overlap of the subsampled dataset with respect to the original dataset (Fig. \ref{fig:evalMethods}b).
On the other hand, a worse compression algorithm quickly leads to information loss as soon as the dataset size is reduced, especially for low-redundancy datasets.
This loss of information is undesirable in a range of scenarios in MLIPs, given that the cost associated with generating and labeling the data may be higher than including an additional point to the training set.

\noindent\textbf{Dataset efficiency:} Similarly to the concept of ``efficiency'' in information theory,\cite{shannon1948mathematical} we define the efficiency of a dataset containing $n$ environments as the information entropy of that dataset divided by the maximum entropy of the dataset ($\log n$).
A dataset with 100\% efficiency necessarily has a uniform distribution of environments, thus without any overlaps between environments.
On the other hand, datasets with low efficiency contain a limited amount of information, but high number of environments, indicating oversampling or redundancy in the data.

\noindent\textbf{Long tail of force distributions:}
Finally, beyond descriptor distributions, we evaluate the performance of compressed datasets according to the distribution of force magnitudes across all environments.
Given that high-force configurations are typically rarer than low-force configurations, preserving the ``long tail'' of the distribution of forces is important when subsampling datasets.
A more effective compression algorithm is expected to produce a force distribution with a longer tail, i.e., one that captures a wider range of force magnitudes, indicating broader coverage of the original dataset and retention of outliers.
To represent this long tail, we compute a cumulative distribution function (CDF) of force magnitudes, $\mathrm{CDF}(x < f)$, as a function of the threshold $f$ (Fig. \ref{fig:evalMethods}c).
With this definition, a reduced loss of outliers is represented by a higher value of the CDF at higher thresholds $f$.

\subsection{Baseline dataset compression methods}

To assess common baseline subsampling algorithms, we implemented three main algorithms: mean farthest-point sampling (FPS), k-means clustering, and random selection.
For all these algorithms, we focus on datasets of structures with periodic boundary conditions where each cell may exhibit large variability in the number of atoms.
The assumption of the compression methods in this work is that all environments of structures in the dataset will be used during the training/testing process, regardless of their (possible) internal redundancies.
Many methods, on the other hand, do not require entire structures, but only selected environments for training.
In these cases, all algorithms could be used in exactly the same way, but with a per-atom treatment rather than a per-structure one.

\noindent \textbf{Random Sampling:} Within random sampling, the most trivial of subsampling strategies, structures (i.e., sets of descriptors $\Xset$) are randomly selected given a target fraction of the dataset size (see Algorithm \ref{alg:randomAlgorithm}).
In this case, representations are not needed in order to subsample the datasets.
Sometimes, random sampling from uniform bins (e.g., energy levels) is used to increase the diversity of the data,\cite{botu2017machine} but in this work we consider only random samples of the dataset.

\begin{algorithm}
    \caption{Random sampling}\label{alg:randomAlgorithm}
    \begin{algorithmic}[1]
        \REQUIRE Set of $N$ structures $\mathcal{S} = \{S_i\}$ with environments $S_i = \{\X_1^{(i)}, \ldots , \X_n^{(i)}\}$, target size $K \leq N$
        \STATE Randomly select $K$ structures $S_i^{(c)}$ from $\mathcal{S}$ without replacement
        \RETURN $\mathcal{C} = \{S_1^{(c)}, \ldots, S_K^{(c)}\}$
    \end{algorithmic}
\end{algorithm}

\noindent \noindent \textbf{$k$-means sampling:} Within $k$-means sampling, we represent each structure of the dataset by taking the mean of all per-atom representations in the structure.
In particular, the representation we adopted is the one within the QUESTS method (see Eqs. \eqref{eq:x1-short} and \eqref{eq:x2-short}), as it is later used to compute entropy or diversity values.
Then, we cluster these per-structure vectors into $k$ groups using a typical $k$-means clustering method and randomly sample one structure from each cluster.
The sampled structures are used to create a compressed dataset (see Algorithm \ref{alg:KMeans}).
We used the $k$-means clustering method implemented in \texttt{scikit-learn} (v. 1.7.0).
This approach has been used multiple times in the development of datasets for MLIPs.\cite{williams2024active,lebeda2024k}
Variations of this method employ hierarchical clustering rather than $k$-means to stratify the space of features in clusters which contain similar information.\cite{sivaraman2020machine,qi2024robust} 

\begin{algorithm}[H]
    \caption{$k$-means sampling}\label{alg:KMeans}
    \begin{algorithmic}[1]
        \REQUIRE Set of $N$ structures $\mathcal{S} = \{S_i\}$ with environments $S_i = \{\X_1^{(i)}, \ldots , \X_n^{(i)}\}$, target size $K \leq N$
        \STATE Calculate a per-structure representation $\mathbf{S}_i$ from per-atom representations of $S_i$, $\mathbf{S}_i = \langle \X_{j}^{(i)} \rangle$
        \STATE Cluster the set of structure vectors $\{ \mathbf{S}_i \}$ in $K$ clusters using a $k$-means clustering method
        \FOR{$n = 1$ \textbf{to} $K$}
            \STATE \textbf{1.} Select a random element from cluster $n$
            \STATE \textbf{2.} Add the selected structure $S_n^{(c)}$  to the compressed set $\mathcal{C}$
        \ENDFOR
        \RETURN $\mathcal{C} = \{S_1^{(c)}, \ldots, S_K^{(c)}\}$
    \end{algorithmic}
\end{algorithm}

\noindent \textbf{Mean Farthest Point Sampling:}
In mean farthest point sampling (FPS), we begin with per-structure representations obtained by averaging the per-atom QUESTS representations across each structure, as used for the $k$-means sampling method.
The algorithm starts by randomly selecting one structure vector as the initial seed for the compressed dataset.
Next, we iteratively compute the Euclidean distances between all vectors in the compressed dataset and those remaining in the full dataset.
At each iteration, the structure whose mean vector has the largest average distance to all vectors in the compressed dataset is selected, added to the compressed dataset, and removed from the pool of remaining candidates.
This process continues until the compressed dataset reaches the size specified by the user. (see Algorithm \ref{alg:MFPS}).
Variations of FPS have been extensively employed in subsampling datasets,\cite{eldar1997farthest} including datasets for MLIPs.\cite{Bartok2017,imbalzano2018automatic,rowe2020accurate,wengert2022hybrid,boulangeot2024hydrogen,li2024local,zhang2025exploring,gurlek2025accurate,trestman2025GradientGuided}
Related algorithms based on matrix factorization methods such as CUR\cite{mahoney2009cur} have also been used to subsample atomistic datasets by maximizing the distance between representations.\cite{deringer2017machine,imbalzano2018automatic,dragoni2018achieving,bernstein2019novo,bartok2018machine}

\begin{algorithm}[H]
    \caption{Mean farthest point sampling}\label{alg:MFPS}
    \begin{algorithmic}[1]
        \REQUIRE Set of $N$ structures $\mathcal{S} = \{S_i\}$ with environments $S_i = \{\X_1^{(i)}, \ldots , \X_n^{(i)}\}$, target size $K \leq N$
        \STATE Calculate a per-structure representation $\mathbf{S}_i$ from per-atom representations of $S_i$, $\mathbf{S}_i = \langle \X_{j}^{(i)} \rangle$
        \STATE Randomly select a structure $S_i^{(c)}$ from $\mathcal{S}$ and add it to the initialized compressed set $\mathcal{C}_1$
        \FOR{$n = 1$ \textbf{to} $K - 1$}
            \STATE \textbf{1.} Calculate the distances $d\left(S_i, S_j^{(c)}\right) =
            \norm{\mathbf{S}_i - \mathbf{S}_j^{(c)}}_2$ between $S_i \in \mathcal{S} \setminus \mathcal{C}_n$ and $S_j^{(c)} \in \mathcal{C}_n$
            \STATE \textbf{2.} Find $S_{n+1}^{(c)} = \displaystyle\argmax_i \displaystyle\sum_{j=1}^{n} d\left(S_i, S_j^{(c)}\right)$
            \STATE \textbf{3.} Define $\mathcal{C}_{n + 1} = \mathcal{C}_n \cup \{ S_{n+1}^{(c)} \}$
            \ENDFOR
        \RETURN $\mathcal{C} = \{S_1^{(c)}, \ldots, S_K^{(c)}\}$
    \end{algorithmic}
\end{algorithm}

\subsection{Compressing atomistic datasets with a minimum set cover algorithm}
\label{sec:algo}

One of the main problems of baseline methods is the reliance on per-structure representations, as most modern MLIPs require entire structures (i.e., collections of atoms) to be added to a training set rather than individual environments.
In this context, one can easily propose counterexamples for which data subsampling is less efficient due to average, per-structure representations rather than sets of per-atom vectors (see \supptext).
To address this problem, we propose a new atomistic dataset subsampling algorithm as an instance of the minimum set cover (MSC) problem.
The set cover problem is a classical mathematical formulation defined as follows: given a domain $\mathcal{U}$ and a family of subsets $\mathcal{S}$ of $\mathcal{U}$, a set cover is a subfamily $\mathcal{C} \subseteq \mathcal{S}$ such that the union of all sets in $\mathcal{C}$ equals $\mathcal{U}$.\cite{korte2018Combinatorial}
The minimum set coverage corresponds to the smallest subfamily $\mathcal{C}$ that fully covers the domain $\mathcal{U}$.
From an atomistic perspective, the domain $\mathcal{U}$ represents the set of all distinct atomic environments $\X_j$ contained in the original dataset, and thus differs slightly from the original MSC problem due to its continuous space.
Nevertheless, these environments are grouped into different sets of environments (i.e., structures) $S_i = \Xset_j^{(i)} $, and $\mathcal{S} = \{S_1, \ldots, S_N\}$ denotes the initial dataset containing $N$ structures.
The sets $S_i$ may have overlapping environments and typically contain different numbers of atoms.
Thus, the goal of atomistic dataset compression is analogous to the set cover problem, aiming to identify a subset of structures $\mathcal{C} = \{ S_1^{(c)}, \ldots, S_K^{(c)} \} \subseteq \mathcal{S}$, with $K \leq N$, such that the union of all environments in $\mathcal{C}$ span the original universe $\mathcal{U}$.

Though the optimization of the set cover problem is NP-hard,\cite{lund1994Hardness} we used a greedy algorithm to approximate the MSC of atomistic datasets in polynomial time.\cite{chvatal1979Greedy}
Within this approach, the main rule is to iteratively select sets with the largest number of uncovered elements until the desired dataset size is achieved, thus maximizing the coverage of environments with the minimum number of sets.
Using the information-theoretical quantities defined in the Methods, the atomistic version of this algorithm is initialized by first selecting the element with the highest per-structure information entropy, i.e., the structure that exhibits a large number of distinct environments and has the minimum possible redundancy.
Then, at each step $n$, we iteratively compute the values of $\dH (\mathcal{S} \setminus \mathcal{C}_n \ |\ \mathcal{C}_n)$, where $\mathcal{C}_n$ is the set of selected structures at iteration $n$.
To maximize the coverage, the new structure $S_{n+1}$ is selected according to

\begin{equation}\label{eq:msc-rule}
    S_{n+1} = \argmax_i \left[
        \max_{\X_j^{(i)} \in S_i} \dH(\X_j^{(i)} | \mathcal{C}_n) + \Info(S_i)
    \right],
\end{equation}

\noindent where $\dH$ is computed for all environments $\X_j^{(i)}$ of each structure $S_i \in \mathcal{S} \setminus \mathcal{C}_n$.
Finally, $\mathcal{C}_{n+1} = \mathcal{C}_n \cup \{ S_{n+1} \}$.
The selection rule from Eq. \eqref{eq:msc-rule} balances two objectives:
(1) it prioritizes new structures that contain a large number of uncovered environments according to the environments already present in the compressed dataset ($\max \dH$);
and (2) it selects structures that are diverse rather than structures exhibiting redundant environments ($\Info$).
The pseudocode for this proposed method is shown in Algorithm \ref{alg:msc}.

\begin{algorithm}[H]
    \caption{Minimum set coverage}\label{alg:msc}
    \begin{algorithmic}[1]
        \REQUIRE Set of $N$ structures $\mathcal{S} = \{S_i\}$ with environments $S_i = \{\X_1^{(i)}, \ldots , \X_n^{(i)}\}$, target size $K \leq N$
        \STATE Calculate the values of per-structure information entropy $\Info(S_i)$
        \STATE Initialize the compressed set $\mathcal{C}_1 = \{ \displaystyle\argmax_i \Info(S_i) \}$
        \FOR{$n = 1$ \textbf{to} $K - 1$}
            \STATE \textbf{1.} For each structure $S_i = \{ \X_j^{(i)} \}$, compute $\dH (\X_j^{(i)} | \Xset_n^{(c)})$, where $\Xset_n^{(c)}$ are all envs. in $\mathcal{C}_{n}$
            \STATE \textbf{2.} Choose $S_{n + 1}^{(c)}$ according to Eq. \eqref{eq:msc-rule}
            \STATE \textbf{3.} Define $\mathcal{C}_{n + 1} = \mathcal{C}_n \cup \{ S_{n+1}^{(c)} \}$
        \ENDFOR
        \RETURN $\mathcal{C} = \{S_1^{(c)}, \ldots, S_K^{(c)}\}$
    \label{algorithm:MSC}
    \end{algorithmic}
\end{algorithm}

\subsection{Machine learning interatomic potential}

\noindent\textbf{Model training approach:} To evaluate the effect of sampling method on MLIP performance, we trained models using subsampled datasets containing 10\%, 25\%, 50\%, 75\%, and 100\% of the original data.
Each model was trained for a fixed total number of structure-epochs to compare performances across dataset sizes.
Accordingly, models trained on 10\%, 25\%, 50\%, 75\%, and 100\% of the data were run for 10,000, 4,000, 2,000, 1,333, and 1,000 epochs, respectively.
Across all the dataset sizes, a batch size of 40 was used, except for the 10\% dataset size, where a batch size of 4 was used.
We randomly split each subsampled dataset at ratios of 80:10:10 for train:validation:test, where a single test set was shared among all experiments.
The best performing model at each training run was determined as the one minimizing the validation loss.

\noindent\textbf{SevenNet architecture:}
The MLIP used in this work was the SevenNet architecture from Park \textit{et al.},\cite{Park2024seven} which is based on the NequIP architecture.\cite{Batzner2022}
We used the SevenNet code available at \url{https://github.com/MDIL-SNU/SevenNet} (v. 0.2.0).
The model was created with three equivariant blocks with $L = 2$ and node features with irreducible representations of 32 scalars ($\ell = 0$), 32 vectors ($\ell = 1$), and 32 tensors ($\ell = 2$).
The convolutional filter used a cutoff radius $r_c$ of 5 \AA~ and a tensor product of learnable radial functions with 8 radial Bessel functions and spherical harmonics up to $\ell = 2$.

\noindent\textbf{SevenNet training:}
The SevenNet models were trained with the Adam optimizer,\cite{kingma2015Adam,reddi2018convergence} starting with a learning rate of 0.005, which decayed by a factor of 0.99 per epoch.
We used the Huber loss function\cite{huber1964Robust} as defined in the SevenNet paper.\cite{Park2024seven}
The weight of the loss due to errors in forces was selected as 0.1, whereas the weight due to stresses was set to $10^{-6}$.
The value of $\delta$ in the Huber loss that controls the transition between the mean squared error loss to a mean absolute error\cite{huber1964Robust} was set to 0.01.
The dataset was shifted according to the mean per-atom energy and rescaled according to the root mean square of the forces, statistics that were obtained from each training dataset.

\section{Results}

\subsection{Comparing compression algorithms with the GAP-20 carbon dataset}
\label{sec:comparing}

\begin{figure}[htb!]
    \centering
    \includegraphics[width=\textwidth]{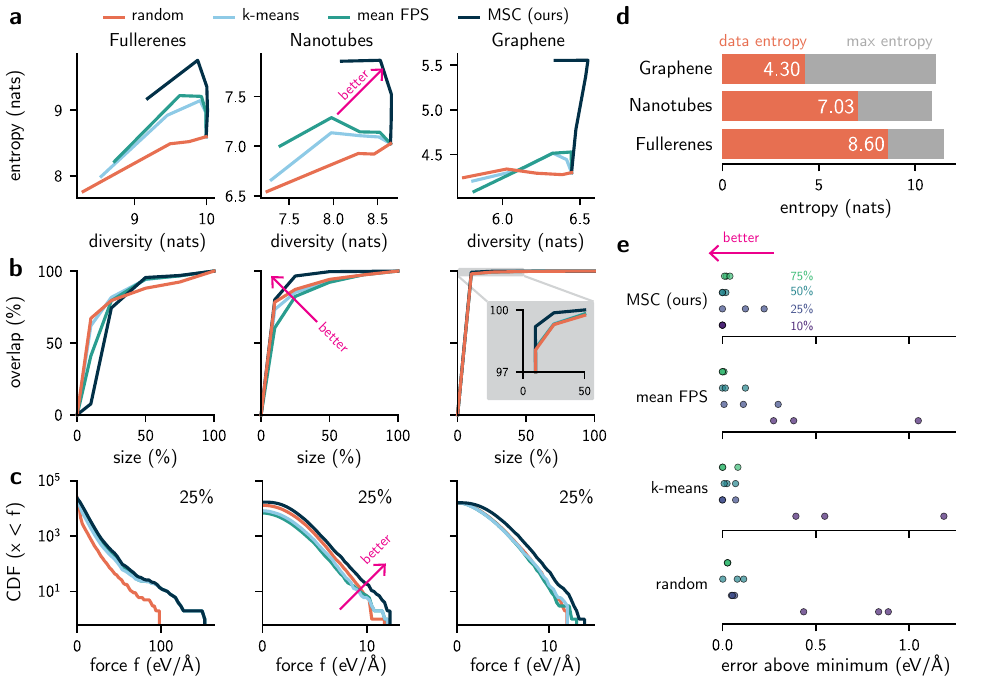}
    \caption{
    Performance of compression algorithms in subsampling three subsets of the GAP-20 dataset (Fullerenes, Nanotubes, Graphene).
    Each dataset was subsampled at four different sizes: 75, 50, 25, and 10\%, and compared against the full, original dataset.
    \textbf{a}, our MSC method exhibits the best behavior in terms of simultaneous entropy increases and diversity retention for the selected datasets across dataset sizes. On the other hand, random sampling quickly leads to information loss, as shown by immediate diversity decreases upon subsampling.
    \textbf{b}, datasets compressed with MSC exhibit the higher overlap with the original dataset, with a single exception of Fullerenes sampled at 10\% of its original size.
    \textbf{c}, MSC preserves the long tail of the distribution of forces of the environments, as shown by a higher cumulative distribution function (CDF) given a force threshold for datasets compressed at 25\% of their original size (see also Fig. \ref{fig:si:LongTailGAP20}).
    \textbf{d}, Information entropy (orange) of the three subsets of GAP-20 and the maximum value of entropy that would be possible in a dataset with that size (gray). The numerical values of entropy (in nats) are shown with white numbers. The differences between the gray and orange bars show that the subsets exhibit different levels of redundancy, explaining the results in \textbf{a}--\textbf{c}.
    \textbf{e}, Average force error (eV/\AA) above the minimum value of error given a subset (i.e., Graphene, Nanotubes, or Fullerenes) and a dataset size (i.e., 10, 25, 50, or 75\%) across algorithms.
    While no algorithm consistently outperforms the others, models trained on datasets compressed using our MSC method exhibit a smaller range of errors above the minimum across sampling sizes, with errors more tightly grouped around optimal performance compared to other methods.
}
    \label{fig:gap20}
\end{figure}

\begin{figure}[h]  
    \centering
    \includegraphics[width=\textwidth]{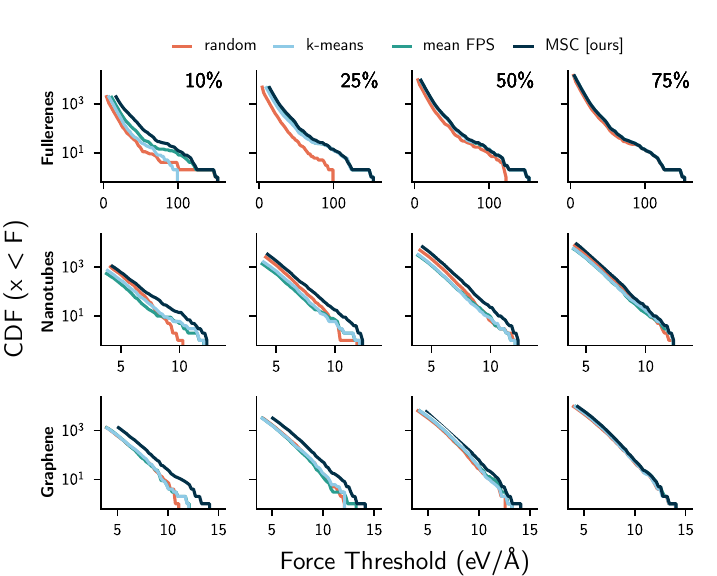}
    \caption{Distribution of forces in selected GAP-20 datasets (Fullerenes, Nanotubes, Graphene) subsampled at various compression levels (75\%, 50\%, 25\%, and 10\%) using different sampling methods. All distributions are plotted starting from the 80th percentile of the corresponding full GAP-20 dataset. The consistently higher coverage of large-magnitude forces achieved by MSC across all compression levels highlights the algorithm’s ability to better preserve the long tail of the force distribution compared to other methods.}
    \label{fig:si:LongTailGAP20}
\end{figure}

Using the figures of merit and methods defined above, we compared the performance of our information-theoretical algorithm from Sec. \ref{sec:algo} against the baseline ones by compressing three subsets of the GAP-20 carbon dataset from Rowe \textit{et al.}\cite{rowe2020accurate}
In our previous work, we showed that the subsets of this dataset exhibit varying degrees of redundancy, leading to distinct behaviors upon random sampling.\cite{schwalbekoda2025information}
Here, we investigated whether improved sampling efficiency could be achieved when different algorithms are used to subsample the data.
Figures \ref{fig:gap20}a--c compare the performance of all four algorithms in compressing the Graphene, Nanotubes, and Fullerenes subsets of GAP-20.
In all metrics and datasets, our MSC method outperforms all other baseline algorithms in compression efficiency.
Figure \ref{fig:gap20}a shows that MSC preserves the diversity of the dataset as much as possible upon compression, while other baseline decrease data diversity at lower compression rates than MSC.
The figure also shows that MSC leads to the highest increase in entropy of the subsampled dataset across all algorithms, demonstrating that the method efficiently removes redundancies in the data prior to information loss.
Figures \ref{fig:gap20}b,c corroborate these results, showing that MSC-compressed datasets have the highest overlap with the original dataset and the longest tail of the distribution of forces (see also Fig. \ref{fig:si:LongTailGAP20}).
The only exception to this result is the Fullerenes subset of GAP-20 compressed to 10\% of its original size, where all algorithms exhibit large information loss, including MSC (see analysis in Sec. \ref{sec:outliers}).
This is partially a consequence of the overall higher dataset efficiency of Fullerenes compared to Nanotubes or Graphene (Fig. \ref{fig:gap20}d), for which higher data compression implies information loss.

The results above show that it is possible to measure information loss in subsampled datasets without training MLIPs.
This model-free approach is useful, as it avoids expensive model training runs and provides a algorithms independent of model architecture.
Furthermore, these metrics provide a way to define an optimal compression level instead of relying on error scans.
Nevertheless, it is still relevant to measure whether MLIPs trained on subsampled datasets exhibit a reasonable range of test errors despite being trained on a small data regime.
Using the SevenNet architecture\cite{Park2024seven} as described in the Methods, we trained models on compressed datasets from each algorithm and size, and evaluated them on a common test set to measure the behavior of force errors when data compression is achieved.
Figure \ref{fig:gap20}e compares these errors for all models and data sizes by subtracting the minimum error from each dataset type and size across algorithms.
In this case, points closer to zero would indicate that the algorithm that created the subsampled dataset leads to models with lower errors.
On the other hand, higher errors indicate that the algorithm often leads to training sets that degrade model performance on the test set.
As shown on Fig. \ref{fig:gap20}e, no algorithm systematically outperforms the others when it comes to the lowest possible error, which can be partly explained by the variability in the model training process and the specific test datasets used to quantify error.
However, MLIPs trained on datasets compressed with MSC are more likely to exhibit lower errors compared to MLIPs trained on datasets compressed with other algorithms, as seen by smaller ranges of the distribution of errors under MSC.
This result is particularly pronounced at the very low data regimes (10\%), where MSC achieves the lowest possible errors among the three subsets of the GAP-20 dataset.
This demonstrates that dataset compression methods can have a large impact in model generalization when datasets are undersampled (low-data regime), while preserving the original distribution of points and errors when the data redundancy is high.
All numerical results supporting these conclusions are shown in the Supplementary Information (Tables \ref{tab:si:gap-gr-entropy}-\ref{tab:si:gap-errors}).

\subsection{Explaining information loss upon dataset compression}
\label{sec:outliers}

\begin{figure}[htb!]
    \centering
    \includegraphics[width=\textwidth]{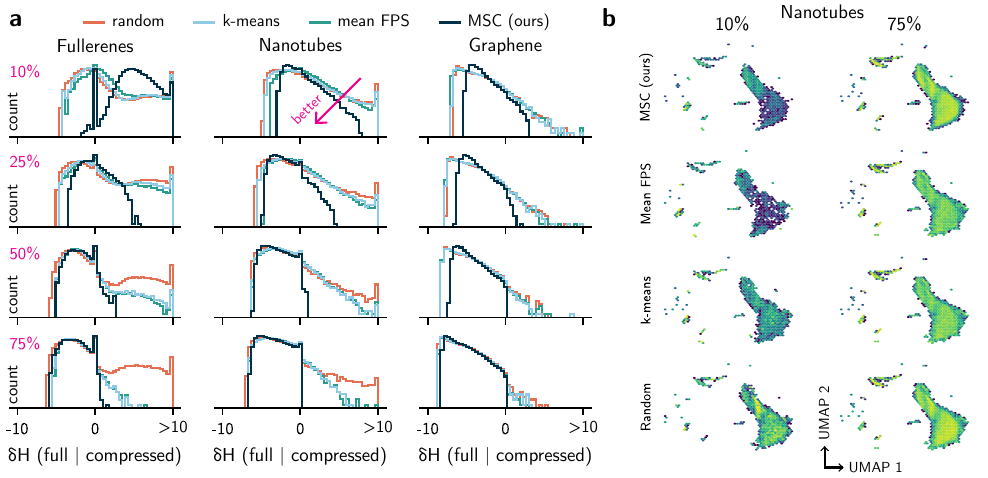}
    \caption{
        Analysis of outlier loss across algorithms and dataset sizes for the three subsets of GAP-20 under study (Graphene, Nanotubes, and Fullerenes).
        \textbf{a}, the distribution of $\dH$ of environments in the full dataset w.r.t the compressed dataset across different methods indicates that MSC minimizes outlier loss, as shown by distributions of $\dH$ shifted towards less positive values.
        The only exception to this is the case of Fullerenes compressed at 10\% of its original size, where the algorithm prioritized reducing the number of extreme outliers (i.e., those with $\dH > 10$ nats) at the expense of a reduction of true overlap ($\dH \leq 0$) with the dataset.
        \textbf{b}, a low-dimensional representation of the datasets, obtained with a UMAP visualization, illustrates how outlier loss is not trivially detected with these methods. When comparing the distribution of environments within the 10\% and 75\% datasets of Nanotubes, the information loss is at best qualitative, and at worst misleading. On all UMAP plots, the axes are the arbitrary axes of UMAP and brighter colors indicate a larger number of points within each region.
    }
    \label{fig:umap}
\end{figure}

\begin{figure}[h]  
    \centering
    \includegraphics[width=0.8\textwidth]{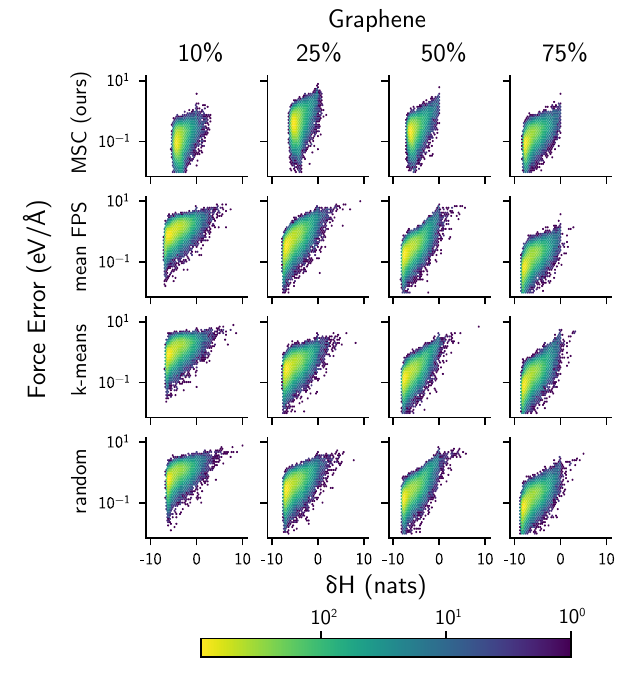}
    \caption{
    Distribution of forces error for a SevenNet model trained on compressed datasets and tested on the original, full dataset of Graphene in GAP-20.
    The distribution is correlated with the $\dH(\mathrm{full} | \mathrm{compressed})$, with very positive values of $\dH$ representing environments in the original dataset that are outside of the distribution of environments in the compressed dataset.
    Across all subsampling rates, MSC results in fewer very positive $\dH$, demonstrating a smaller loss of outliers without compromising accuracy.
    Brighter colors represent a higher number of environments in each region of the space.
    }
    \label{fig:si:ErrordHGraphene}
\end{figure}

\begin{figure}[h]  
    \centering
    \includegraphics[width=0.8\textwidth]{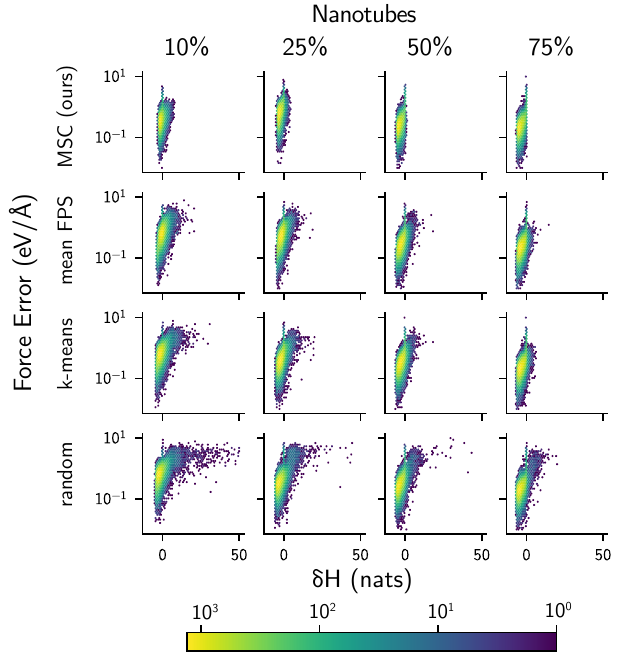}
    \caption{
    Distribution of forces error for a SevenNet model trained on compressed datasets and tested on the original, full dataset of Nanotubes in GAP-20.
    The distribution is correlated with the $\dH(\mathrm{full} | \mathrm{compressed})$, with very positive values of $\dH$ representing environments in the original dataset that are outside of the distribution of environments in the compressed dataset.
    Across all subsampling rates, MSC results in fewer very positive $\dH$, demonstrating a smaller loss of outliers without compromising accuracy.
    Brighter colors represent a higher number of environments in each region of the space.
    }
    \label{fig:si:ErrordHNanotubes}
\end{figure}

\begin{figure}[h]  
    \centering
    \includegraphics[width=0.8\textwidth]{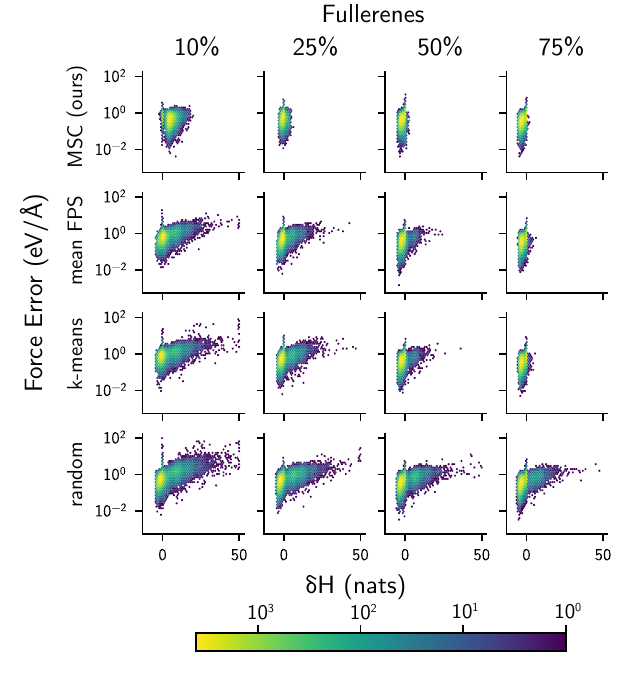}
    \caption{
        Distribution of forces error for a SevenNet model trained on compressed datasets and tested on the original, full dataset of Fullerenes in GAP-20.
        The distribution is correlated with the $\dH(\mathrm{full} | \mathrm{compressed})$, with very positive values of $\dH$ representing environments in the original dataset that are outside of the distribution of environments in the compressed dataset.
        Across all subsampling rates, MSC results in fewer very positive $\dH$, demonstrating a smaller loss of outliers without compromising accuracy.
        Brighter colors represent a higher number of environments in each region of the space.
        }
    \label{fig:si:ErrordHFullerenes}
\end{figure}

To explain the results from the previous section, we investigated how the algorithms lead to information loss when compressing datasets.
However, instead of relying on the aggregate dataset metrics from Figs. \ref{fig:gap20}a--c, we analyzed individual environments in each dataset and tested their impact in the model errors shown in Fig. \ref{fig:gap20}e.
Figure \ref{fig:umap}a illustrates the distribution of $\dH(\X_i | \mathcal{C})$ as a function of compression level and algorithm, where $\X_i$ are all environments in the original, complete dataset.
Given that $\dH$ measures how close each environment $\X_i$ is to the distribution of environments in the compressed dataset $\mathcal{C}$, an ideal algorithm would lead to as many negative $\dH$ values  as possible.
In all but one case in Fig. \ref{fig:umap}a, datasets compressed with MSC exhibit $\dH$ distributions with fewer positive values, indicating that the compressed datasets indeed cover as much as possible of the original dataset.
On the other hand, the baseline algorithms almost always lead to loss of information, as seen by distributions with high values of $\dH$.
The only exception to this scenario is the Fullerenes subset of GAP-20 compressed to 10\% of its original size.
To explain this discrepancy, we note that the MSC algorithm attempted to minimize the number of environments very far away from the original distribution, as illustrated by the number of data points with $\dH > 10$ nats in the plot.
In the process, the subsampled structures end up reasonably close to the original dataset (see range of $0 < \dH < 10$ nats for 10\% Fullerenes), creating a dataset with low overlap with the original, but fewer true outliers as well.
These results refine the conclusions from the previous section.
When the errors from Fig. \ref{fig:gap20}e are broken down by environment rather than treated as average results, it becomes clear that MSC reduces the number of data points treated as outliers and, as a consequence, constrain generalization errors from the MLIPs.
Figures \ref{fig:si:ErrordHGraphene}--\ref{fig:si:ErrordHFullerenes} further show that, by reducing the number of environments from the original dataset that become outliers (i.e., exhibit large $\dH$ values) in the compressed data, models are less likely to operate under ``out-of-distribution'' scenarios ($\dH > 0$).

\begin{figure}[h]  
    \centering
    \includegraphics[width=0.8\textwidth]{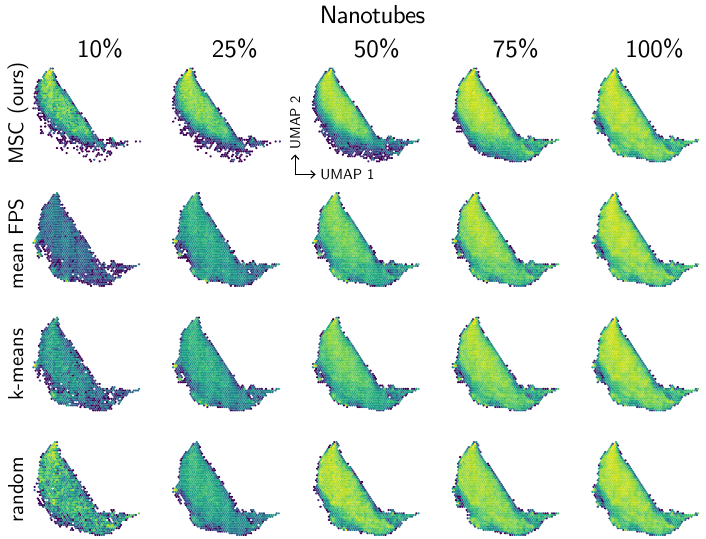}
    \caption{Two-dimensional projection of per-atom descriptors from compressed datasets of the Graphene subset of GAP-20. The projection was performed using UMAP, and the two axes correspond to the two components of UMAP. Brighter colors represent a higher number of environments in each region of the 2D space. As discussed in the main text, the representations provide little information on the ability of each algorithm to compress the dataset.}
    \label{fig:si:UMAPGraphene}
\end{figure}

\begin{figure}[h]  
    \centering
    \includegraphics[width=0.8\textwidth]{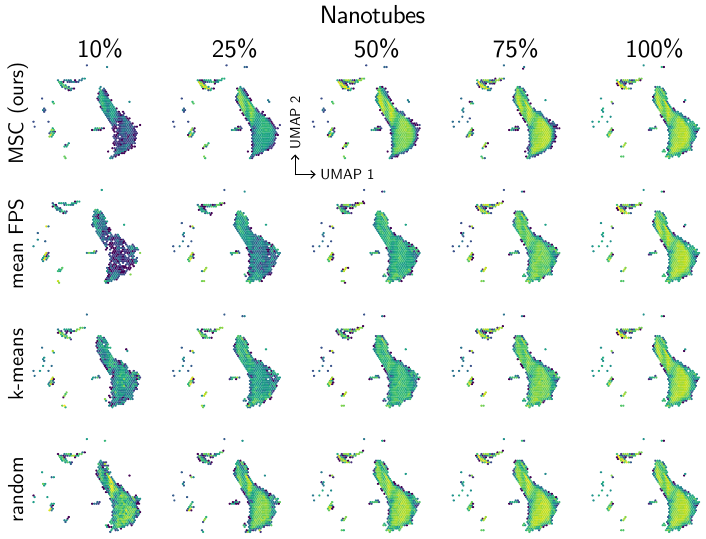}
    \caption{Two-dimensional projection of per-atom descriptors from compressed datasets of the Nanotubes subset of GAP-20. The projection was performed using UMAP, and the two axes correspond to the two components of UMAP. Brighter colors represent a higher number of environments in each region of the 2D space. As discussed in the main text, the representations provide little information on the ability of each algorithm to compress the dataset.}
    \label{fig:si:UMAPNanotubes}
\end{figure}

\begin{figure}[h]  
    \centering
    \includegraphics[width=0.8\textwidth]{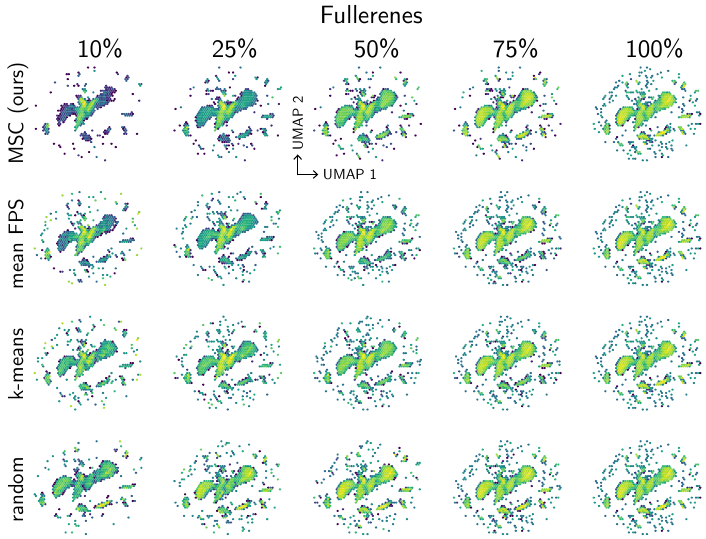}
    \caption{Two-dimensional projection of per-atom descriptors from compressed datasets of the Fullerenes subset of GAP-20. The projection was performed using UMAP, and the two axes correspond to the two components of UMAP. Brighter colors represent a higher number of environments in each region of the 2D space. As discussed in the main text, the representations provide little information on the ability of each algorithm to compress the dataset.}
    \label{fig:si:UMAPFullerenes}
\end{figure}

As an alternative to our $\dH$ analysis, dataset subsampling is often examined in the literature using dimensionality reduction techniques.
In such approaches, a 2D projection of the subsampled dataset is overlaid on top of that of the original dataset, and the overlap between the two distributions is qualitatively analyzed.
To evaluate whether this visualization method could provide insights into the information loss caused by subsampling similarly to $\dH$, we used a visualization created with UMAP\cite{McInnes2018,mcinnes2020UMAP} to compare datasets subsampled with different algorithms.
The UMAP transformation was fitted to the distribution of QUESTS descriptors from the original, full subsets of the GAP-20 dataset and then applied consistently to all subsampled datasets.
Figure \ref{fig:umap}b illustrates how the 2D distribution of points for the Nanotubes subset changes as we reduce the dataset size from 75\% to 10\% (see all results in Figs. \ref{fig:si:UMAPGraphene}--\ref{fig:si:UMAPFullerenes}).
In contrast with the the metrics in Fig. \ref{fig:gap20} or the outlier analysis in Fig. \ref{fig:umap}a, this 2D visualization is unable to quantify information losses in the data.
It can be misleading to interpret the ``blank areas'' within the main region of the 2D UMAP plots at 10\% size as a sign of higher ``information losses''.
If that assumption were true, it could suggest that random sampling is the best algorithm, as it seemingly covers more of the two-dimensional space represented with the UMAP visualization.
As shown before, the actual coverage is minimized with the random sampling algorithm, which exhibits the worst information loss when subsampling atomistic datasets.
Similar results can be found when comparing all 2D visualizations for the Graphene and Fullerene datasets (Figs. \ref{fig:si:UMAPGraphene},\ref{fig:si:UMAPFullerenes}), where the 2D distribution of data points is less representative of the true dataset coverage than an analysis in higher-dimensional spaces.
This provides a cautionary tale on the use of low-dimensional representations when compressing datasets, supporting a more quantitative measure of distributions and coverage of local environments using information theory.

\subsection{Comparing compression algorithms with the TM23 dataset}

\begin{figure}[h]  
    \centering
    \includegraphics[width=\textwidth]{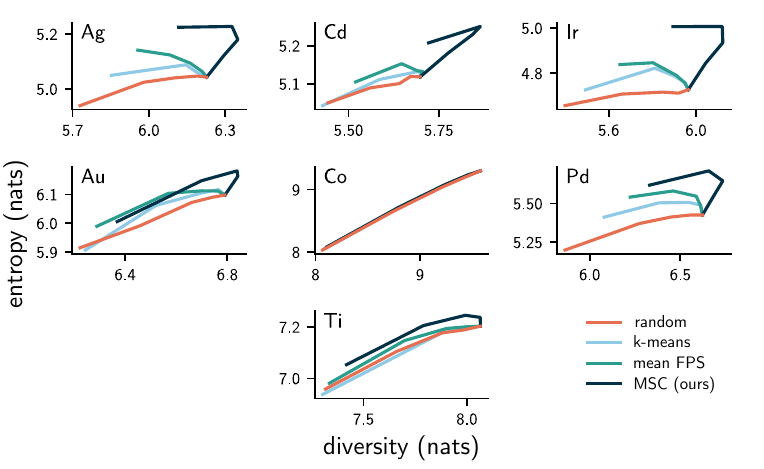}
    \caption{
    Performance of compression algorithms in subsampling seven subsets of the TM23 dataset according to their overlap with the original dataset.
    Each dataset was subsampled at four different sizes: 75, 50, 25, and 10\%, and compared against the full, original dataset.
    In nearly all cases, our MSC method exhibits the best behavior in terms of simultaneous entropy ($H$) increases and diversity ($D$) retention for the selected datasets across dataset sizes.
    On the other hand, random sampling quickly leads to information loss, as shown by immediate diversity decreases upon subsampling.
    }
    \label{fig:si:TM23_H_Div}
\end{figure}

\begin{figure}[h]  
    \centering
    \includegraphics[width=\textwidth]{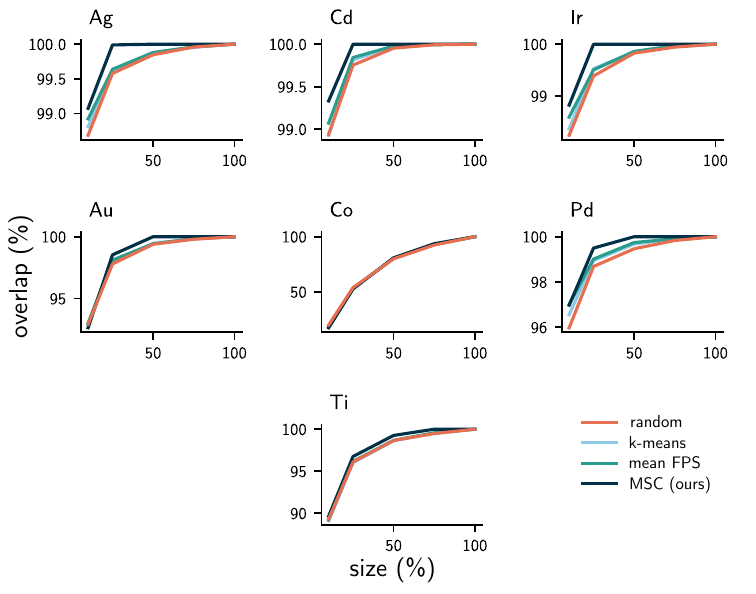}
    \caption{Performance of compression algorithms in subsampling seven subsets of the TM23 dataset according to their overlap with the original dataset.
    Each dataset was subsampled at four different sizes: 75, 50, 25, and 10\%, and compared against the full, original dataset.
    In all cases, our MSC method exhibits the best behavior in terms of retaining a high overlap with the original datasets upon compression.
    }
    \label{fig:si:TM23_Overlap}
\end{figure}

\begin{figure}[h]  
    \centering
    \includegraphics[width=0.8\textwidth]{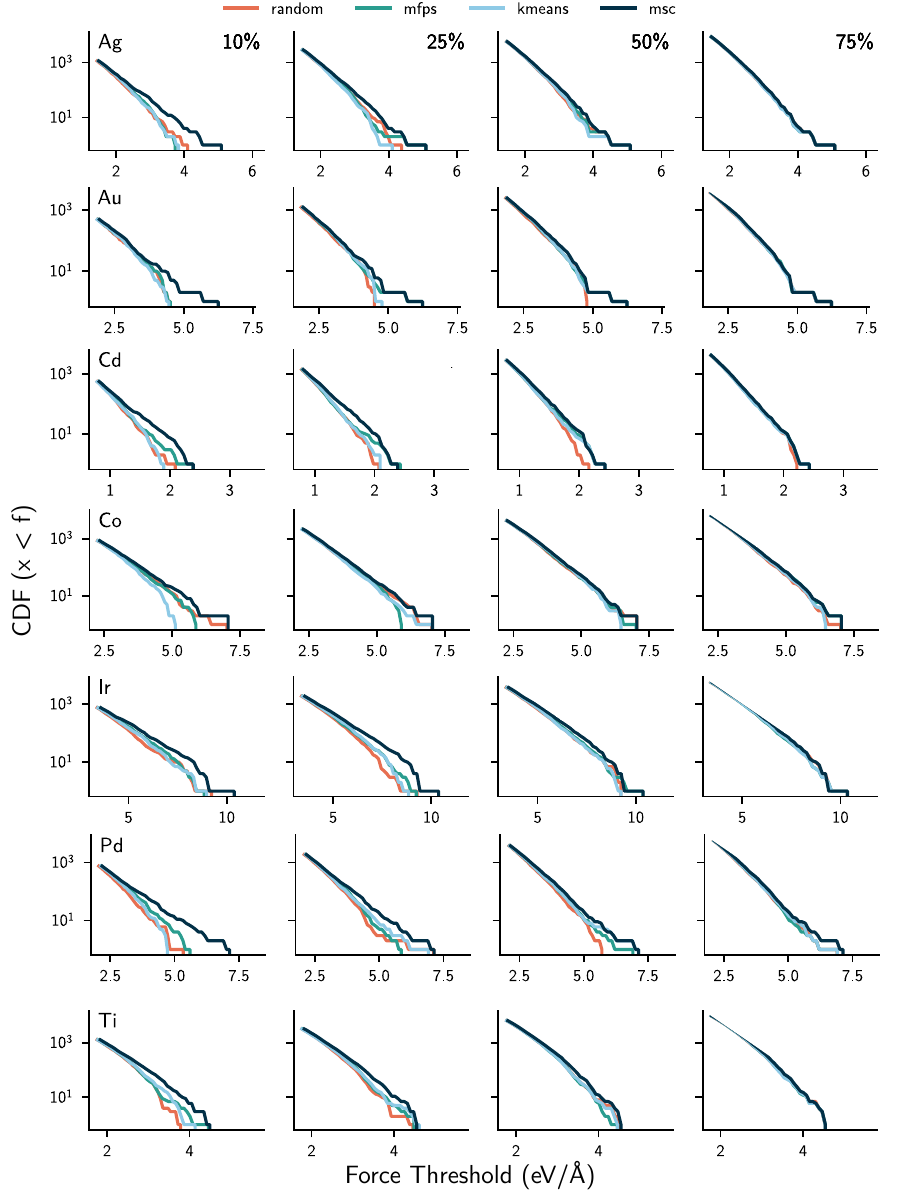}
    \caption{
    Performance of compression algorithms in subsampling seven subsets of the TM23 dataset according to their ability to retain the long tail of forces.
    Each dataset was subsampled at four different sizes: 75, 50, 25, and 10\%.
    In all cases, our MSC method exhibits the highest ability to preserve higher (and more rare) forces upon compression, as evidenced by the higher values of the cumulative density function (CDF).}
    \label{fig:si:TM23_PDF}
\end{figure}

As a second case study, we analyzed the compression of the TM23 dataset from Owen \textit{et al}.\cite{owen2024complexity}
This dataset exhibits several PESes, but here we focus on seven elements whose errors are representative of the original TM23 analysis: Ag, Au, Cd, Co, Ir, Pd, and Ti, all within the ``warm'' subset of the dataset (i.e., sampled at 75\% of the melting point of each metal).
Specifically, results for four elements are shown in the main text: Ag (very low test errors of MLIPs), Ir (low errors), Co (medium errors), and Ti (high errors), with the remaining data shown in the Supplementary Information.
Figures~\ref{fig:tm23}a--c show that MSC outperforms all other algorithms in preserving dataset diversity, increasing entropy, maintaining higher overlap with the original distribution, and retaining the long tail behavior of the distribution of forces (see also Figs. \ref{fig:si:TM23_H_Div}--\ref{fig:si:TM23_PDF}, Tables \ref{tab:si:tm23-ag}--\ref{tab:si:tm23-ti}).
Interestingly, the datasets exhibited different behavior under compression.
Ag and Ir have large redundancies in the data (see also Fig. \ref{fig:tm23}d), as demonstrated by overlaps close to 100\% across dataset sizes.
Even dataset sizes as small as 10\% retain most of the original information in the data, with high data diversity and overlaps around 99\% for Ag and Ir across algorithms (see discussion on the diversity increases in the \supptext).
On the other hand, Ti and Co rapidly lose information upon compression, reaching 90\% and 20\% overlaps with the original dataset, respectively, when compressed to 10\% of their original sizes.
The smaller compressibility of Co is explained by the small redundancy in the original data (Fig. \ref{fig:tm23}d).
Whereas a slight entropy increase can be seen when compressing Ti (Fig. \ref{fig:tm23}a), any subsampling in Co leads to information loss.

\begin{figure}[h]  
    \centering
    \includegraphics[width=0.8\textwidth]{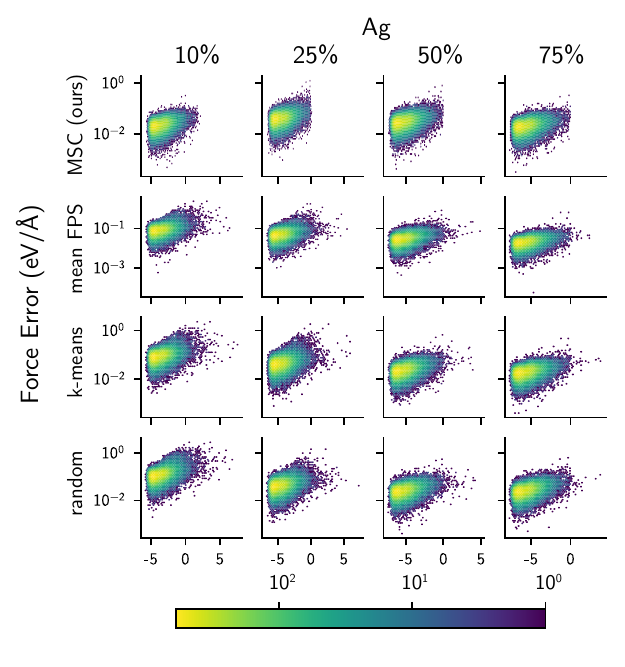}
    \caption{
    Distribution of forces error for a SevenNet model trained on compressed datasets and tested on the original, full dataset of Ag (warm) in TM23.
    The distribution is correlated with the $\dH(\mathrm{full} | \mathrm{compressed})$, with very positive values of $\dH$ representing environments in the original dataset that are outside of the distribution of environments in the compressed dataset.
    Brighter colors represent a higher number of environments in each region of the space.
    }
    \label{fig:si:Ag_Error_dH}
\end{figure}

\begin{figure}[h]  
    \centering
    \includegraphics[width=0.8\textwidth]{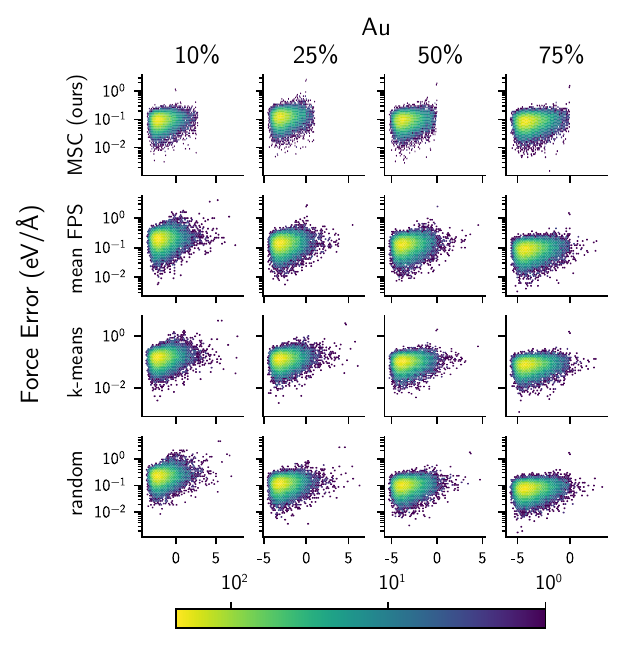}
    \caption{Distribution of forces error for a SevenNet model trained on compressed datasets and tested on the original, full dataset of Au (warm) in TM23.
    The distribution is correlated with the $\dH(\mathrm{full} | \mathrm{compressed})$, with very positive values of $\dH$ representing environments in the original dataset that are outside of the distribution of environments in the compressed dataset.
    Brighter colors represent a higher number of environments in each region of the space.}
    \label{fig:si:Au_Error_dH}
\end{figure}

\begin{figure}[h]  
    \centering
    \includegraphics[width=0.8\textwidth]{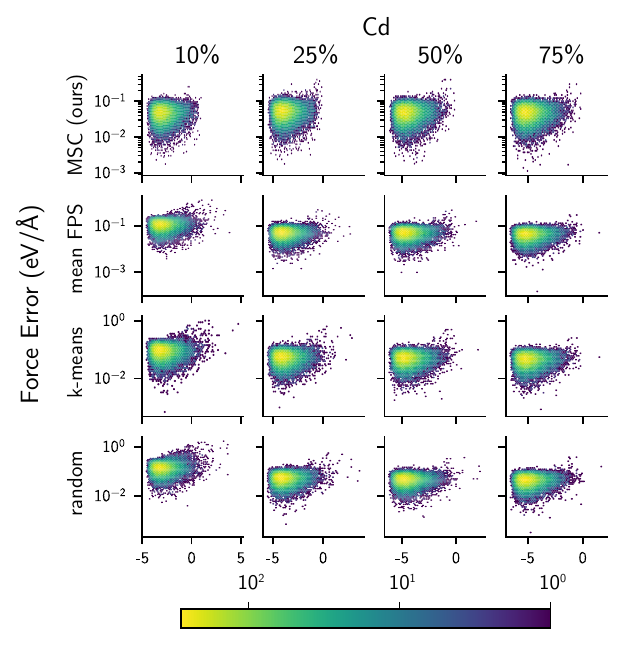}
    \caption{Distribution of forces error for a SevenNet model trained on compressed datasets and tested on the original, full dataset of Cd (warm) in TM23.
    The distribution is correlated with the $\dH(\mathrm{full} | \mathrm{compressed})$, with very positive values of $\dH$ representing environments in the original dataset that are outside of the distribution of environments in the compressed dataset.
    Brighter colors represent a higher number of environments in each region of the space.}
    \label{fig:si:Cd_Error_dH}
\end{figure}

\begin{figure}[h]  
    \centering
    \includegraphics[width=0.8\textwidth]{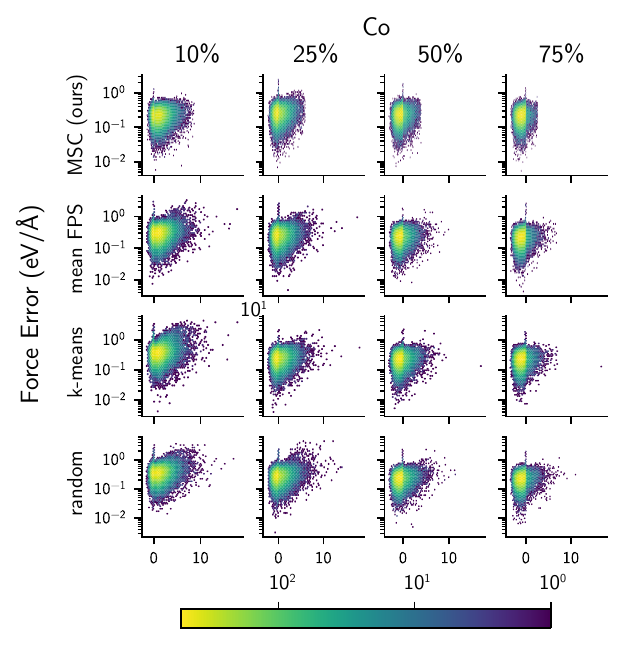}
    \caption{Distribution of forces error for a SevenNet model trained on compressed datasets and tested on the original, full dataset of Co (warm) in TM23.
    The distribution is correlated with the $\dH(\mathrm{full} | \mathrm{compressed})$, with very positive values of $\dH$ representing environments in the original dataset that are outside of the distribution of environments in the compressed dataset.
    Brighter colors represent a higher number of environments in each region of the space.}
    \label{fig:si:Co_Error_dH}
\end{figure}

\begin{figure}[h]  
    \centering
    \includegraphics[width=0.8\textwidth]{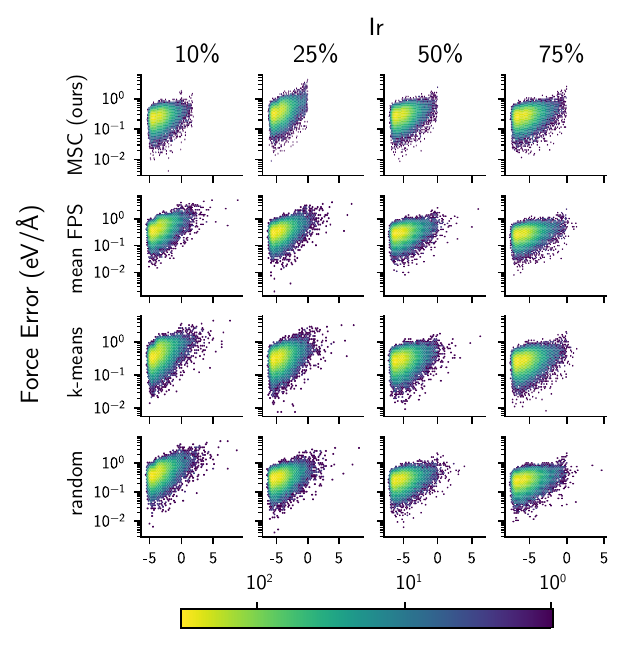}
    \caption{Distribution of forces error for a SevenNet model trained on compressed datasets and tested on the original, full dataset of Ir (warm) in TM23.
    The distribution is correlated with the $\dH(\mathrm{full} | \mathrm{compressed})$, with very positive values of $\dH$ representing environments in the original dataset that are outside of the distribution of environments in the compressed dataset.
    Brighter colors represent a higher number of environments in each region of the space.}
    \label{fig:si:Ir_Error_dH}
\end{figure}

\begin{figure}[h]  
    \centering
    \includegraphics[width=0.8\textwidth]{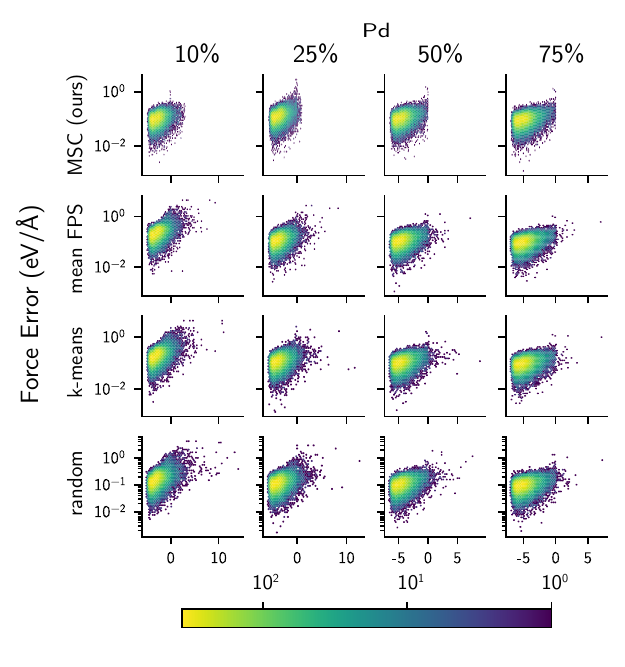}
    \caption{Distribution of forces error for a SevenNet model trained on compressed datasets and tested on the original, full dataset of Pd (warm) in TM23.
    The distribution is correlated with the $\dH(\mathrm{full} | \mathrm{compressed})$, with very positive values of $\dH$ representing environments in the original dataset that are outside of the distribution of environments in the compressed dataset.
    Brighter colors represent a higher number of environments in each region of the space.}
    \label{fig:si:Pd_Error_dH}
\end{figure}

\begin{figure}[h]  
    \centering
    \includegraphics[width=0.8\textwidth]{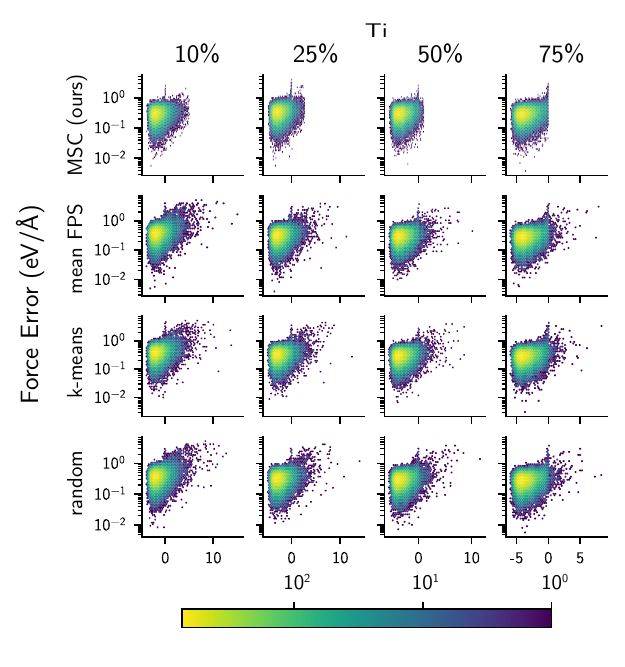}
    \caption{Distribution of forces error for a SevenNet model trained on compressed datasets and tested on the original, full dataset of Ti (warm) in TM23.
    The distribution is correlated with the $\dH(\mathrm{full} | \mathrm{compressed})$, with very positive values of $\dH$ representing environments in the original dataset that are outside of the distribution of environments in the compressed dataset.
    Brighter colors represent a higher number of environments in each region of the space.}
    \label{fig:si:Ti_Error_dH}
\end{figure}

\begin{figure}[h]  
    \centering
    \includegraphics[width=0.8\textwidth]{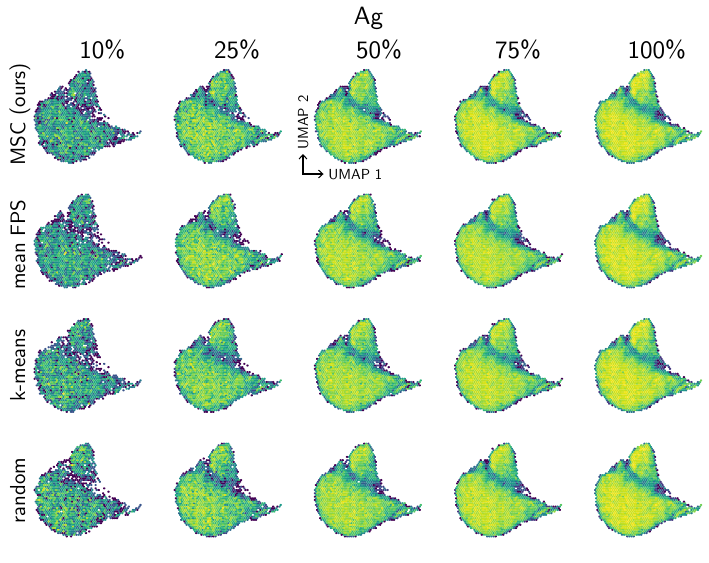}
    \caption{Two-dimensional projection of per-atom descriptors from compressed datasets of the Ag (warm) subset of TM23. The projection was performed using UMAP, and the two axes correspond to the two components of UMAP. Brighter colors represent a higher number of environments in each region of the 2D space.}
    \label{fig:si:Ag_UMAP}
\end{figure}

\begin{figure}[h]  
    \centering
    \includegraphics[width=0.8\textwidth]{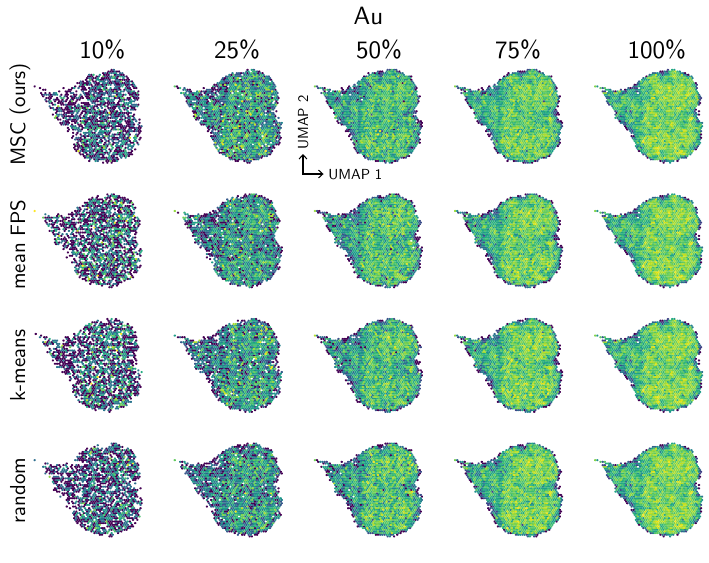}
    \caption{Two-dimensional projection of per-atom descriptors from compressed datasets of the Au (warm) subset of TM23. The projection was performed using UMAP, and the two axes correspond to the two components of UMAP. Brighter colors represent a higher number of environments in each region of the 2D space.}
    \label{fig:si:Au_UMAP}
\end{figure}

\begin{figure}[h]  
    \centering
    \includegraphics[width=0.8\textwidth]{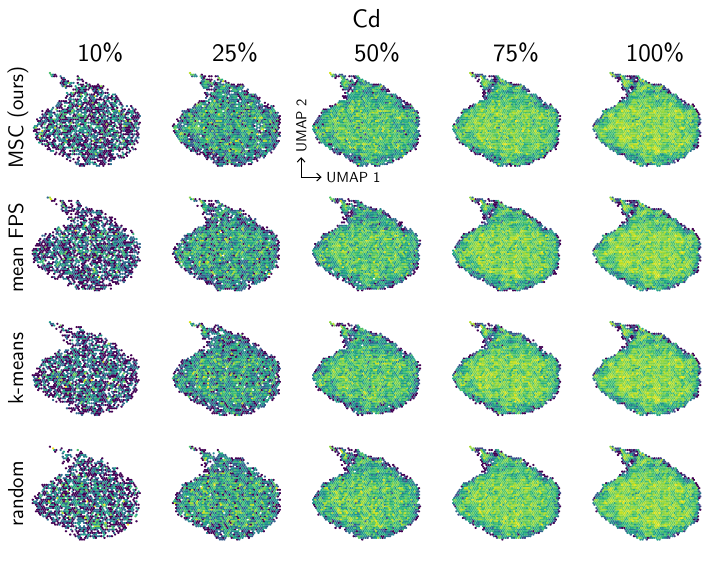}
    \caption{Two-dimensional projection of per-atom descriptors from compressed datasets of the Cd (warm) subset of TM23. The projection was performed using UMAP, and the two axes correspond to the two components of UMAP. Brighter colors represent a higher number of environments in each region of the 2D space.}
    \label{fig:si:Cd_UMAP}
\end{figure}

\begin{figure}[h]  
    \centering
    \includegraphics[width=0.8\textwidth]{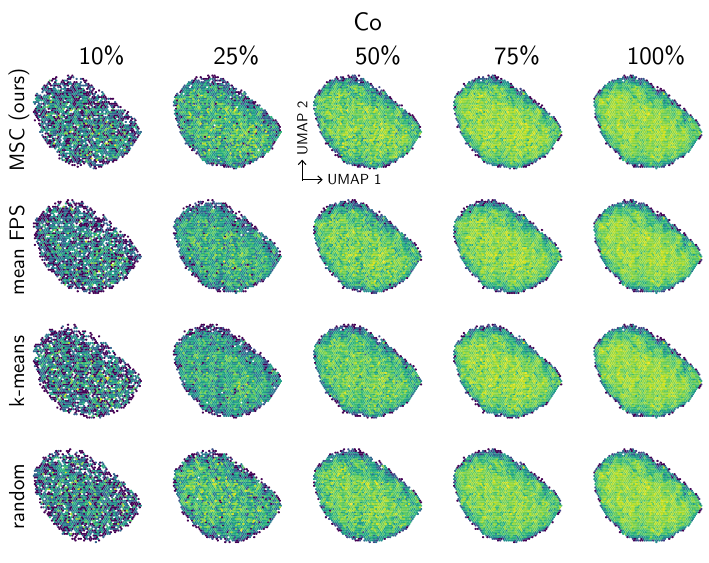}
    \caption{Two-dimensional projection of per-atom descriptors from compressed datasets of the Co (warm) subset of TM23. The projection was performed using UMAP, and the two axes correspond to the two components of UMAP. Brighter colors represent a higher number of environments in each region of the 2D space.}
    \label{fig:si:Co_UMAP}
\end{figure}

\begin{figure}[h]  
    \centering
    \includegraphics[width=0.8\textwidth]{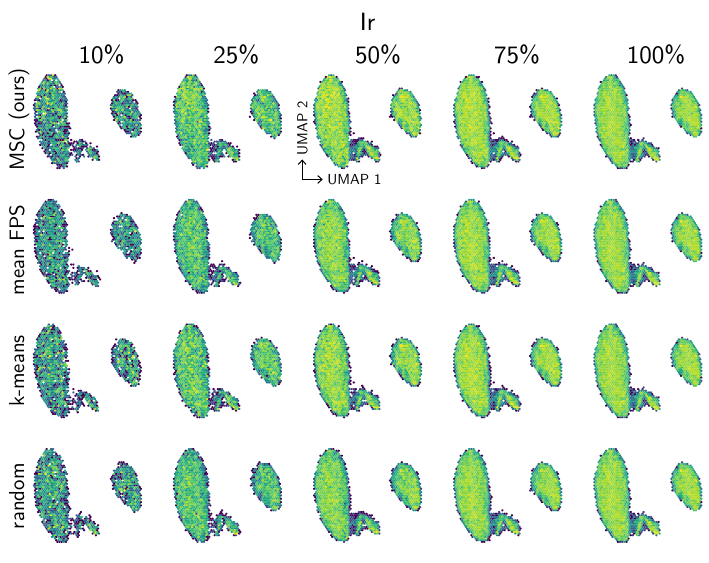}
    \caption{Two-dimensional projection of per-atom descriptors from compressed datasets of the Ir (warm) subset of TM23. The projection was performed using UMAP, and the two axes correspond to the two components of UMAP. Brighter colors represent a higher number of environments in each region of the 2D space.}
    \label{fig:si:Ir_UMAP}
\end{figure}

\begin{figure}[h]  
    \centering
    \includegraphics[width=0.8\textwidth]{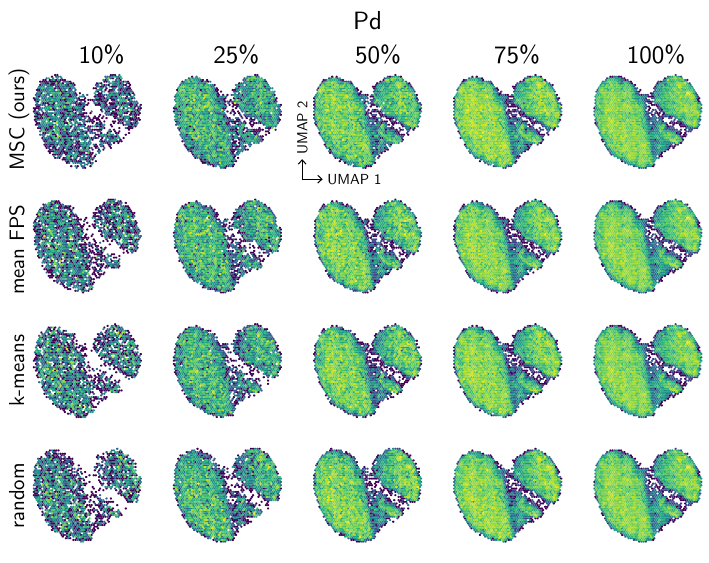}
    \caption{Two-dimensional projection of per-atom descriptors from compressed datasets of the Pd (warm) subset of TM23. The projection was performed using UMAP, and the two axes correspond to the two components of UMAP. Brighter colors represent a higher number of environments in each region of the 2D space.}
    \label{fig:si:Pd_UMAP}
\end{figure}

\begin{figure}[h]  
    \centering
    \includegraphics[width=0.8\textwidth]{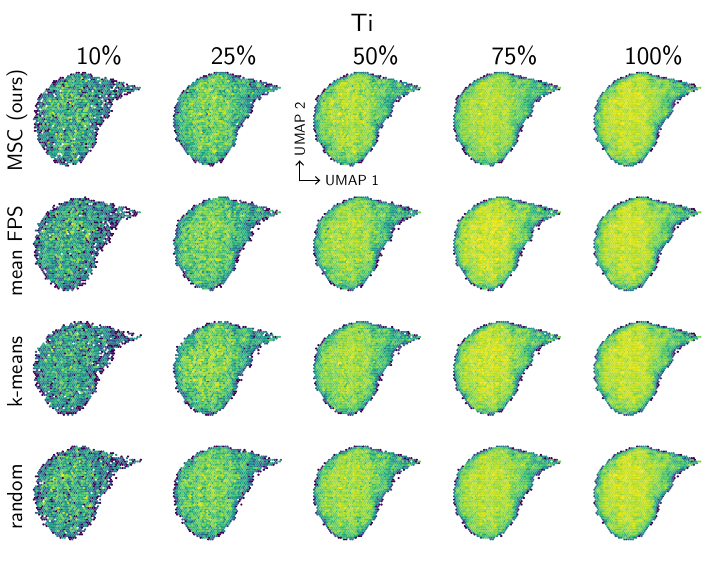}
    \caption{Two-dimensional projection of per-atom descriptors from compressed datasets of the Ti (warm) subset of TM23. The projection was performed using UMAP, and the two axes correspond to the two components of UMAP. Brighter colors represent a higher number of environments in each region of the 2D space.}
    \label{fig:si:Ti_UMAP}
\end{figure}

To evaluate the performance of MLIPs on the compressed datasets, we trained SevenNet models to each of the compressed subsets of TM23 and evaluated their error on shared, held-out test sets.
As seen with the GAP-20 dataset, errors of models trained on MSC-compressed datasets are often constrained compared to the other algorithms, especially in low-data regimes (Tables \ref{tab:si:tm23-errors-ag}--\ref{tab:si:tm23-errors-ti}).
Similarly to Fig. \ref{fig:umap}a, we observed that our MSC method leads to less outliers compared to the other algorithms even for the Co dataset (Figs. \ref{fig:si:Ag_Error_dH}--\ref{fig:si:Ti_Error_dH}), highlighting the importance of minimizing information loss and extreme outliers for MLIPs.
Though variability in model training and errors can make it challenging to compare errors across data sizes and subsampled datasets, our information-theoretical algorithm provides a robust method to compress atomistic datasets.
On the other hand, 2D visualizations created with UMAP are unable to provide quantitative insights on the dataset compression (Figs. \ref{fig:si:Ag_UMAP}--\ref{fig:si:Ti_UMAP}).

\begin{figure}[htb!]
    \centering
    \includegraphics[width=\textwidth]{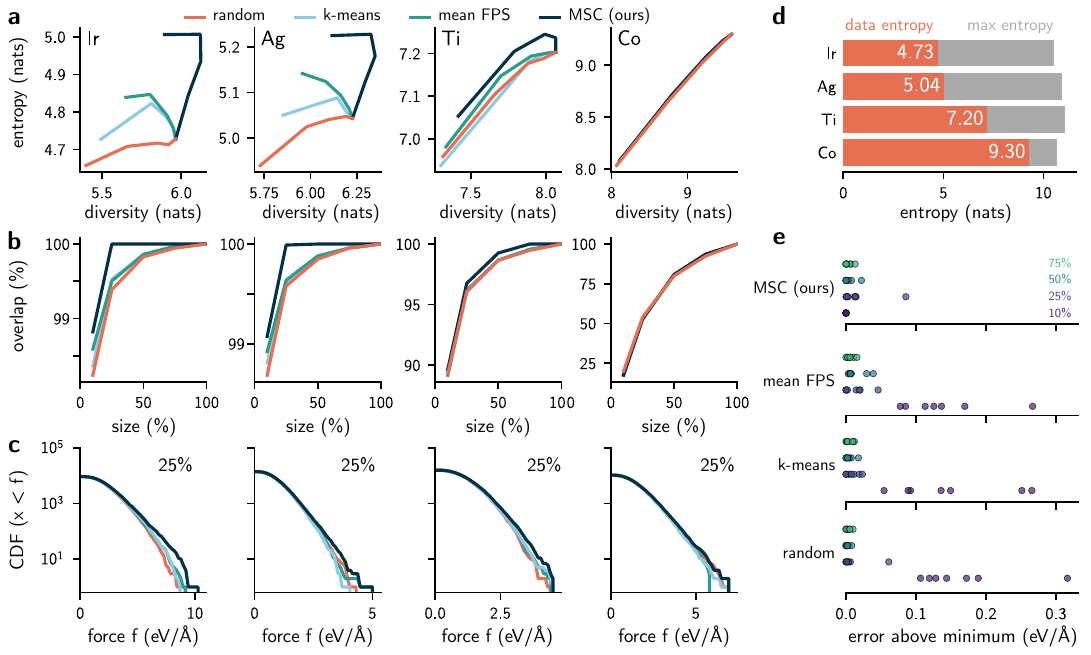}
    \caption{
        Performance of subsampling algorithms across a subset of the TM23 dataset (elements Ir, Ag, Ti, Co).
        The datasets are subsampled 75, 50, 25, and 10\% of their original sizes, and compared against the original datasets.
        \textbf{a}, our MSC method exhibits the best behavior in terms of simultaneous entropy increases and diversity retention for the subsets of TM23 across dataset sizes. In cases where there is lower data redundancy, such as the case of Ti or Co, information loss is inevitable.
        \textbf{b}, datasets compressed with MSC often exhibit the higher overlap with the original dataset compared to those obtained with other algorithms, except when redundancy in the original data is low, in which case all overlaps are similar.
        \textbf{c}, MSC preserves the long tail of the distribution of forces of the environments, as shown by a higher cumulative density function (CDF) given a force threshold $f$.
        \textbf{d}, Information entropy (orange) of the four subsets of TM23 and the maximum value of entropy that would be possible in a dataset with that size (gray). The numerical values of entropy are shown with white numbers, where the entropy is expressed in units of nats. The differences between the gray and orange bars show that the subsets contain different redundancy levels, explaining trends across elements in \textbf{a}--\textbf{c}.
        \textbf{e}, Average force error (eV/\AA) above the minimum value of error given a subset (i.e., Ir, Ag, Ti, or Co) and a dataset size (i.e., 10, 25, 50, or 75\%).
    Similarly to the GAP-20 example, while no algorithm consistently outperforms the other in terms of force errors, models trained on datasets compressed using our MSC method exhibit a smaller range of errors above the minimum across sampling sizes, with errors more tightly grouped around optimal performance in the low-data regime.
}
    \label{fig:tm23}
\end{figure}

\subsection{Compression of 64 single-element datasets from the ColabFit repository}

\begin{figure}[htb!]
    \centering
    \includegraphics[width=0.85\textwidth]{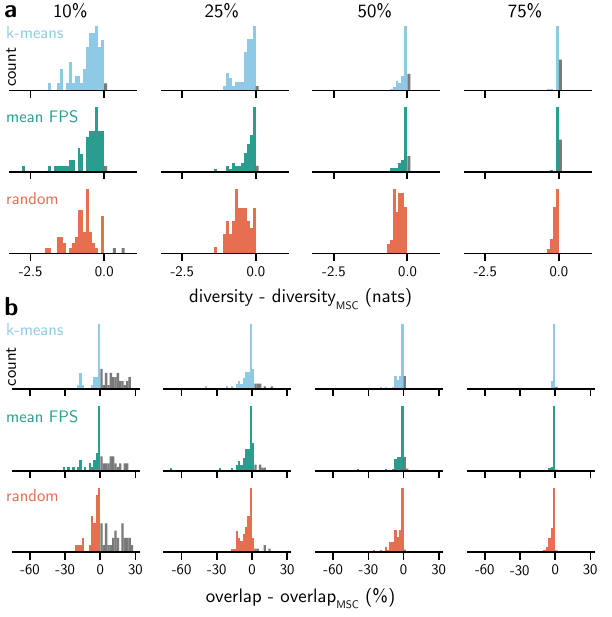}
    \caption{
    Performance of baseline algorithms against MSC across 64 different atomistic datasets from the ColabFit repository compressed to 10, 25, 50, and 75\% of their original sizes.
    Gray bars represent datasets for which the performance of MSC is worse than the baseline algorithm according to each metric.
    \textbf{a}, The difference in diversity between each method and MSC is mostly negative across all sizes and algorithms, demonstrating that MSC is a robust algorithm to preserve the original data distribution.
    \textbf{b}, Similarly, baseline algorithms lead to datasets with lower overlap against the original data compared to datasets subsampled with MSC.
    In the low-data regime, some algorithms showcase a higher overlap even though their diversity is lower.
    Similar to the Fullerenes example in Fig. \ref{fig:umap}, overlap does not fully account for how close outliers are to the distribution of points, but only for which points are entirely contained by a distribution.
    }
    \label{fig:misc}
\end{figure}

\begin{figure}[ht]  
    \centering
    \includegraphics[width=0.8\textwidth]{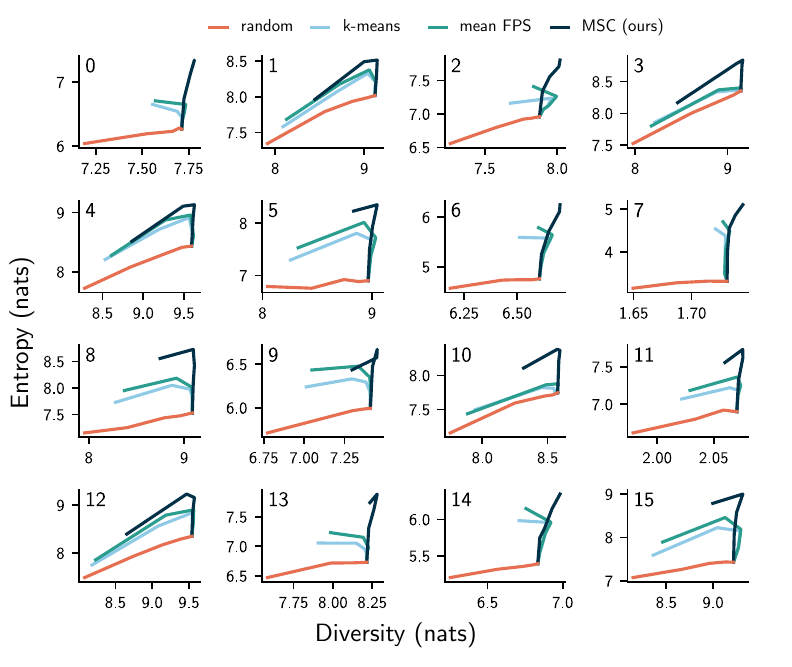}
    \caption{Relationship between entropy and diversity of miscellaneous datasets obtained from the ColabFit repository sampled at different sizes (100\%, 75\%, 50\%, 25\%, 10\%) and with different methods. Outside of a few outliers, datasets compressed with MSC resulted in higher diversity and entropy relative to datasets compressed with other methods, demonstrating that our compression algorithm is the most efficient in removing the redundancy of the data. The dataset numbers, shown on the top left of each plot, correspond to the ones shown in Table \ref{tab:si:colabfit-names}.}
    \label{fig:si:Misc_H_Div_0}
\end{figure}

\begin{figure}[h]  
    \centering
    \includegraphics[width=0.8\textwidth]{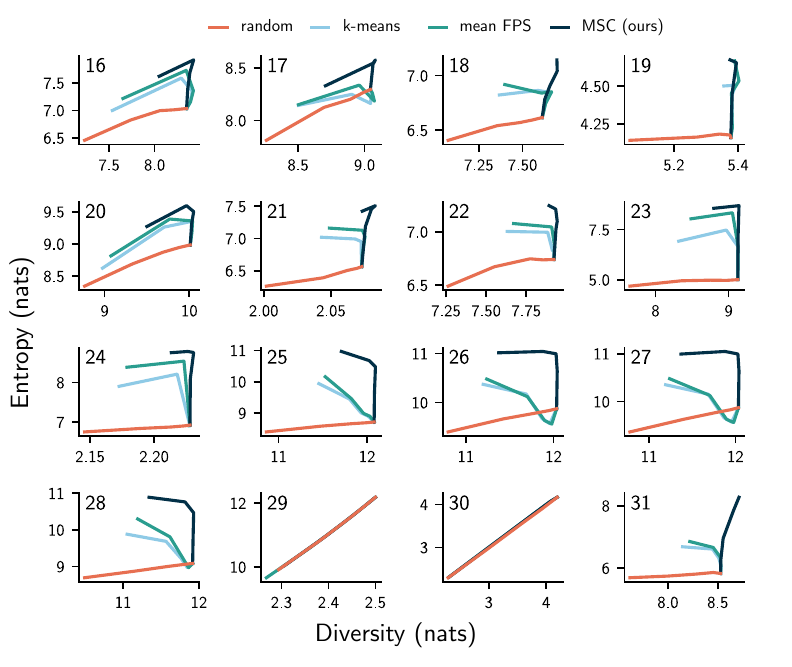}
    \caption{Relationship between entropy and diversity of miscellaneous datasets obtained from the ColabFit repository sampled at different sizes (100\%, 75\%, 50\%, 25\%, 10\%) and with different methods. Outside of a few outliers, datasets compressed with MSC resulted in higher diversity and entropy relative to datasets compressed with other methods, demonstrating that our compression algorithm is the most efficient in removing the redundancy of the data. The dataset numbers, shown on the top left of each plot, correspond to the ones shown in Table \ref{tab:si:colabfit-names}.}
    \label{fig:si:Misc_H_Div_1}
\end{figure}

\begin{figure}[h]  
    \centering
    \includegraphics[width=0.8\textwidth]{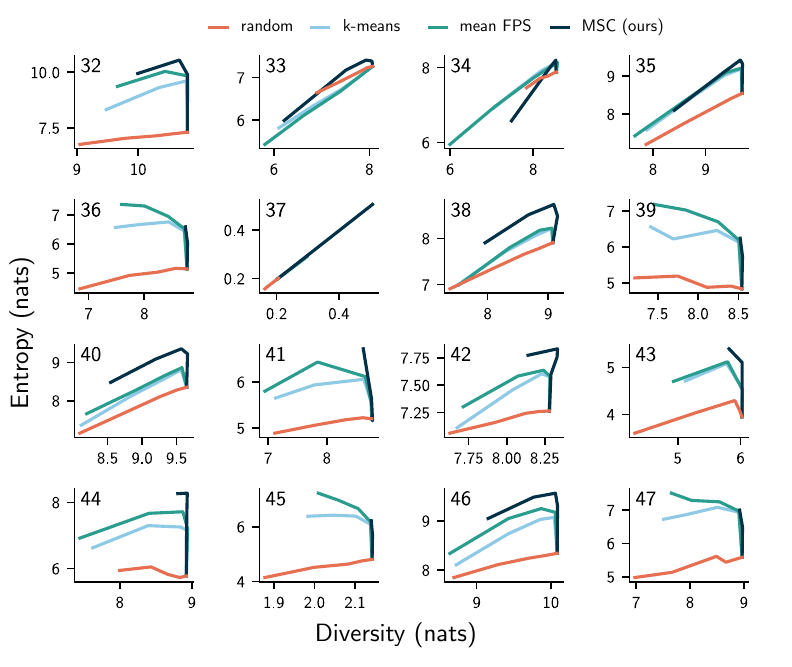}
    \caption{Relationship between entropy and diversity of miscellaneous datasets obtained from the ColabFit repository sampled at different sizes (100\%, 75\%, 50\%, 25\%, 10\%) and with different methods. Outside of a few outliers, datasets compressed with MSC resulted in higher diversity and entropy relative to datasets compressed with other methods, demonstrating that our compression algorithm is the most efficient in removing the redundancy of the data. The dataset numbers, shown on the top left of each plot, correspond to the ones shown in Table \ref{tab:si:colabfit-names}.}
    \label{fig:si:Misc_H_Div_2}
\end{figure}

\begin{figure}[h]  
    \centering
    \includegraphics[width=0.8\textwidth]{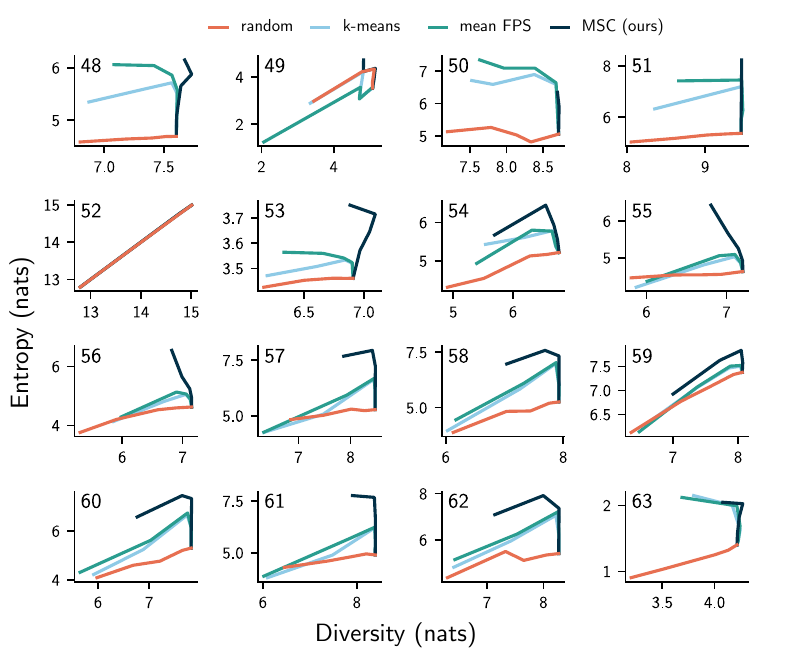}
    \caption{Relationship between entropy and diversity of miscellaneous datasets obtained from the ColabFit repository sampled at different sizes (100\%, 75\%, 50\%, 25\%, 10\%) and with different methods. Outside of a few outliers, datasets compressed with MSC resulted in higher diversity and entropy relative to datasets compressed with other methods, demonstrating that our compression algorithm is the most efficient in removing the redundancy of the data. The dataset numbers, shown on the top left of each plot, correspond to the ones shown in Table \ref{tab:si:colabfit-names}.}
    \label{fig:si:Misc_H_Div_3}
\end{figure}

\begin{figure}[h]  
    \centering
    \includegraphics[width=0.8\textwidth]{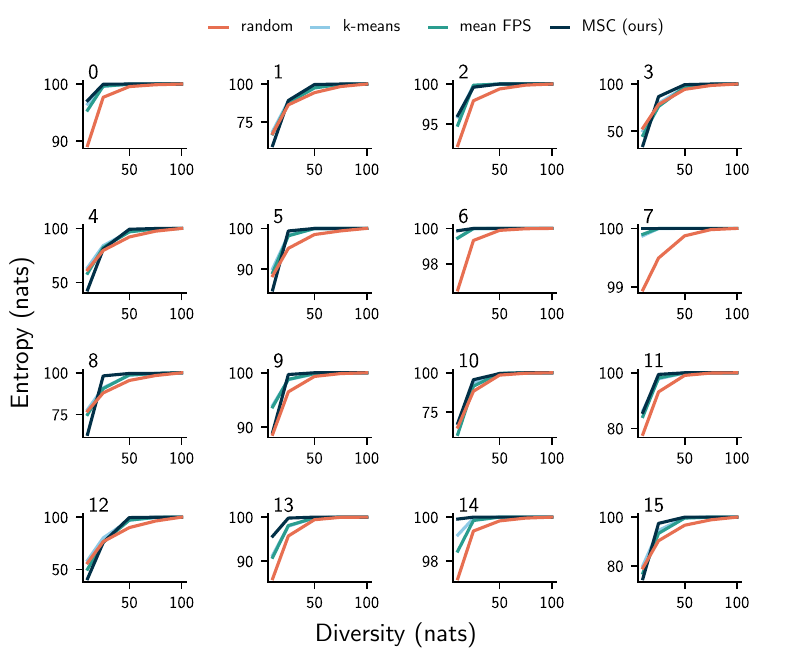}
    \caption{
    Relationship between overlap of several compressed datasets obtained from the ColabFit repository sampled at different sizes (100\%, 75\%, 50\%, 25\%, 10\%) with respect to their own, respective full datasets. Aside from a few outliers, datasets compressed with MSC consistently have higher overlap with the full dataset compared to other algorithms, demonstrating that our algorithm more efficiently preserves the distribution of the dataset. The dataset numbers, shown on the top left of each plot, correspond to the ones shown in Table \ref{tab:si:colabfit-names}.
    }
    \label{fig:si:Misc_Overlap_0}
\end{figure}

\begin{figure}[h]  
    \centering
    \includegraphics[width=0.8\textwidth]{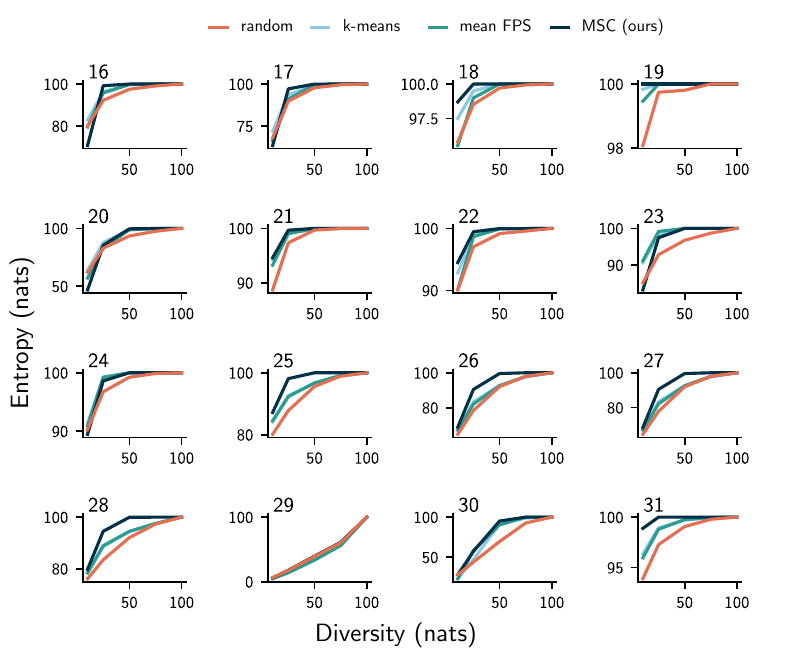}
    \caption{Relationship between overlap of several compressed datasets obtained from the ColabFit repository sampled at different sizes (100\%, 75\%, 50\%, 25\%, 10\%) with respect to their own, respective full datasets. Aside from a few outliers, datasets compressed with MSC consistently have higher overlap with the full dataset compared to other algorithms, demonstrating that our algorithm more efficiently preserves the distribution of the dataset. The dataset numbers, shown on the top left of each plot, correspond to the ones shown in Table \ref{tab:si:colabfit-names}.}
    \label{fig:si:Misc_Overlap_1}
\end{figure}

\begin{figure}[h]  
    \centering
    \includegraphics[width=0.8\textwidth]{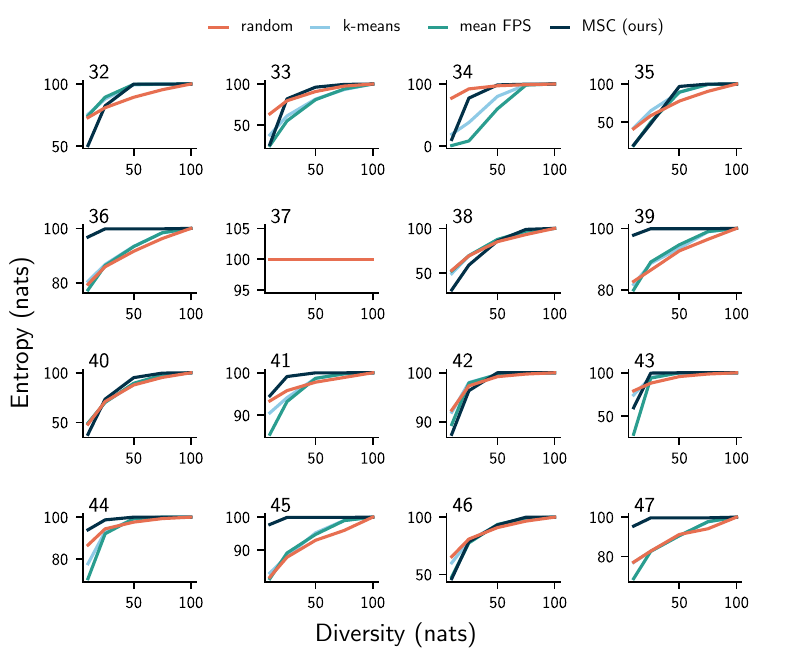}
        \caption{Relationship between overlap of several compressed datasets obtained from the ColabFit repository sampled at different sizes (100\%, 75\%, 50\%, 25\%, 10\%) with respect to their own, respective full datasets. Aside from a few outliers, datasets compressed with MSC consistently have higher overlap with the full dataset compared to other algorithms, demonstrating that our algorithm more efficiently preserves the distribution of the dataset. The dataset numbers, shown on the top left of each plot, correspond to the ones shown in Table \ref{tab:si:colabfit-names}.}
    \label{fig:si:Misc_Overlap_2}
\end{figure}

\begin{figure}[h]  
    \centering
    \includegraphics[width=0.8\textwidth]{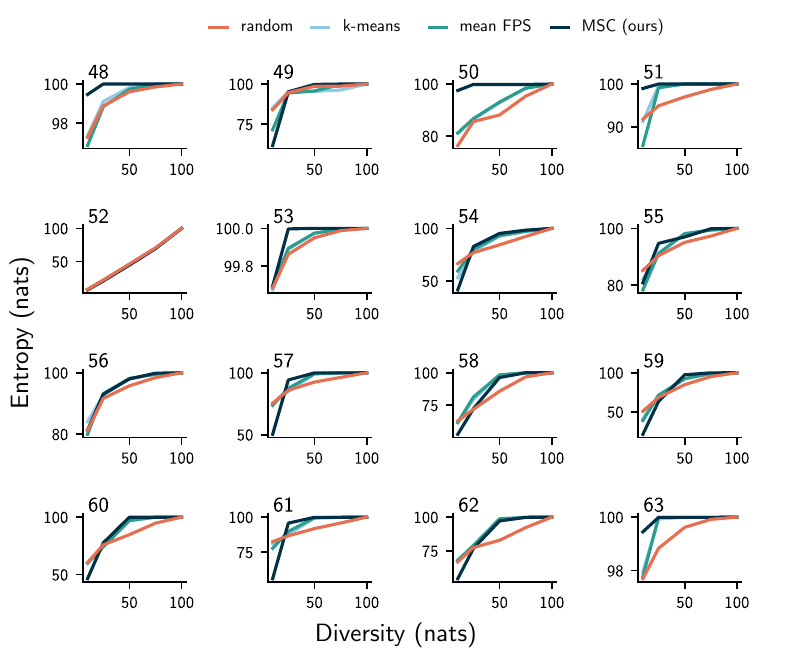}
        \caption{Relationship between overlap of several compressed datasets obtained from the ColabFit repository sampled at different sizes (100\%, 75\%, 50\%, 25\%, 10\%) with respect to their own, respective full datasets. Aside from a few outliers, datasets compressed with MSC consistently have higher overlap with the full dataset compared to other algorithms, demonstrating that our algorithm more efficiently preserves the distribution of the dataset. The dataset numbers, shown on the top left of each plot, correspond to the ones shown in Table \ref{tab:si:colabfit-names}.}
    \label{fig:si:Misc_Overlap_3}
\end{figure}

Beyond the examples above, we evaluated all algorithms in this work on a larger and more diverse collection of datasets used for training MLIPs.
Specifically, we selected 64 single-element datasets from the ColabFit repository\cite{vita2023colabfit} spanning a range of materials, structures, and phases (see Table \ref{tab:si:colabfit-names} for full dataset names).
We compressed each of the datasets by the same ratios adopted for the other examples in this work (10, 25, 50, and 75\%) and compared the resulting performance metrics.
Figure \ref{fig:misc} contrasts the diversity (Fig. \ref{fig:misc}a) and overlap (Fig. \ref{fig:misc}b) of datasets compressed with MSC against the other algorithms.
The negative shifts in distributions show that MSC consistently outperforms all other algorithms across nearly all 64 datasets.
With a few exceptions, the diversity of MSC-compressed datasets is substantially higher than that obtained from other algorithms (see full results in Figs.~\ref{fig:si:Misc_H_Div_0}--\ref{fig:si:Misc_H_Div_3} and Tables~\ref{tab:si:misc-0}--\ref{tab:si:misc-63}), and overlaps can be up to 30\% higher (and 60\% higher in one case) compared to other methods.
Similar to what was observed in the Fullerenes subset of GAP-20 (Fig. \ref{fig:umap}a), variability in overlap is observed when the datasets are compressed to 10\% of their original size even though data diversity is preserved (Figs. \ref{fig:si:Misc_Overlap_0}--\ref{fig:si:Misc_Overlap_3}).
Nevertheless, these results demonstrate that our MSC method is a robust dataset compression technique and often outperforms other algorithms in reducing dataset sizes while also minimizing information loss.

\section{Discussion}

By interpreting the problem of subsampling atomistic datasets as a set cover problem, we demonstrated that a greedy algorithm can be used to obtain a minimum set cover and reduce loss of information of atomistic datasets upon compression.
These results rely on both the information-theoretical formalism used to identify high-novelty data points as well as the avoidance of per-structure representations used in typical baseline algorithms.
In addition to the algorithm, we proposed model-free figures of merit to quantify the efficiency of the compression algorithms across datasets and showed that our method outperforms baseline algorithms in the majority of cases.

Multiple results in this work invite additional discussion.
First, when MLIPs were trained on compressed datasets, their performance was variable, but errors mostly increased compared to training MLIPs to the full dataset, as expected.
Nevertheless, it is possible that subsampling algorithms could lead to different ``learning curves'' for several datasets, decreasing, on average, errors at the low-data regime.
This result, however, may depend on biases on test sets, which are often randomly sampled from the full dataset.
In some cases, specific test samples may contain more or less outliers, increasing the variance in test errors without accounting for the ``fairness'' of the test distribution.
For instance, recent work has showed that variability in dataset construction and model training can lead to robust models even with random sampling of datasets.\cite{stolte2025Random}
While a full analysis of test distributions has been the scope of other studies,\cite{vita2023losslands} this work provides a tool that bypasses the use of test errors as sole figure-of-merit to evaluate data compression.
Moreover, eliminating data redundancies is not only useful to reduce MLIP training costs, but also to augment training sets with new data, as is performed in active learning loops.
For instance, instead of subsampling molecular dynamics trajectories at constant time steps, which may overlook outlier frames, our MSC algorithm can be used to robustly select frames that are most diverse from each other, maximizing the amount of information that can be used to build a new generation of training data points.

At the large-data regime, despite the impressive ability of universal NNIPs to simulate materials across the periodic table, recent works have shown that their degree of generalization (as measured by the $\dH$ of each environment with respect to the training data) is correlated with test errors.\cite{schwalbekoda2025information,chiang2025mliparena}
Given this strong correlation between $\dH$ and errors, ensuring that MLIPs avoid extreme cases of out-of-distribution inference requires minimizing the ``surprise'' of outliers given known training datasets.
We showed that our algorithm reduces the loss of extreme outliers even in cases where information loss is inevitable, such as in the TM23 dataset.
While this work focuses on single-component systems, similar approaches can be performed for multi-component materials, especially if learned representations can be used for each environment.
The definitions of entropy, diversity, and $\dH$ are agnostic to the choice of representation, and $\dH$ was demonstrated to work efficiently with multi-component systems even with the composition-agnostic QUESTS descriptor.\cite{schwalbekoda2025information}

Interestingly, we showed that dimensionality reduction techniques can be insufficient to demonstrate data compression.
While qualitative behavior on dataset coverage can be sometimes extracted from these graphics, we provided examples where outlier loss is far from obvious in a UMAP plot.
On the other hand, our information-theoretical metrics can effectively assess information loss, outlier behavior, and dataset compressibility, allowing a quantitative treatment of dataset subsampling while also being model-free.

One drawback of our MSC algorithm is the increased time complexity to run the algorithm.
While most of the datasets in this work were compressed using a personal laptop in a matter of minutes, datasets with $\mathcal{O}(10^5)-\mathcal{O}(10^6)$ environments may require a few CPU-hours to be compressed using our method.
The software implementation in the QUESTS package provides GPU acceleration, but an information-theoretical approach to data compression will always be more costly than randomly sampling structures.
Some dedicated production applications may strongly benefit from our MSC approach, especially for limiting outlier loss.
Nevertheless, baseline algorithms such as $k$-means sampling are scalable, simple to implement, and remain an improvement over random sampling.

\section{Conclusions}

In summary, we proposed a strategy to subsample datasets for machine learning interatomic potentials with minimal information loss using information theory.
We started with the minimum set cover (MSC) problem definition in combinatorics, which tries to find the minimum number of sets that covers the entire universe of elements belonging to any of these sets, and applied their solutions to atomistic data.
Then, we suggested that selecting a minimum number of structures that maximally cover the space of atom-centered environments is analogous to solving the MSC problem.
Within this formalism, we created an algorithm that avoids information losses when subsampling an atomistic dataset compared to other baseline algorithms.
These results were quantified by multiple figures of merit, including dataset entropy-diversity trade-offs, overlap with the original distribution of data points, and long tail of the force distributions.
When compressed datasets were used to train MLIPs, models trained on MSC-compressed datasets were less prone to ``out-of-distribution'' predictions compared to models trained on datasets compressed with other algorithms.
These results were extensively discussed in the case of GAP-20 and TM23, and remained consistent when a larger analysis was performed for 64 other atomistic datasets, demonstrating the robustness of our MSC method.
All algorithms are implemented in and available as part of the QUESTS package at \url{https://github.com/dskoda/quests} and can be used in a wide range of applications, such as subsampling molecular dynamics trajectories, compressing atomistic datasets for MLIP training, performing active learning loops, and more.

\section*{Data Availability}

The datasets used for training/testing ML potentials were obtained from the original sources at:

\begin{itemize}
    \item GAP-20: \url{https://doi.org/10.17863/CAM.54529}
    \item TM23: \url{https://doi.org/10.24435/materialscloud:6c-b3}
\end{itemize}

The 64 datasets used for the large-scale analysis of the compression algorithms were obtained from the ColabFit repository. The references and DOIs of the datasets are provided in Table \ref{tab:si:colabfit-names} in the Supplementary Information.

The raw data and analysis/plotting code necessary to reproduce all figures and results of this work are available on GitHub at \url{https://github.com/digital-synthesis-lab/2025-compression-data}, with persistent data storage on Zenodo under the DOIs: \href{https://10.5281/zenodo.17536234}{10.5281/zenodo.17536234} and \href{https://doi.org/10.5281/zenodo.17602768}{10.5281/zenodo.17602768}.
Most of the raw data from this work is also available as tables in the the Supplementary Information.

\section*{Code Availability}

The code for QUESTS is available on GitHub at the link \url{https://github.com/dskoda/quests}.
The version of the code used in this work (v.2025.09.29) is deposited on Zenodo for persistent storage under the DOI \href{https://doi.org/10.5281/zenodo.17229448}{10.5281/zenodo.17229448}.

\section*{Acknowledgements}

This work was supported by the U.S. Department of Energy, Office of Science, Office of Basic Energy Sciences under Award Number DE-SC0025642.
V.L.'s contributions were performed under the auspices of the U.S. Department of Energy by Lawrence Livermore National Laboratory under Contract DE-AC52-07NA27344, funded by the Laboratory Directed Research and Development Program at LLNL under project tracking code 23-SI-006.
Manuscript released under the tracking code LLNL-JRNL-2013284-DRAFT.
The authors acknowledge helpful discussions with Joshua Vita.

\section*{Conflicts of Interest}

The authors have no conflicts to disclose.

\section*{Author Contributions}

\noindent\textbf{Benjamin Yu:} Formal Analysis; Investigation; Software; Validation; Visualization; Writing - Original Draft; Writing - Review \& Editing.
\noindent\textbf{Vincenzo Lordi:} Conceptualization; Writing - Review \& Editing; Funding Acquisition.
\noindent\textbf{Daniel Schwalbe-Koda:} Conceptualization; Formal Analysis; Investigation; Methodology; Project Administration; Software; Validation; Visualization; Writing - Original Draft; Writing - Review \& Editing; Funding Acquisition; Supervision.

\clearpage
\beginsupplement

\beginsuppinfo
\customlabel{sec:sinfo}{Supplementary Information}

\section{Supplementary Text}
\customlabel{sec:stext}{Supplementary Text}


\subsection{Example of suboptimal dataset compression with per-structure representations}

Assume that a dataset to be compressed contains two structures representing the same relaxed, bulk crystal with a vacancy, but one of the structures has 255 atoms and the other one has 511 atoms.
These otherwise identical structures contain mostly similar atomic environments, including the few atoms around the vacancy that deviate from bulk-like environments, even though they should exhibit small differences in electronic structures due to the concentration of defects.
As a result, the two per-structure representations --- obtained from the average representations of each cell --- will necessarily be similar because of the large number of atoms of both structures.
According to their per-structure representation, therefore, these two structures are nearly indistinguishable, and may be selected almost interchangeably if subsampled from a diverse dataset.
However, the 511-atom supercell has a significantly higher memory overhead compared to the 255-atom supercell even though both carry the same amount of information.
From a dataset compression perspective, therefore, it is desirable to take into account smaller supercells that cover the same space of atomic environments.

An alternative example regarding the problem above would be to consider a manual selection rule, where the structure containing the smallest number of environments is selected for each neighborhood of the high-dimensional representation space.
In the example above, this rule would lead to the selection of the 255-atom supercell rather than the 511-atom supercell.
However, given this rule, one can also propose a counterexample that leads to outlier loss.
For instance, take a small cell representing the idealized bulk structure rather than a defected cell.
Because of the large size of the supercell and the fact that the defect may be a small fraction of the cell, the per-structure representation of the small cell is likely close to the per-structure representation of the defected cell.
According to the manual selection rule, the information of the defect may be lost simply because it is averaged out in a per-structure representation.

\subsection{Diversity increases upon dataset compression}

Figure \ref{fig:tm23}a of the main text showcases examples where the diversity of a dataset increases upon compression.
However, within a strict definition, the diversity of a dataset should not decrease as the dataset size increases.
The definition of diversity in Eq. \eqref{eq:diversity} was proposed in our previous work to recover two limiting cases:
(1) when the dataset is a degenerate case where all environments are identical, the diversity should be 0 nats; and
(2) when the dataset is a degenerate case where all environments are distinct, the diversity should be equal to $\log n$, where $n$ is the number of environments.
Moreover, the current definition of diversity approximates the non-decreasing behavior of a true diversity metric, but does not guarantee it.
Especially in near-degenerate cases where there is large redundancy in the data, as is the case of the Ag and Ir subsets of TM23, an increase in data redundancy can slightly decrease the dataset diversity, as the support of the data distribution cannot be easily measured in the continuous space of atomic environments.
Therefore, the diversity in Eq. \eqref{eq:diversity} is an approximation of dataset diversity rather than an actual measure of the support of the distribution.
Future work can focus on proposing more rigorous measures of dataset diversity, but the analysis in this and previous work shows that the current definition is sufficient to demonstrate the results of dataset compression.

\section{Supplementary Figures}

\clearpage
\section{Supplementary Tables}

\subsection{Results of dataset compression for GAP-20}

\begin{table*}[ht!]
\centering
\tiny
\caption{Entropy, Diversity, and Overlap results for the Graphene subset of GAP-20}
\label{tab:si:gap-gr-entropy}
\begin{tabular}{l|cccc|cccc|cccc}
\hline
\multicolumn{1}{c}{\textbf{}} & \multicolumn{4}{c}{\textbf{Entropy (nats)}} & \multicolumn{4}{c}{\textbf{Diversity (nats)}} & \multicolumn{4}{c}{\textbf{Overlap (\%)}} \\
\hline
Size (\%) & Random & K-means & Mean FPS & MSC (ours) & Random & K-means & Mean FPS & MSC (ours) & Random & K-means & Mean FPS & MSC (ours) \\
\hline

10 & 4.24 & 4.20 & 4.09 & 5.55 & 5.75 & 5.81 & 5.82 & 6.34 & 0.98 & 0.98 & 0.98 & 0.99 \\
25 & 4.34 & 4.33 & 4.31 & 5.55 & 6.03 & 6.11 & 6.08 & 6.55 & 0.99 & 0.99 & 0.99 & 1.00 \\
50 & 4.29 & 4.52 & 4.52 & 5.37 & 6.22 & 6.32 & 6.32 & 6.54 & 1.00 & 1.00 & 1.00 & 1.00 \\
75 & 4.28 & 4.44 & 4.53 & 4.78 & 6.39 & 6.41 & 6.44 & 6.47 & 1.00 & 1.00 & 1.00 & 1.00 \\
100 & 4.30 & 4.30 & 4.30 & 4.30 & 6.45 & 6.45 & 6.45 & 6.45 & 1.00 & 1.00 & 1.00 & 1.00 \\
\hline
\end{tabular}
\end{table*}

\begin{table*}[h!]
\centering
\tiny
\caption{Entropy, Diversity, and Overlap results for the Nanotubes subset of GAP-20}
\label{tab:si:gap-nt-entropy}
\begin{tabular}{l|cccc|cccc|cccc}
\hline
\multicolumn{1}{c}{\textbf{}} & \multicolumn{4}{c}{\textbf{Entropy (nats)}} & \multicolumn{4}{c}{\textbf{Diversity (nats)}} & \multicolumn{4}{c}{\textbf{Overlap (\%)}} \\
\hline
Size (\%) & Random & K-means & Mean FPS & MSC (ours) & Random & K-means & Mean FPS & MSC (ours) & Random & K-means & Mean FPS & MSC (ours) \\
\hline

10 & 6.54 & 6.66 & 7.00 & 7.86 & 7.25 & 7.30 & 7.40 & 8.09 & 0.78 & 0.73 & 0.61 & 0.80 \\
25 & 6.77 & 7.14 & 7.29 & 7.87 & 7.85 & 7.97 & 7.98 & 8.53 & 0.87 & 0.85 & 0.82 & 0.97 \\
50 & 6.93 & 7.11 & 7.14 & 7.52 & 8.28 & 8.30 & 8.30 & 8.66 & 0.94 & 0.93 & 0.92 & 1.00 \\
75 & 6.92 & 7.09 & 7.14 & 7.26 & 8.46 & 8.52 & 8.52 & 8.65 & 0.97 & 0.97 & 0.97 & 1.00 \\
100 & 7.03 & 7.03 & 7.03 & 7.03 & 8.65 & 8.65 & 8.65 & 8.65 & 1.00 & 1.00 & 1.00 & 1.00 \\
\hline
\end{tabular}
\end{table*}

\begin{table*}[h!]
\centering
\tiny
\caption{Entropy, Diversity, and Overlap results for the Fullerenes subset of GAP-20}
\label{tab:si:gap-fu-entropy}
\begin{tabular}{l|cccc|cccc|cccc}
\hline
\multicolumn{1}{c}{\textbf{}} & \multicolumn{4}{c}{\textbf{Entropy (nats)}} & \multicolumn{4}{c}{\textbf{Diversity (nats)}} & \multicolumn{4}{c}{\textbf{Overlap (\%)}} \\
\hline
Size (\%) & Random & K-means & Mean FPS & MSC (ours) & Random & K-means & Mean FPS & MSC (ours) & Random & K-means & Mean FPS & MSC (ours) \\
\hline

10 & 7.76 & 7.99 & 8.22 & 9.18 & 8.29 & 8.54 & 8.72 & 9.18 & 0.67 & 0.62 & 0.41 & 0.08 \\
25 & 8.12 & 8.92 & 9.22 & 9.76 & 8.88 & 9.45 & 9.62 & 9.88 & 0.80 & 0.82 & 0.79 & 0.74 \\
50 & 8.49 & 9.15 & 9.21 & 9.36 & 9.49 & 9.91 & 9.93 & 10.00 & 0.88 & 0.94 & 0.94 & 0.96 \\
75 & 8.52 & 8.96 & 8.98 & 8.99 & 9.76 & 9.99 & 9.99 & 10.01 & 0.92 & 0.97 & 0.97 & 0.97 \\
100 & 8.60 & 8.60 & 8.60 & 8.60 & 9.99 & 9.99 & 9.99 & 9.99 & 1.00 & 1.00 & 1.00 & 1.00 \\
\hline
\end{tabular}
\end{table*}

\begin{table*}[h!]
\centering
\tiny
\caption{Force errors for a SevenNet model trained on subsampled datasets of the Graphene, Nanotubes, and Fullerenes subsets of GAP-20. The models were tested on the original, full datasets.}
\label{tab:si:gap-errors}
\begin{tabular}{l|cccc|cccc|cccc}
\hline
\multicolumn{1}{c}{\textbf{}} & \multicolumn{4}{c}{\textbf{Graphene Error (eV/Å)}} & \multicolumn{4}{c}{\textbf{Fullerenes Error (eV/Å)}} & \multicolumn{4}{c}{\textbf{Nanotubes Error (eV/Å)}} \\
\hline
Size (\%) & Random & K-means & Mean FPS & MSC (ours) & Random & K-means & Mean FPS & MSC (ours) & Random & K-means & Mean FPS & MSC (ours) \\
\hline

10 & 0.47 & 0.49 & 0.45 & 0.24 & 3.34 & 6.22 & 6.33 & 2.73 & 0.86 & 1.07 & 1.07 & 0.71 \\
25 & 0.23 & 0.26 & 0.29 & 0.27 & 2.99 & 2.96 & 3.08 & 3.13 & 0.70 & 0.72 & 0.80 & 0.70 \\
50 & 0.18 & 0.17 & 0.20 & 0.17 & 2.88 & 2.58 & 3.09 & 2.49 & 0.71 & 0.55 & 0.70 & 0.53 \\
75 & 0.13 & 0.14 & 0.14 & 0.14 & 2.61 & 3.45 & 2.76 & 2.55 & 0.61 & 0.63 & 0.59 & 0.56 \\
100 & 0.11 & 0.11 & 0.11 & 0.11 & 2.90 & 2.90 & 2.90 & 2.90 & 0.69 & 0.69 & 0.69 & 0.69 \\
\hline
\end{tabular}
\end{table*}

\subsection{Results of dataset compression for TM23}

\begin{table*}[h!]
\centering
\tiny
\caption{Entropy, Diversity, and Overlap results for the Ag (warm) subset of TM23}
\label{tab:si:tm23-ag}
\begin{tabular}{l|cccc|cccc|cccc}
\hline
\multicolumn{1}{c}{\textbf{}} & \multicolumn{4}{c}{\textbf{Entropy (nats)}} & \multicolumn{4}{c}{\textbf{Diversity (nats)}} & \multicolumn{4}{c}{\textbf{Overlap (\%)}} \\
\hline
Size (\%) & Random & K-means & Mean FPS & MSC (ours) & Random & K-means & Mean FPS & MSC (ours) & Random & K-means & Mean FPS & MSC (ours) \\
\hline

10 & 3.78 & 3.87 & 3.93 & 4.02 & 4.53 & 4.58 & 4.66 & 4.97 & 0.99 & 0.99 & 0.99 & 1.00 \\
25 & 3.83 & 3.87 & 3.90 & 4.00 & 4.65 & 4.69 & 4.75 & 5.00 & 1.00 & 1.00 & 1.00 & 1.00 \\
50 & 3.83 & 3.87 & 3.88 & 3.95 & 4.73 & 4.77 & 4.77 & 4.94 & 1.00 & 1.00 & 1.00 & 1.00 \\
75 & 3.84 & 3.85 & 3.85 & 3.90 & 4.76 & 4.77 & 4.78 & 4.86 & 1.00 & 1.00 & 1.00 & 1.00 \\
100 & 3.84 & 3.84 & 3.84 & 3.84 & 4.79 & 4.79 & 4.79 & 4.79 & 1.00 & 1.00 & 1.00 & 1.00 \\
\hline
\end{tabular}
\end{table*}

\begin{table*}[h!]
\centering
\tiny
\caption{Entropy, Diversity, and Overlap results for the Au (warm) subset of TM23}
\label{tab:si:tm23-au}
\begin{tabular}{l|cccc|cccc|cccc}
\hline
\multicolumn{1}{c}{\textbf{}} & \multicolumn{4}{c}{\textbf{Entropy (nats)}} & \multicolumn{4}{c}{\textbf{Diversity (nats)}} & \multicolumn{4}{c}{\textbf{Overlap (\%)}} \\
\hline
Size (\%) & Random & K-means & Mean FPS & MSC (ours) & Random & K-means & Mean FPS & MSC (ours) & Random & K-means & Mean FPS & MSC (ours) \\
\hline

10 & 4.91 & 4.90 & 4.99 & 5.14 & 5.21 & 5.24 & 5.31 & 5.53 & 0.96 & 0.96 & 0.96 & 0.96 \\
25 & 4.93 & 5.02 & 5.05 & 5.18 & 5.33 & 5.43 & 5.48 & 5.66 & 0.99 & 0.99 & 0.99 & 0.99 \\
50 & 4.99 & 5.03 & 5.03 & 5.14 & 5.46 & 5.51 & 5.52 & 5.66 & 1.00 & 1.00 & 1.00 & 1.00 \\
75 & 5.00 & 5.02 & 5.02 & 5.08 & 5.50 & 5.53 & 5.53 & 5.61 & 1.00 & 1.00 & 1.00 & 1.00 \\
100 & 5.00 & 5.00 & 5.00 & 5.00 & 5.53 & 5.53 & 5.53 & 5.53 & 1.00 & 1.00 & 1.00 & 1.00 \\
\hline
\end{tabular}
\end{table*}

\begin{table*}[h!]
\centering
\tiny
\caption{Entropy, Diversity, and Overlap results for the Cd (warm) subset of TM23}
\label{tab:si:tm23-cd}
\begin{tabular}{l|cccc|cccc|cccc}
\hline
\multicolumn{1}{c}{\textbf{}} & \multicolumn{4}{c}{\textbf{Entropy (nats)}} & \multicolumn{4}{c}{\textbf{Diversity (nats)}} & \multicolumn{4}{c}{\textbf{Overlap (\%)}} \\
\hline
Size (\%) & Random & K-means & Mean FPS & MSC (ours) & Random & K-means & Mean FPS & MSC (ours) & Random & K-means & Mean FPS & MSC (ours) \\
\hline

10 & 4.67 & 4.67 & 4.73 & 4.83 & 4.99 & 5.00 & 5.09 & 5.28 & 0.99 & 0.99 & 1.00 & 1.00 \\
25 & 4.71 & 4.73 & 4.76 & 4.85 & 5.08 & 5.12 & 5.17 & 5.35 & 1.00 & 1.00 & 1.00 & 1.00 \\
50 & 4.72 & 4.74 & 4.75 & 4.82 & 5.13 & 5.18 & 5.18 & 5.31 & 1.00 & 1.00 & 1.00 & 1.00 \\
75 & 4.73 & 4.74 & 4.74 & 4.78 & 5.16 & 5.18 & 5.18 & 5.24 & 1.00 & 1.00 & 1.00 & 1.00 \\
100 & 4.73 & 4.73 & 4.73 & 4.73 & 5.17 & 5.17 & 5.17 & 5.17 & 1.00 & 1.00 & 1.00 & 1.00 \\
\hline
\end{tabular}
\end{table*}

\begin{table*}[h!]
\centering
\tiny
\caption{Entropy, Diversity, and Overlap results for the Co (warm) subset of TM23}
\label{tab:si:tm23-co}
\begin{tabular}{l|cccc|cccc|cccc}
\hline
\multicolumn{1}{c}{\textbf{}} & \multicolumn{4}{c}{\textbf{Entropy (nats)}} & \multicolumn{4}{c}{\textbf{Diversity (nats)}} & \multicolumn{4}{c}{\textbf{Overlap (\%)}} \\
\hline
Size (\%) & Random & K-means & Mean FPS & MSC (ours) & Random & K-means & Mean FPS & MSC (ours) & Random & K-means & Mean FPS & MSC (ours) \\
\hline

10 & 7.76 & 7.81 & 7.85 & 7.88 & 7.84 & 7.88 & 7.91 & 7.94 & 0.41 & 0.39 & 0.38 & 0.37 \\
25 & 8.27 & 8.32 & 8.34 & 8.37 & 8.43 & 8.47 & 8.49 & 8.52 & 0.70 & 0.70 & 0.70 & 0.70 \\
50 & 8.53 & 8.58 & 8.58 & 8.61 & 8.79 & 8.83 & 8.83 & 8.87 & 0.88 & 0.88 & 0.89 & 0.89 \\
75 & 8.65 & 8.67 & 8.68 & 8.70 & 8.97 & 8.99 & 8.99 & 9.02 & 0.96 & 0.96 & 0.96 & 0.97 \\
100 & 8.72 & 8.72 & 8.72 & 8.72 & 9.09 & 9.09 & 9.09 & 9.09 & 1.00 & 1.00 & 1.00 & 1.00 \\
\hline
\end{tabular}
\end{table*}

\begin{table*}[h!]
\centering
\tiny
\caption{Entropy, Diversity, and Overlap results for the Ir (warm) subset of TM23}
\label{tab:si:tm23-ir}
\begin{tabular}{l|cccc|cccc|cccc}
\hline
\multicolumn{1}{c}{\textbf{}} & \multicolumn{4}{c}{\textbf{Entropy (nats)}} & \multicolumn{4}{c}{\textbf{Diversity (nats)}} & \multicolumn{4}{c}{\textbf{Overlap (\%)}} \\
\hline
Size (\%) & Random & K-means & Mean FPS & MSC (ours) & Random & K-means & Mean FPS & MSC (ours) & Random & K-means & Mean FPS & MSC (ours) \\
\hline

10 & 3.63 & 3.72 & 3.82 & 3.97 & 4.32 & 4.43 & 4.57 & 4.83 & 0.99 & 0.99 & 0.99 & 0.99 \\
25 & 3.64 & 3.75 & 3.78 & 3.92 & 4.43 & 4.59 & 4.60 & 4.86 & 1.00 & 1.00 & 1.00 & 1.00 \\
50 & 3.67 & 3.72 & 3.73 & 3.83 & 4.58 & 4.62 & 4.62 & 4.79 & 1.00 & 1.00 & 1.00 & 1.00 \\
75 & 3.66 & 3.70 & 3.70 & 3.76 & 4.60 & 4.63 & 4.64 & 4.70 & 1.00 & 1.00 & 1.00 & 1.00 \\
100 & 3.66 & 3.66 & 3.66 & 3.66 & 4.62 & 4.62 & 4.62 & 4.62 & 1.00 & 1.00 & 1.00 & 1.00 \\
\hline
\end{tabular}
\end{table*}

\begin{table*}[h!]
\centering
\tiny
\caption{Entropy, Diversity, and Overlap results for the Pd (warm) subset of TM23}
\label{tab:si:tm23-pd}
\begin{tabular}{l|cccc|cccc|cccc}
\hline
\multicolumn{1}{c}{\textbf{}} & \multicolumn{4}{c}{\textbf{Entropy (nats)}} & \multicolumn{4}{c}{\textbf{Diversity (nats)}} & \multicolumn{4}{c}{\textbf{Overlap (\%)}} \\
\hline
Size (\%) & Random & K-means & Mean FPS & MSC (ours) & Random & K-means & Mean FPS & MSC (ours) & Random & K-means & Mean FPS & MSC (ours) \\
\hline

10 & 4.01 & 4.12 & 4.23 & 4.36 & 4.64 & 4.78 & 4.87 & 5.16 & 0.98 & 0.98 & 0.98 & 0.98 \\
25 & 4.07 & 4.18 & 4.23 & 4.37 & 4.84 & 4.97 & 5.00 & 5.25 & 0.99 & 0.99 & 1.00 & 1.00 \\
50 & 4.10 & 4.18 & 4.20 & 4.29 & 4.96 & 5.05 & 5.06 & 5.21 & 1.00 & 1.00 & 1.00 & 1.00 \\
75 & 4.10 & 4.15 & 4.15 & 4.21 & 5.02 & 5.07 & 5.07 & 5.13 & 1.00 & 1.00 & 1.00 & 1.00 \\
100 & 4.10 & 4.10 & 4.10 & 4.10 & 5.05 & 5.05 & 5.05 & 5.05 & 1.00 & 1.00 & 1.00 & 1.00 \\
\hline
\end{tabular}
\end{table*}
\clearpage

\begin{table*}[h!]
\centering
\tiny
\caption{Entropy, Diversity, and Overlap results for the Ti (warm) subset of TM23}
\label{tab:si:tm23-ti}
\begin{tabular}{l|cccc|cccc|cccc}
\hline
\multicolumn{1}{c}{\textbf{}} & \multicolumn{4}{c}{\textbf{Entropy (nats)}} & \multicolumn{4}{c}{\textbf{Diversity (nats)}} & \multicolumn{4}{c}{\textbf{Overlap (\%)}} \\
\hline
Size (\%) & Random & K-means & Mean FPS & MSC (ours) & Random & K-means & Mean FPS & MSC (ours) & Random & K-means & Mean FPS & MSC (ours) \\
\hline

10 & 6.02 & 6.01 & 6.06 & 6.12 & 6.54 & 6.53 & 6.60 & 6.73 & 0.94 & 0.94 & 0.94 & 0.94 \\
25 & 6.09 & 6.11 & 6.15 & 6.20 & 6.78 & 6.80 & 6.85 & 7.01 & 0.98 & 0.98 & 0.98 & 0.98 \\
50 & 6.15 & 6.16 & 6.17 & 6.22 & 6.97 & 6.98 & 6.99 & 7.13 & 0.99 & 0.99 & 0.99 & 1.00 \\
75 & 6.15 & 6.17 & 6.16 & 6.20 & 7.04 & 7.06 & 7.06 & 7.13 & 1.00 & 1.00 & 1.00 & 1.00 \\
100 & 6.16 & 6.16 & 6.16 & 6.16 & 7.09 & 7.09 & 7.09 & 7.09 & 1.00 & 1.00 & 1.00 & 1.00 \\
\hline
\end{tabular}
\end{table*}

\begin{table*}[h!]
\centering
\tiny
\caption{Force errors for a SevenNet model trained on subsampled datasets of the Ag (warm) subset of TM23. The models were tested on the original test splits corresponding to the subset.}
\label{tab:si:tm23-errors-ag}
\begin{tabular}{l|cccc|cccc|cccc}
\hline
\multicolumn{1}{c}{\textbf{}} & \multicolumn{4}{c}{\textbf{Cold Error (eV/Å)}} & \multicolumn{4}{c}{\textbf{Warm Error (eV/Å)}} & \multicolumn{4}{c}{\textbf{Melt Error (eV/Å)}} \\
\hline
Size (\%) & Random & K-means & Mean FPS & MSC (ours) & Random & K-means & Mean FPS & MSC (ours) & Random & K-means & Mean FPS & MSC (ours) \\
\hline

10 & 0.08 & 0.06 & 0.06 & 0.02 & 0.23 & 0.20 & 0.22 & 0.05 & 0.58 & 0.50 & 0.54 & 0.17 \\
25 & 0.03 & 0.03 & 0.03 & 0.02 & 0.10 & 0.12 & 0.10 & 0.10 & 0.31 & 0.35 & 0.31 & 0.29 \\
50 & 0.02 & 0.02 & 0.02 & 0.02 & 0.07 & 0.08 & 0.09 & 0.07 & 0.22 & 0.25 & 0.26 & 0.22 \\
75 & 0.02 & 0.02 & 0.02 & 0.02 & 0.06 & 0.08 & 0.07 & 0.07 & 0.21 & 0.25 & 0.24 & 0.21 \\
100 & 0.01 & 0.01 & 0.01 & 0.01 & 0.07 & 0.07 & 0.07 & 0.07 & 0.21 & 0.21 & 0.21 & 0.21 \\
\hline
\end{tabular}
\end{table*}

\begin{table*}[h!]
\centering
\tiny
\caption{Force errors for a SevenNet model trained on subsampled datasets of the Au (warm) subset of TM23. The models were tested on the original test splits corresponding to the subset.}
\label{tab:si:tm23-errors-au}
\begin{tabular}{l|cccc|cccc|cccc}
\hline
\multicolumn{1}{c}{\textbf{}} & \multicolumn{4}{c}{\textbf{Cold Error (eV/Å)}} & \multicolumn{4}{c}{\textbf{Warm Error (eV/Å)}} & \multicolumn{4}{c}{\textbf{Melt Error (eV/Å)}} \\
\hline
Size (\%) & Random & K-means & Mean FPS & MSC (ours) & Random & K-means & Mean FPS & MSC (ours) & Random & K-means & Mean FPS & MSC (ours) \\
\hline

10 & 0.14 & 0.25 & 0.14 & 0.07 & 0.30 & 0.50 & 0.29 & 0.13 & 0.54 & 0.81 & 0.53 & 0.21 \\
25 & 0.09 & 0.09 & 0.10 & 0.09 & 0.19 & 0.22 & 0.21 & 0.20 & 0.33 & 0.40 & 0.37 & 0.35 \\
50 & 0.07 & 0.07 & 0.08 & 0.07 & 0.16 & 0.15 & 0.17 & 0.15 & 0.28 & 0.26 & 0.30 & 0.25 \\
75 & 0.06 & 0.06 & 0.06 & 0.06 & 0.13 & 0.12 & 0.13 & 0.13 & 0.23 & 0.21 & 0.23 & 0.23 \\
100 & 0.05 & 0.05 & 0.05 & 0.05 & 0.11 & 0.11 & 0.11 & 0.11 & 0.20 & 0.20 & 0.20 & 0.20 \\
\hline
\end{tabular}
\end{table*}

\begin{table*}[h!]
\centering
\tiny
\caption{Force errors for a SevenNet model trained on subsampled datasets of the Cd (warm) subset of TM23. The models were tested on the original test splits corresponding to the subset.}
\label{tab:si:tm23-errors-cd}
\begin{tabular}{l|cccc|cccc|cccc}
\hline
\multicolumn{1}{c}{\textbf{}} & \multicolumn{4}{c}{\textbf{Cold Error (eV/Å)}} & \multicolumn{4}{c}{\textbf{Warm Error (eV/Å)}} & \multicolumn{4}{c}{\textbf{Melt Error (eV/Å)}} \\
\hline
Size (\%) & Random & K-means & Mean FPS & MSC (ours) & Random & K-means & Mean FPS & MSC (ours) & Random & K-means & Mean FPS & MSC (ours) \\
\hline

10 & 0.12 & 0.10 & 0.11 & 0.04 & 0.20 & 0.17 & 0.18 & 0.06 & 0.36 & 0.30 & 0.31 & 0.12 \\
25 & 0.05 & 0.05 & 0.05 & 0.05 & 0.07 & 0.08 & 0.07 & 0.07 & 0.15 & 0.16 & 0.14 & 0.15 \\
50 & 0.04 & 0.04 & 0.04 & 0.04 & 0.07 & 0.06 & 0.07 & 0.07 & 0.14 & 0.12 & 0.14 & 0.14 \\
75 & 0.04 & 0.04 & 0.04 & 0.04 & 0.06 & 0.06 & 0.06 & 0.07 & 0.13 & 0.11 & 0.13 & 0.13 \\
100 & 0.04 & 0.04 & 0.04 & 0.04 & 0.06 & 0.06 & 0.06 & 0.06 & 0.13 & 0.13 & 0.13 & 0.13 \\
\hline
\end{tabular}
\end{table*}

\begin{table*}[h!]
\centering
\tiny
\caption{Force errors for a SevenNet model trained on subsampled datasets of the Co (warm) subset of TM23. The models were tested on the original test splits corresponding to the subset.}
\label{tab:si:tm23-errors-co}
\begin{tabular}{l|cccc|cccc|cccc}
\hline
\multicolumn{1}{c}{\textbf{}} & \multicolumn{4}{c}{\textbf{Cold Error (eV/Å)}} & \multicolumn{4}{c}{\textbf{Warm Error (eV/Å)}} & \multicolumn{4}{c}{\textbf{Melt Error (eV/Å)}} \\
\hline
Size (\%) & Random & K-means & Mean FPS & MSC (ours) & Random & K-means & Mean FPS & MSC (ours) & Random & K-means & Mean FPS & MSC (ours) \\
\hline

10 & 0.32 & 0.29 & 0.31 & 0.18 & 0.64 & 0.63 & 0.61 & 0.30 & 1.22 & 1.26 & 1.12 & 0.61 \\
25 & 0.21 & 0.20 & 0.21 & 0.20 & 0.38 & 0.36 & 0.39 & 0.34 & 0.80 & 0.70 & 0.79 & 0.66 \\
50 & 0.18 & 0.18 & 0.18 & 0.18 & 0.31 & 0.33 & 0.30 & 0.31 & 0.61 & 0.65 & 0.54 & 0.58 \\
75 & 0.18 & 0.18 & 0.18 & 0.18 & 0.29 & 0.28 & 0.29 & 0.28 & 0.56 & 0.51 & 0.59 & 0.49 \\
100 & 0.17 & 0.17 & 0.17 & 0.17 & 0.28 & 0.28 & 0.28 & 0.28 & 0.53 & 0.53 & 0.53 & 0.53 \\
\hline
\end{tabular}
\end{table*}

\begin{table*}[h!]
\centering
\tiny
\caption{Force errors for a SevenNet model trained on subsampled datasets of the Ir (warm) subset of TM23. The models were tested on the original test splits corresponding to the subset.}
\label{tab:si:tm23-errors-ir}
\begin{tabular}{l|cccc|cccc|cccc}
\hline
\multicolumn{1}{c}{\textbf{}} & \multicolumn{4}{c}{\textbf{Cold Error (eV/Å)}} & \multicolumn{4}{c}{\textbf{Warm Error (eV/Å)}} & \multicolumn{4}{c}{\textbf{Melt Error (eV/Å)}} \\
\hline
Size (\%) & Random & K-means & Mean FPS & MSC (ours) & Random & K-means & Mean FPS & MSC (ours) & Random & K-means & Mean FPS & MSC (ours) \\
\hline

10 & 0.41 & 0.42 & 0.34 & 0.20 & 0.82 & 0.84 & 0.63 & 0.34 & 1.54 & 1.46 & 1.18 & 0.64 \\
25 & 0.25 & 0.23 & 0.24 & 0.26 & 0.51 & 0.49 & 0.54 & 0.51 & 0.96 & 0.94 & 1.07 & 1.00 \\
50 & 0.21 & 0.21 & 0.21 & 0.21 & 0.39 & 0.39 & 0.43 & 0.39 & 0.74 & 0.73 & 0.82 & 0.72 \\
75 & 0.20 & 0.20 & 0.20 & 0.20 & 0.36 & 0.37 & 0.40 & 0.40 & 0.63 & 0.68 & 0.76 & 0.80 \\
100 & 0.19 & 0.19 & 0.19 & 0.19 & 0.36 & 0.36 & 0.36 & 0.36 & 0.71 & 0.71 & 0.71 & 0.71 \\
\hline
\end{tabular}
\end{table*}

\begin{table*}[h!]
\centering
\tiny
\caption{Force errors for a SevenNet model trained on subsampled datasets of the Pd (warm) subset of TM23. The models were tested on the original test splits corresponding to the subset.}
\label{tab:si:tm23-errors-pd}
\begin{tabular}{l|cccc|cccc|cccc}
\hline
\multicolumn{1}{c}{\textbf{}} & \multicolumn{4}{c}{\textbf{Cold Error (eV/Å)}} & \multicolumn{4}{c}{\textbf{Warm Error (eV/Å)}} & \multicolumn{4}{c}{\textbf{Melt Error (eV/Å)}} \\
\hline
Size (\%) & Random & K-means & Mean FPS & MSC (ours) & Random & K-means & Mean FPS & MSC (ours) & Random & K-means & Mean FPS & MSC (ours) \\
\hline

10 & 0.18 & 0.21 & 0.20 & 0.07 & 0.52 & 0.54 & 0.50 & 0.17 & 1.14 & 1.18 & 1.15 & 0.48 \\
25 & 0.09 & 0.09 & 0.09 & 0.09 & 0.31 & 0.27 & 0.27 & 0.26 & 0.79 & 0.69 & 0.69 & 0.68 \\
50 & 0.07 & 0.07 & 0.07 & 0.07 & 0.19 & 0.20 & 0.20 & 0.20 & 0.52 & 0.54 & 0.53 & 0.54 \\
75 & 0.06 & 0.07 & 0.07 & 0.07 & 0.16 & 0.15 & 0.17 & 0.15 & 0.44 & 0.42 & 0.47 & 0.44 \\
100 & 0.06 & 0.06 & 0.06 & 0.06 & 0.16 & 0.16 & 0.16 & 0.16 & 0.48 & 0.48 & 0.48 & 0.48 \\
\hline
\end{tabular}
\end{table*}
\clearpage

\begin{table*}[h!]
\centering
\tiny
\caption{Force errors for a SevenNet model trained on subsampled datasets of the Ti (warm) subset of TM23. The models were tested on the original test splits corresponding to the subset.}
\label{tab:si:tm23-errors-ti}
\begin{tabular}{l|cccc|cccc|cccc}
\hline
\multicolumn{1}{c}{\textbf{}} & \multicolumn{4}{c}{\textbf{Cold Error (eV/Å)}} & \multicolumn{4}{c}{\textbf{Warm Error (eV/Å)}} & \multicolumn{4}{c}{\textbf{Melt Error (eV/Å)}} \\
\hline
Size (\%) & Random & K-means & Mean FPS & MSC (ours) & Random & K-means & Mean FPS & MSC (ours) & Random & K-means & Mean FPS & MSC (ours) \\
\hline

10 & 0.36 & 0.39 & 0.37 & 0.26 & 0.55 & 0.64 & 0.59 & 0.37 & 1.22 & 1.28 & 1.27 & 0.93 \\
25 & 0.31 & 0.31 & 0.31 & 0.30 & 0.46 & 0.44 & 0.46 & 0.44 & 1.10 & 1.03 & 1.08 & 1.00 \\
50 & 0.27 & 0.28 & 0.27 & 0.26 & 0.41 & 0.40 & 0.39 & 0.40 & 1.08 & 1.00 & 0.92 & 1.00 \\
75 & 0.25 & 0.25 & 0.25 & 0.25 & 0.39 & 0.39 & 0.42 & 0.39 & 1.01 & 1.10 & 1.12 & 0.98 \\
100 & 0.24 & 0.24 & 0.24 & 0.24 & 0.37 & 0.37 & 0.37 & 0.37 & 0.88 & 0.88 & 0.88 & 0.88 \\
\hline
\end{tabular}
\end{table*}

\subsection{Results of dataset compression for 64 different datasets obtained from ColabFit}

{\small
\begin{longtable}{c|p{10cm}|p{3cm}}
\caption{Names of the 64 single-component datasets obtained from the ColabFit repository. 
The dataset numbers provided here are used in Figs.~\ref{fig:si:Misc_H_Div_0}--\ref{fig:si:Misc_Overlap_3} for brevity.}
\label{tab:si:colabfit-names} \\

\hline
\textbf{Number} & \textbf{Name in ColabFit} & \textbf{DOI} \\
\hline
\endfirsthead

\multicolumn{3}{c}{{\bfseries Table \thetable\ (continued)}} \\
\hline
\textbf{Number} & \textbf{Name in ColabFit} & \textbf{DOI} \\
\hline
\endhead

\hline
\multicolumn{3}{r}{{Continued on next page}} \\
\endfoot

\hline
\endlastfoot

0  & \texttt{23-Single-Element-DNPs RSCDD 2023-Ag} &
\href{https://doi.org/10.60732/7cf58a2f}{10.60732/7cf58a2f}\,\citeSupp{https://doi.org/10.60732/7cf58a2f} \\

1  & \texttt{23-Single-Element-DNPs RSCDD 2023-Al} &
\href{https://doi.org/10.60732/ef6f5966}{10.60732/ef6f5966}\,\citeSupp{https://doi.org/10.60732/ef6f5966} \\

2  & \texttt{23-Single-Element-DNPs RSCDD 2023-Au} &
\href{https://doi.org/10.60732/7c0dfbc8}{10.60732/7c0dfbc8}\,\citeSupp{https://doi.org/10.60732/7c0dfbc8} \\

3  & \texttt{23-Single-Element-DNPs RSCDD 2023-Co} &
\href{https://doi.org/10.60732/15bb1dca}{10.60732/15bb1dca}\,\citeSupp{https://doi.org/10.60732/15bb1dca} \\

4  & \texttt{23-Single-Element-DNPs RSCDD 2023-Cu} &
\href{https://doi.org/10.60732/e0a72dd8}{10.60732/e0a72dd8}\,\citeSupp{https://doi.org/10.60732/e0a72dd8} \\

5  & \texttt{23-Single-Element-DNPs RSCDD 2023-Ge} &
\href{https://doi.org/10.60732/33767c4f}{10.60732/33767c4f}\,\citeSupp{https://doi.org/10.60732/33767c4f} \\

6  & \texttt{23-Single-Element-DNPs RSCDD 2023-I} &
\href{https://doi.org/10.60732/57c54149}{10.60732/57c54149}\,\citeSupp{https://doi.org/10.60732/57c54149} \\

7  & \texttt{23-Single-Element-DNPs RSCDD 2023-Kr} &
\href{https://doi.org/10.60732/6a060a77}{
10.60732/6a060a77}\,\citeSupp{https://doi.org/10.60732/6a060a77} \\
\end{longtable}
}

\begin{table*}[h!]
\centering
\tiny
\caption{Entropy, Diversity, and Overlap results for dataset 23-Single-Element-DNPs RSCDD 2023-Ag}
\label{tab:si:misc-0}
\begin{tabular}{l|cccc|cccc|cccc}
\hline
\multicolumn{1}{c}{\textbf{}} & \multicolumn{4}{c}{\textbf{Entropy (nats)}} & \multicolumn{4}{c}{\textbf{Diversity (nats)}} & \multicolumn{4}{c}{\textbf{Overlap (\%)}} \\
\hline
Size (\%) & Random & K-means & Mean FPS & MSC (ours) & Random & K-means & Mean FPS & MSC (ours) & Random & K-means & Mean FPS & MSC (ours) \\
\hline

10.0 & 6.04 & 6.65 & 6.71 & 7.25 & 7.19 & 7.56 & 7.57 & 7.77 & 0.89 & 0.96 & 0.95 & 0.97 \\
25.0 & 6.19 & 6.54 & 6.65 & 7.34 & 7.52 & 7.69 & 7.73 & 7.78 & 0.98 & 1.00 & 1.00 & 1.00 \\
50.0 & 6.23 & 6.45 & 6.51 & 7.14 & 7.66 & 7.71 & 7.73 & 7.76 & 1.00 & 1.00 & 1.00 & 1.00 \\
75.0 & 6.28 & 6.39 & 6.39 & 6.74 & 7.70 & 7.71 & 7.72 & 7.72 & 1.00 & 1.00 & 1.00 & 1.00 \\
100.0 & 6.26 & 6.26 & 6.26 & 6.26 & 7.71 & 7.71 & 7.71 & 7.71 & 1.00 & 1.00 & 1.00 & 1.00 \\
\hline
\end{tabular}
\end{table*}

\begin{table*}[h!]
\centering
\tiny
\caption{Entropy, Diversity, and Overlap results for dataset 23-Single-Element-DNPs RSCDD 2023-Al}
\label{tab:si:misc-1}
\begin{tabular}{l|cccc|cccc|cccc}
\hline
\multicolumn{1}{c}{\textbf{}} & \multicolumn{4}{c}{\textbf{Entropy (nats)}} & \multicolumn{4}{c}{\textbf{Diversity (nats)}} & \multicolumn{4}{c}{\textbf{Overlap (\%)}} \\
\hline
Size (\%) & Random & K-means & Mean FPS & MSC (ours) & Random & K-means & Mean FPS & MSC (ours) & Random & K-means & Mean FPS & MSC (ours) \\
\hline

10.0 & 7.34 & 7.58 & 7.68 & 7.96 & 7.91 & 8.09 & 8.13 & 8.45 & 0.67 & 0.69 & 0.67 & 0.59 \\
25.0 & 7.79 & 8.07 & 8.17 & 8.49 & 8.56 & 8.70 & 8.72 & 9.00 & 0.86 & 0.89 & 0.88 & 0.89 \\
50.0 & 7.94 & 8.33 & 8.38 & 8.51 & 8.87 & 9.04 & 9.06 & 9.15 & 0.94 & 0.98 & 0.97 & 1.00 \\
75.0 & 7.98 & 8.21 & 8.21 & 8.29 & 9.03 & 9.13 & 9.13 & 9.14 & 0.98 & 1.00 & 1.00 & 1.00 \\
100.0 & 8.02 & 8.02 & 8.02 & 8.02 & 9.12 & 9.12 & 9.12 & 9.12 & 1.00 & 1.00 & 1.00 & 1.00 \\
\hline
\end{tabular}
\end{table*}

\begin{table*}[h!]
\centering
\tiny
\caption{Entropy, Diversity, and Overlap results for dataset 23-Single-Element-DNPs RSCDD 2023-Au}
\label{tab:si:misc-2}
\begin{tabular}{l|cccc|cccc|cccc}
\hline
\multicolumn{1}{c}{\textbf{}} & \multicolumn{4}{c}{\textbf{Entropy (nats)}} & \multicolumn{4}{c}{\textbf{Diversity (nats)}} & \multicolumn{4}{c}{\textbf{Overlap (\%)}} \\
\hline
Size (\%) & Random & K-means & Mean FPS & MSC (ours) & Random & K-means & Mean FPS & MSC (ours) & Random & K-means & Mean FPS & MSC (ours) \\
\hline

10.0 & 6.56 & 7.16 & 7.41 & 7.81 & 7.26 & 7.68 & 7.84 & 8.02 & 0.92 & 0.96 & 0.95 & 0.96 \\
25.0 & 6.80 & 7.24 & 7.26 & 7.71 & 7.58 & 7.96 & 8.00 & 8.01 & 0.98 & 1.00 & 1.00 & 1.00 \\
50.0 & 6.92 & 7.11 & 7.13 & 7.55 & 7.77 & 7.93 & 7.94 & 7.95 & 0.99 & 1.00 & 1.00 & 1.00 \\
75.0 & 6.94 & 7.07 & 7.07 & 7.34 & 7.82 & 7.90 & 7.91 & 7.90 & 1.00 & 1.00 & 1.00 & 1.00 \\
100.0 & 6.96 & 6.96 & 6.96 & 6.96 & 7.88 & 7.88 & 7.88 & 7.88 & 1.00 & 1.00 & 1.00 & 1.00 \\
\hline
\end{tabular}
\end{table*}

\begin{table*}[h!]
\centering
\tiny
\caption{Entropy, Diversity, and Overlap results for dataset 23-Single-Element-DNPs RSCDD 2023-Co}
\label{tab:si:misc-3}
\begin{tabular}{l|cccc|cccc|cccc}
\hline
\multicolumn{1}{c}{\textbf{}} & \multicolumn{4}{c}{\textbf{Entropy (nats)}} & \multicolumn{4}{c}{\textbf{Diversity (nats)}} & \multicolumn{4}{c}{\textbf{Overlap (\%)}} \\
\hline
Size (\%) & Random & K-means & Mean FPS & MSC (ours) & Random & K-means & Mean FPS & MSC (ours) & Random & K-means & Mean FPS & MSC (ours) \\
\hline

10.0 & 7.53 & 7.86 & 7.80 & 8.17 & 7.98 & 8.22 & 8.18 & 8.46 & 0.53 & 0.50 & 0.46 & 0.35 \\
25.0 & 8.00 & 8.34 & 8.37 & 8.78 & 8.61 & 8.88 & 8.90 & 9.10 & 0.78 & 0.80 & 0.76 & 0.87 \\
50.0 & 8.20 & 8.36 & 8.40 & 8.84 & 8.92 & 9.10 & 9.13 & 9.16 & 0.94 & 0.97 & 0.97 & 0.99 \\
75.0 & 8.29 & 8.39 & 8.39 & 8.76 & 9.07 & 9.13 & 9.15 & 9.16 & 0.98 & 0.99 & 1.00 & 1.00 \\
100.0 & 8.35 & 8.35 & 8.35 & 8.35 & 9.14 & 9.14 & 9.14 & 9.14 & 1.00 & 1.00 & 1.00 & 1.00 \\
\hline
\end{tabular}
\end{table*}

\begin{table*}[h!]
\centering
\tiny
\caption{Entropy, Diversity, and Overlap results for dataset 23-Single-Element-DNPs RSCDD 2023-Cu}
\label{tab:si:misc-4}
\begin{tabular}{l|cccc|cccc|cccc}
\hline
\multicolumn{1}{c}{\textbf{}} & \multicolumn{4}{c}{\textbf{Entropy (nats)}} & \multicolumn{4}{c}{\textbf{Diversity (nats)}} & \multicolumn{4}{c}{\textbf{Overlap (\%)}} \\
\hline
Size (\%) & Random & K-means & Mean FPS & MSC (ours) & Random & K-means & Mean FPS & MSC (ours) & Random & K-means & Mean FPS & MSC (ours) \\
\hline

10.0 & 7.72 & 8.21 & 8.28 & 8.51 & 8.28 & 8.53 & 8.61 & 8.86 & 0.62 & 0.64 & 0.58 & 0.43 \\
25.0 & 8.08 & 8.72 & 8.87 & 9.11 & 8.85 & 9.20 & 9.27 & 9.49 & 0.80 & 0.84 & 0.82 & 0.81 \\
50.0 & 8.30 & 8.92 & 8.96 & 9.13 & 9.27 & 9.57 & 9.58 & 9.63 & 0.92 & 0.97 & 0.97 & 0.99 \\
75.0 & 8.41 & 8.61 & 8.61 & 8.92 & 9.48 & 9.61 & 9.62 & 9.61 & 0.97 & 1.00 & 1.00 & 1.00 \\
100.0 & 8.43 & 8.43 & 8.43 & 8.43 & 9.59 & 9.59 & 9.59 & 9.59 & 1.00 & 1.00 & 1.00 & 1.00 \\
\hline
\end{tabular}
\end{table*}

\begin{table*}[h!]
\centering
\tiny
\caption{Entropy, Diversity, and Overlap results for dataset 23-Single-Element-DNPs RSCDD 2023-Ge}
\label{tab:si:misc-5}
\begin{tabular}{l|cccc|cccc|cccc}
\hline
\multicolumn{1}{c}{\textbf{}} & \multicolumn{4}{c}{\textbf{Entropy (nats)}} & \multicolumn{4}{c}{\textbf{Diversity (nats)}} & \multicolumn{4}{c}{\textbf{Overlap (\%)}} \\
\hline
Size (\%) & Random & K-means & Mean FPS & MSC (ours) & Random & K-means & Mean FPS & MSC (ours) & Random & K-means & Mean FPS & MSC (ours) \\
\hline

10.0 & 6.79 & 7.29 & 7.53 & 8.23 & 8.05 & 8.26 & 8.33 & 8.83 & 0.88 & 0.90 & 0.89 & 0.85 \\
25.0 & 6.75 & 7.81 & 8.02 & 8.35 & 8.45 & 8.86 & 8.93 & 9.05 & 0.95 & 0.98 & 0.98 & 0.99 \\
50.0 & 6.92 & 7.65 & 7.72 & 7.93 & 8.75 & 9.03 & 9.03 & 9.01 & 0.98 & 1.00 & 1.00 & 1.00 \\
75.0 & 6.88 & 7.38 & 7.38 & 7.50 & 8.88 & 8.99 & 8.99 & 8.98 & 0.99 & 1.00 & 1.00 & 1.00 \\
100.0 & 6.90 & 6.90 & 6.90 & 6.90 & 8.97 & 8.97 & 8.97 & 8.97 & 1.00 & 1.00 & 1.00 & 1.00 \\
\hline
\end{tabular}
\end{table*}
\clearpage

\begin{table*}[h!]
\centering
\tiny
\caption{Entropy, Diversity, and Overlap results for dataset 23-Single-Element-DNPs RSCDD 2023-I}
\label{tab:si:misc-6}
\begin{tabular}{l|cccc|cccc|cccc}
\hline
\multicolumn{1}{c}{\textbf{}} & \multicolumn{4}{c}{\textbf{Entropy (nats)}} & \multicolumn{4}{c}{\textbf{Diversity (nats)}} & \multicolumn{4}{c}{\textbf{Overlap (\%)}} \\
\hline
Size (\%) & Random & K-means & Mean FPS & MSC (ours) & Random & K-means & Mean FPS & MSC (ours) & Random & K-means & Mean FPS & MSC (ours) \\
\hline

10.0 & 4.56 & 5.59 & 5.79 & 6.25 & 6.18 & 6.51 & 6.60 & 6.70 & 0.97 & 0.99 & 0.99 & 1.00 \\
25.0 & 4.73 & 5.58 & 5.64 & 6.11 & 6.43 & 6.66 & 6.67 & 6.70 & 0.99 & 1.00 & 1.00 & 1.00 \\
50.0 & 4.74 & 5.26 & 5.29 & 5.71 & 6.51 & 6.63 & 6.63 & 6.65 & 1.00 & 1.00 & 1.00 & 1.00 \\
75.0 & 4.73 & 4.97 & 5.01 & 5.25 & 6.57 & 6.61 & 6.61 & 6.62 & 1.00 & 1.00 & 1.00 & 1.00 \\
100.0 & 4.76 & 4.76 & 4.76 & 4.76 & 6.61 & 6.61 & 6.61 & 6.61 & 1.00 & 1.00 & 1.00 & 1.00 \\
\hline
\end{tabular}
\end{table*}

\begin{table*}[h!]
\centering
\tiny
\caption{Entropy, Diversity, and Overlap results for dataset 23-Single-Element-DNPs RSCDD 2023-Kr}
\label{tab:si:misc-7}
\begin{tabular}{l|cccc|cccc|cccc}
\hline
\multicolumn{1}{c}{\textbf{}} & \multicolumn{4}{c}{\textbf{Entropy (nats)}} & \multicolumn{4}{c}{\textbf{Diversity (nats)}} & \multicolumn{4}{c}{\textbf{Overlap (\%)}} \\
\hline
Size (\%) & Random & K-means & Mean FPS & MSC (ours) & Random & K-means & Mean FPS & MSC (ours) & Random & K-means & Mean FPS & MSC (ours) \\
\hline

10.0 & 3.15 & 4.54 & 4.70 & 5.10 & 1.65 & 1.72 & 1.73 & 1.74 & 0.99 & 1.00 & 1.00 & 1.00 \\
25.0 & 3.28 & 4.38 & 4.49 & 4.86 & 1.69 & 1.73 & 1.73 & 1.74 & 0.99 & 1.00 & 1.00 & 1.00 \\
50.0 & 3.32 & 4.05 & 4.07 & 4.55 & 1.71 & 1.73 & 1.73 & 1.73 & 1.00 & 1.00 & 1.00 & 1.00 \\
75.0 & 3.31 & 3.49 & 3.49 & 3.99 & 1.73 & 1.73 & 1.73 & 1.73 & 1.00 & 1.00 & 1.00 & 1.00 \\
100.0 & 3.32 & 3.32 & 3.32 & 3.32 & 1.73 & 1.73 & 1.73 & 1.73 & 1.00 & 1.00 & 1.00 & 1.00 \\
\hline
\end{tabular}
\end{table*}

\begin{table*}[h!]
\centering
\tiny
\caption{Entropy, Diversity, and Overlap results for dataset 23-Single-Element-DNPs RSCDD 2023-Li}
\label{tab:si:misc-8}
\begin{tabular}{l|cccc|cccc|cccc}
\hline
\multicolumn{1}{c}{\textbf{}} & \multicolumn{4}{c}{\textbf{Entropy (nats)}} & \multicolumn{4}{c}{\textbf{Diversity (nats)}} & \multicolumn{4}{c}{\textbf{Overlap (\%)}} \\
\hline
Size (\%) & Random & K-means & Mean FPS & MSC (ours) & Random & K-means & Mean FPS & MSC (ours) & Random & K-means & Mean FPS & MSC (ours) \\
\hline

10.0 & 7.15 & 7.73 & 7.95 & 8.56 & 7.96 & 8.28 & 8.37 & 8.75 & 0.77 & 0.78 & 0.75 & 0.63 \\
25.0 & 7.25 & 8.05 & 8.18 & 8.73 & 8.40 & 8.87 & 8.92 & 9.10 & 0.88 & 0.91 & 0.91 & 0.98 \\
50.0 & 7.44 & 7.98 & 8.01 & 8.42 & 8.79 & 9.06 & 9.08 & 9.11 & 0.95 & 0.99 & 0.99 & 1.00 \\
75.0 & 7.48 & 7.84 & 7.86 & 8.10 & 8.96 & 9.08 & 9.09 & 9.10 & 0.98 & 1.00 & 1.00 & 1.00 \\
100.0 & 7.53 & 7.53 & 7.53 & 7.53 & 9.09 & 9.09 & 9.09 & 9.09 & 1.00 & 1.00 & 1.00 & 1.00 \\
\hline
\end{tabular}
\end{table*}

\begin{table*}[h!]
\centering
\tiny
\caption{Entropy, Diversity, and Overlap results for dataset 23-Single-Element-DNPs RSCDD 2023-Mg}
\label{tab:si:misc-9}
\begin{tabular}{l|cccc|cccc|cccc}
\hline
\multicolumn{1}{c}{\textbf{}} & \multicolumn{4}{c}{\textbf{Entropy (nats)}} & \multicolumn{4}{c}{\textbf{Diversity (nats)}} & \multicolumn{4}{c}{\textbf{Overlap (\%)}} \\
\hline
Size (\%) & Random & K-means & Mean FPS & MSC (ours) & Random & K-means & Mean FPS & MSC (ours) & Random & K-means & Mean FPS & MSC (ours) \\
\hline

10.0 & 5.71 & 6.24 & 6.43 & 6.44 & 6.77 & 7.01 & 7.05 & 7.30 & 0.89 & 0.94 & 0.94 & 0.89 \\
25.0 & 5.88 & 6.33 & 6.48 & 6.58 & 7.11 & 7.29 & 7.33 & 7.45 & 0.97 & 0.99 & 0.99 & 1.00 \\
50.0 & 5.97 & 6.30 & 6.35 & 6.67 & 7.30 & 7.38 & 7.40 & 7.45 & 0.99 & 1.00 & 1.00 & 1.00 \\
75.0 & 5.99 & 6.19 & 6.19 & 6.47 & 7.37 & 7.41 & 7.41 & 7.42 & 1.00 & 1.00 & 1.00 & 1.00 \\
100.0 & 5.99 & 5.99 & 5.99 & 5.99 & 7.41 & 7.41 & 7.41 & 7.41 & 1.00 & 1.00 & 1.00 & 1.00 \\
\hline
\end{tabular}
\end{table*}

\begin{table*}[h!]
\centering
\tiny
\caption{Entropy, Diversity, and Overlap results for dataset 23-Single-Element-DNPs RSCDD 2023-Mo}
\label{tab:si:misc-10}
\begin{tabular}{l|cccc|cccc|cccc}
\hline
\multicolumn{1}{c}{\textbf{}} & \multicolumn{4}{c}{\textbf{Entropy (nats)}} & \multicolumn{4}{c}{\textbf{Diversity (nats)}} & \multicolumn{4}{c}{\textbf{Overlap (\%)}} \\
\hline
Size (\%) & Random & K-means & Mean FPS & MSC (ours) & Random & K-means & Mean FPS & MSC (ours) & Random & K-means & Mean FPS & MSC (ours) \\
\hline

10.0 & 7.16 & 7.50 & 7.44 & 8.11 & 7.76 & 7.95 & 7.89 & 8.32 & 0.65 & 0.68 & 0.61 & 0.68 \\
25.0 & 7.59 & 7.82 & 7.86 & 8.38 & 8.25 & 8.45 & 8.49 & 8.57 & 0.88 & 0.94 & 0.91 & 0.96 \\
50.0 & 7.70 & 7.81 & 7.88 & 8.37 & 8.47 & 8.55 & 8.58 & 8.60 & 0.99 & 0.99 & 1.00 & 0.99 \\
75.0 & 7.71 & 7.74 & 7.75 & 8.22 & 8.54 & 8.56 & 8.57 & 8.59 & 1.00 & 1.00 & 1.00 & 1.00 \\
100.0 & 7.74 & 7.74 & 7.74 & 7.74 & 8.58 & 8.58 & 8.58 & 8.58 & 1.00 & 1.00 & 1.00 & 1.00 \\
\hline
\end{tabular}
\end{table*}

\begin{table*}[h!]
\centering
\tiny
\caption{Entropy, Diversity, and Overlap results for dataset 23-Single-Element-DNPs RSCDD 2023-Nb}
\label{tab:si:misc-11}
\begin{tabular}{l|cccc|cccc|cccc}
\hline
\multicolumn{1}{c}{\textbf{}} & \multicolumn{4}{c}{\textbf{Entropy (nats)}} & \multicolumn{4}{c}{\textbf{Diversity (nats)}} & \multicolumn{4}{c}{\textbf{Overlap (\%)}} \\
\hline
Size (\%) & Random & K-means & Mean FPS & MSC (ours) & Random & K-means & Mean FPS & MSC (ours) & Random & K-means & Mean FPS & MSC (ours) \\
\hline

10.0 & 6.62 & 7.07 & 7.18 & 7.56 & 1.98 & 2.02 & 2.03 & 2.06 & 0.78 & 0.86 & 0.84 & 0.86 \\
25.0 & 6.80 & 7.22 & 7.36 & 7.73 & 2.03 & 2.06 & 2.07 & 2.08 & 0.93 & 0.98 & 0.98 & 0.99 \\
50.0 & 6.92 & 7.19 & 7.25 & 7.62 & 2.06 & 2.07 & 2.07 & 2.08 & 0.99 & 1.00 & 1.00 & 1.00 \\
75.0 & 6.91 & 7.12 & 7.12 & 7.32 & 2.07 & 2.07 & 2.07 & 2.07 & 1.00 & 1.00 & 1.00 & 1.00 \\
100.0 & 6.90 & 6.90 & 6.90 & 6.90 & 2.07 & 2.07 & 2.07 & 2.07 & 1.00 & 1.00 & 1.00 & 1.00 \\
\hline
\end{tabular}
\end{table*}
\clearpage

\begin{table*}[h!]
\centering
\tiny
\caption{Entropy, Diversity, and Overlap results for dataset 23-Single-Element-DNPs RSCDD 2023-Ni}
\label{tab:si:misc-12}
\begin{tabular}{l|cccc|cccc|cccc}
\hline
\multicolumn{1}{c}{\textbf{}} & \multicolumn{4}{c}{\textbf{Entropy (nats)}} & \multicolumn{4}{c}{\textbf{Diversity (nats)}} & \multicolumn{4}{c}{\textbf{Overlap (\%)}} \\
\hline
Size (\%) & Random & K-means & Mean FPS & MSC (ours) & Random & K-means & Mean FPS & MSC (ours) & Random & K-means & Mean FPS & MSC (ours) \\
\hline

10.0 & 7.49 & 7.75 & 7.86 & 8.39 & 8.08 & 8.18 & 8.23 & 8.65 & 0.56 & 0.59 & 0.50 & 0.41 \\
25.0 & 7.94 & 8.57 & 8.79 & 9.23 & 8.75 & 9.09 & 9.19 & 9.47 & 0.77 & 0.80 & 0.76 & 0.76 \\
50.0 & 8.17 & 8.84 & 8.90 & 9.16 & 9.14 & 9.54 & 9.57 & 9.57 & 0.90 & 0.97 & 0.97 & 1.00 \\
75.0 & 8.29 & 8.62 & 8.65 & 8.85 & 9.39 & 9.55 & 9.56 & 9.55 & 0.96 & 1.00 & 1.00 & 1.00 \\
100.0 & 8.35 & 8.35 & 8.35 & 8.35 & 9.54 & 9.54 & 9.54 & 9.54 & 1.00 & 1.00 & 1.00 & 1.00 \\
\hline
\end{tabular}
\end{table*}

\begin{table*}[h!]
\centering
\tiny
\caption{Entropy, Diversity, and Overlap results for dataset 23-Single-Element-DNPs RSCDD 2023-Os}
\label{tab:si:misc-13}
\begin{tabular}{l|cccc|cccc|cccc}
\hline
\multicolumn{1}{c}{\textbf{}} & \multicolumn{4}{c}{\textbf{Entropy (nats)}} & \multicolumn{4}{c}{\textbf{Diversity (nats)}} & \multicolumn{4}{c}{\textbf{Overlap (\%)}} \\
\hline
Size (\%) & Random & K-means & Mean FPS & MSC (ours) & Random & K-means & Mean FPS & MSC (ours) & Random & K-means & Mean FPS & MSC (ours) \\
\hline

10.0 & 6.47 & 7.06 & 7.23 & 7.74 & 7.58 & 7.91 & 7.99 & 8.24 & 0.86 & 0.91 & 0.91 & 0.96 \\
25.0 & 6.72 & 7.05 & 7.15 & 7.88 & 7.99 & 8.15 & 8.19 & 8.28 & 0.96 & 0.98 & 0.98 & 1.00 \\
50.0 & 6.72 & 6.95 & 6.98 & 7.63 & 8.13 & 8.20 & 8.23 & 8.26 & 0.99 & 1.00 & 1.00 & 1.00 \\
75.0 & 6.73 & 6.90 & 6.92 & 7.31 & 8.19 & 8.21 & 8.22 & 8.23 & 1.00 & 1.00 & 1.00 & 1.00 \\
100.0 & 6.73 & 6.73 & 6.73 & 6.73 & 8.22 & 8.22 & 8.22 & 8.22 & 1.00 & 1.00 & 1.00 & 1.00 \\
\hline
\end{tabular}
\end{table*}

\begin{table*}[h!]
\centering
\tiny
\caption{Entropy, Diversity, and Overlap results for dataset 23-Single-Element-DNPs RSCDD 2023-Pb}
\label{tab:si:misc-14}
\begin{tabular}{l|cccc|cccc|cccc}
\hline
\multicolumn{1}{c}{\textbf{}} & \multicolumn{4}{c}{\textbf{Entropy (nats)}} & \multicolumn{4}{c}{\textbf{Diversity (nats)}} & \multicolumn{4}{c}{\textbf{Overlap (\%)}} \\
\hline
Size (\%) & Random & K-means & Mean FPS & MSC (ours) & Random & K-means & Mean FPS & MSC (ours) & Random & K-means & Mean FPS & MSC (ours) \\
\hline

10.0 & 5.20 & 5.98 & 6.15 & 6.35 & 6.25 & 6.71 & 6.76 & 6.98 & 0.97 & 0.99 & 0.98 & 1.00 \\
25.0 & 5.32 & 5.96 & 5.96 & 6.15 & 6.57 & 6.90 & 6.92 & 6.93 & 0.99 & 1.00 & 1.00 & 1.00 \\
50.0 & 5.35 & 5.79 & 5.80 & 5.88 & 6.74 & 6.88 & 6.88 & 6.88 & 1.00 & 1.00 & 1.00 & 1.00 \\
75.0 & 5.38 & 5.54 & 5.54 & 5.75 & 6.80 & 6.85 & 6.85 & 6.85 & 1.00 & 1.00 & 1.00 & 1.00 \\
100.0 & 5.39 & 5.39 & 5.39 & 5.39 & 6.83 & 6.83 & 6.83 & 6.83 & 1.00 & 1.00 & 1.00 & 1.00 \\
\hline
\end{tabular}
\end{table*}

\begin{table*}[h!]
\centering
\tiny
\caption{Entropy, Diversity, and Overlap results for dataset 23-Single-Element-DNPs RSCDD 2023-Pd}
\label{tab:si:misc-15}
\begin{tabular}{l|cccc|cccc|cccc}
\hline
\multicolumn{1}{c}{\textbf{}} & \multicolumn{4}{c}{\textbf{Entropy (nats)}} & \multicolumn{4}{c}{\textbf{Diversity (nats)}} & \multicolumn{4}{c}{\textbf{Overlap (\%)}} \\
\hline
Size (\%) & Random & K-means & Mean FPS & MSC (ours) & Random & K-means & Mean FPS & MSC (ours) & Random & K-means & Mean FPS & MSC (ours) \\
\hline

10.0 & 7.07 & 7.59 & 7.89 & 8.78 & 8.14 & 8.36 & 8.46 & 9.00 & 0.79 & 0.80 & 0.77 & 0.75 \\
25.0 & 7.26 & 8.22 & 8.46 & 9.00 & 8.67 & 9.05 & 9.13 & 9.32 & 0.90 & 0.95 & 0.93 & 0.97 \\
50.0 & 7.40 & 8.15 & 8.19 & 8.57 & 8.96 & 9.29 & 9.30 & 9.25 & 0.97 & 1.00 & 1.00 & 1.00 \\
75.0 & 7.43 & 7.75 & 7.77 & 8.07 & 9.12 & 9.27 & 9.27 & 9.23 & 0.99 & 1.00 & 1.00 & 1.00 \\
100.0 & 7.43 & 7.43 & 7.43 & 7.43 & 9.22 & 9.22 & 9.22 & 9.22 & 1.00 & 1.00 & 1.00 & 1.00 \\
\hline
\end{tabular}
\end{table*}

\begin{table*}[h!]
\centering
\tiny
\caption{Entropy, Diversity, and Overlap results for dataset 23-Single-Element-DNPs RSCDD 2023-Pt}
\begin{tabular}{l|cccc|cccc|cccc}
\hline
\multicolumn{1}{c}{\textbf{}} & \multicolumn{4}{c}{\textbf{Entropy (nats)}} & \multicolumn{4}{c}{\textbf{Diversity (nats)}} & \multicolumn{4}{c}{\textbf{Overlap (\%)}} \\
\hline
Size (\%) & Random & K-means & Mean FPS & MSC (ours) & Random & K-means & Mean FPS & MSC (ours) & Random & K-means & Mean FPS & MSC (ours) \\
\hline

10.0 & 6.46 & 7.00 & 7.22 & 7.62 & 7.23 & 7.54 & 7.65 & 8.05 & 0.80 & 0.83 & 0.80 & 0.71 \\
25.0 & 6.82 & 7.59 & 7.73 & 7.92 & 7.74 & 8.29 & 8.35 & 8.43 & 0.92 & 0.96 & 0.96 & 0.99 \\
50.0 & 6.99 & 7.35 & 7.36 & 7.69 & 8.06 & 8.41 & 8.43 & 8.39 & 0.97 & 1.00 & 1.00 & 1.00 \\
75.0 & 7.01 & 7.15 & 7.15 & 7.39 & 8.21 & 8.39 & 8.39 & 8.36 & 0.99 & 1.00 & 1.00 & 1.00 \\
100.0 & 7.03 & 7.03 & 7.03 & 7.03 & 8.35 & 8.35 & 8.35 & 8.35 & 1.00 & 1.00 & 1.00 & 1.00 \\
\hline
\end{tabular}
\end{table*}

\begin{table*}[h!]
\centering
\tiny
\caption{Entropy, Diversity, and Overlap results for dataset 23-Single-Element-DNPs RSCDD 2023-Re}
\begin{tabular}{l|cccc|cccc|cccc}
\hline
\multicolumn{1}{c}{\textbf{}} & \multicolumn{4}{c}{\textbf{Entropy (nats)}} & \multicolumn{4}{c}{\textbf{Diversity (nats)}} & \multicolumn{4}{c}{\textbf{Overlap (\%)}} \\
\hline
Size (\%) & Random & K-means & Mean FPS & MSC (ours) & Random & K-means & Mean FPS & MSC (ours) & Random & K-means & Mean FPS & MSC (ours) \\
\hline

10.0 & 7.81 & 8.14 & 8.15 & 8.33 & 8.26 & 8.50 & 8.50 & 8.71 & 0.68 & 0.72 & 0.66 & 0.63 \\
25.0 & 8.12 & 8.25 & 8.34 & 8.55 & 8.69 & 8.90 & 8.96 & 9.07 & 0.90 & 0.94 & 0.91 & 0.97 \\
50.0 & 8.20 & 8.16 & 8.18 & 8.58 & 8.89 & 9.05 & 9.07 & 9.08 & 0.98 & 0.99 & 1.00 & 1.00 \\
75.0 & 8.27 & 8.25 & 8.26 & 8.54 & 8.99 & 9.05 & 9.06 & 9.06 & 1.00 & 1.00 & 1.00 & 1.00 \\
100.0 & 8.29 & 8.29 & 8.29 & 8.29 & 9.04 & 9.04 & 9.04 & 9.04 & 1.00 & 1.00 & 1.00 & 1.00 \\
\hline
\end{tabular}
\end{table*}
\clearpage

\begin{table*}[h!]
\centering
\tiny
\caption{Entropy, Diversity, and Overlap results for dataset 23-Single-Element-DNPs RSCDD 2023-Sb}
\begin{tabular}{l|cccc|cccc|cccc}
\hline
\multicolumn{1}{c}{\textbf{}} & \multicolumn{4}{c}{\textbf{Entropy (nats)}} & \multicolumn{4}{c}{\textbf{Diversity (nats)}} & \multicolumn{4}{c}{\textbf{Overlap (\%)}} \\
\hline
Size (\%) & Random & K-means & Mean FPS & MSC (ours) & Random & K-means & Mean FPS & MSC (ours) & Random & K-means & Mean FPS & MSC (ours) \\
\hline

10.0 & 6.41 & 6.82 & 6.92 & 7.14 & 7.08 & 7.37 & 7.40 & 7.70 & 0.96 & 0.98 & 0.96 & 0.99 \\
25.0 & 6.54 & 6.86 & 6.84 & 7.04 & 7.36 & 7.59 & 7.61 & 7.70 & 0.99 & 0.99 & 0.99 & 1.00 \\
50.0 & 6.57 & 6.84 & 6.85 & 6.90 & 7.48 & 7.66 & 7.67 & 7.66 & 1.00 & 1.00 & 1.00 & 1.00 \\
75.0 & 6.59 & 6.74 & 6.75 & 6.78 & 7.56 & 7.63 & 7.63 & 7.63 & 1.00 & 1.00 & 1.00 & 1.00 \\
100.0 & 6.62 & 6.62 & 6.62 & 6.62 & 7.61 & 7.61 & 7.61 & 7.61 & 1.00 & 1.00 & 1.00 & 1.00 \\
\hline
\end{tabular}
\end{table*}

\begin{table*}[h!]
\centering
\tiny
\caption{Entropy, Diversity, and Overlap results for dataset 23-Single-Element-DNPs RSCDD 2023-Sr}
\begin{tabular}{l|cccc|cccc|cccc}
\hline
\multicolumn{1}{c}{\textbf{}} & \multicolumn{4}{c}{\textbf{Entropy (nats)}} & \multicolumn{4}{c}{\textbf{Diversity (nats)}} & \multicolumn{4}{c}{\textbf{Overlap (\%)}} \\
\hline
Size (\%) & Random & K-means & Mean FPS & MSC (ours) & Random & K-means & Mean FPS & MSC (ours) & Random & K-means & Mean FPS & MSC (ours) \\
\hline

10.0 & 4.14 & 4.50 & 4.67 & 4.68 & 5.06 & 5.36 & 5.39 & 5.37 & 0.98 & 1.00 & 0.99 & 1.00 \\
25.0 & 4.16 & 4.51 & 4.54 & 4.66 & 5.27 & 5.39 & 5.40 & 5.40 & 1.00 & 1.00 & 1.00 & 1.00 \\
50.0 & 4.18 & 4.43 & 4.45 & 4.60 & 5.34 & 5.38 & 5.38 & 5.39 & 1.00 & 1.00 & 1.00 & 1.00 \\
75.0 & 4.17 & 4.22 & 4.22 & 4.46 & 5.37 & 5.38 & 5.38 & 5.38 & 1.00 & 1.00 & 1.00 & 1.00 \\
100.0 & 4.16 & 4.16 & 4.16 & 4.16 & 5.38 & 5.38 & 5.38 & 5.38 & 1.00 & 1.00 & 1.00 & 1.00 \\
\hline
\end{tabular}
\end{table*}

\begin{table*}[h!]
\centering
\tiny
\caption{Entropy, Diversity, and Overlap results for dataset 23-Single-Element-DNPs RSCDD 2023-Ti}
\begin{tabular}{l|cccc|cccc|cccc}
\hline
\multicolumn{1}{c}{\textbf{}} & \multicolumn{4}{c}{\textbf{Entropy (nats)}} & \multicolumn{4}{c}{\textbf{Diversity (nats)}} & \multicolumn{4}{c}{\textbf{Overlap (\%)}} \\
\hline
Size (\%) & Random & K-means & Mean FPS & MSC (ours) & Random & K-means & Mean FPS & MSC (ours) & Random & K-means & Mean FPS & MSC (ours) \\
\hline

10.0 & 8.35 & 8.62 & 8.81 & 9.27 & 8.76 & 8.98 & 9.08 & 9.50 & 0.62 & 0.64 & 0.57 & 0.47 \\
25.0 & 8.69 & 9.26 & 9.39 & 9.59 & 9.33 & 9.70 & 9.77 & 9.97 & 0.83 & 0.88 & 0.83 & 0.85 \\
50.0 & 8.87 & 9.35 & 9.36 & 9.50 & 9.69 & 10.02 & 10.04 & 10.05 & 0.93 & 0.99 & 0.99 & 1.00 \\
75.0 & 8.94 & 9.18 & 9.18 & 9.32 & 9.88 & 10.03 & 10.03 & 10.03 & 0.98 & 1.00 & 1.00 & 1.00 \\
100.0 & 8.98 & 8.98 & 8.98 & 8.98 & 10.01 & 10.01 & 10.01 & 10.01 & 1.00 & 1.00 & 1.00 & 1.00 \\
\hline
\end{tabular}
\end{table*}

\begin{table*}[h!]
\centering
\tiny
\caption{Entropy, Diversity, and Overlap results for dataset 23-Single-Element-DNPs RSCDD 2023-Zn}
\begin{tabular}{l|cccc|cccc|cccc}
\hline
\multicolumn{1}{c}{\textbf{}} & \multicolumn{4}{c}{\textbf{Entropy (nats)}} & \multicolumn{4}{c}{\textbf{Diversity (nats)}} & \multicolumn{4}{c}{\textbf{Overlap (\%)}} \\
\hline
Size (\%) & Random & K-means & Mean FPS & MSC (ours) & Random & K-means & Mean FPS & MSC (ours) & Random & K-means & Mean FPS & MSC (ours) \\
\hline

10.0 & 6.26 & 7.02 & 7.16 & 7.42 & 2.00 & 2.04 & 2.05 & 2.07 & 0.89 & 0.94 & 0.93 & 0.95 \\
25.0 & 6.39 & 6.99 & 7.13 & 7.51 & 2.04 & 2.07 & 2.07 & 2.08 & 0.97 & 0.99 & 0.99 & 1.00 \\
50.0 & 6.50 & 6.95 & 7.03 & 7.46 & 2.06 & 2.07 & 2.08 & 2.08 & 1.00 & 1.00 & 1.00 & 1.00 \\
75.0 & 6.54 & 6.74 & 6.75 & 7.19 & 2.07 & 2.07 & 2.07 & 2.08 & 1.00 & 1.00 & 1.00 & 1.00 \\
100.0 & 6.56 & 6.56 & 6.56 & 6.56 & 2.07 & 2.07 & 2.07 & 2.07 & 1.00 & 1.00 & 1.00 & 1.00 \\
\hline
\end{tabular}
\end{table*}

\begin{table*}[h!]
\centering
\tiny
\caption{Entropy, Diversity, and Overlap results for dataset 23-Single-Element-DNPs RSCDD 2023-Zr}
\begin{tabular}{l|cccc|cccc|cccc}
\hline
\multicolumn{1}{c}{\textbf{}} & \multicolumn{4}{c}{\textbf{Entropy (nats)}} & \multicolumn{4}{c}{\textbf{Diversity (nats)}} & \multicolumn{4}{c}{\textbf{Overlap (\%)}} \\
\hline
Size (\%) & Random & K-means & Mean FPS & MSC (ours) & Random & K-means & Mean FPS & MSC (ours) & Random & K-means & Mean FPS & MSC (ours) \\
\hline

10.0 & 6.49 & 7.00 & 7.08 & 7.24 & 7.26 & 7.63 & 7.67 & 7.89 & 0.90 & 0.93 & 0.90 & 0.94 \\
25.0 & 6.67 & 7.00 & 7.04 & 7.21 & 7.55 & 7.88 & 7.91 & 7.93 & 0.97 & 0.99 & 0.99 & 0.99 \\
50.0 & 6.75 & 6.82 & 6.85 & 7.10 & 7.77 & 7.91 & 7.93 & 7.94 & 0.99 & 1.00 & 1.00 & 1.00 \\
75.0 & 6.74 & 6.88 & 6.88 & 7.03 & 7.85 & 7.92 & 7.93 & 7.94 & 1.00 & 1.00 & 1.00 & 1.00 \\
100.0 & 6.74 & 6.74 & 6.74 & 6.74 & 7.92 & 7.92 & 7.92 & 7.92 & 1.00 & 1.00 & 1.00 & 1.00 \\
\hline
\end{tabular}
\end{table*}

\begin{table*}[h!]
\centering
\tiny
\caption{Entropy, Diversity, and Overlap results for dataset Ag-PBE MSMSE 2021}
\begin{tabular}{l|cccc|cccc|cccc}
\hline
\multicolumn{1}{c}{\textbf{}} & \multicolumn{4}{c}{\textbf{Entropy (nats)}} & \multicolumn{4}{c}{\textbf{Diversity (nats)}} & \multicolumn{4}{c}{\textbf{Overlap (\%)}} \\
\hline
Size (\%) & Random & K-means & Mean FPS & MSC (ours) & Random & K-means & Mean FPS & MSC (ours) & Random & K-means & Mean FPS & MSC (ours) \\
\hline

10.0 & 4.68 & 6.92 & 8.04 & 8.57 & 7.65 & 8.32 & 8.49 & 8.80 & 0.85 & 0.91 & 0.91 & 0.83 \\
25.0 & 4.95 & 7.48 & 8.34 & 8.70 & 8.36 & 8.97 & 9.05 & 9.14 & 0.93 & 0.98 & 0.99 & 0.97 \\
50.0 & 4.98 & 6.65 & 6.67 & 7.07 & 8.75 & 9.13 & 9.13 & 9.13 & 0.97 & 1.00 & 1.00 & 1.00 \\
75.0 & 4.97 & 5.55 & 5.55 & 5.97 & 8.97 & 9.13 & 9.13 & 9.13 & 0.99 & 1.00 & 1.00 & 1.00 \\
100.0 & 5.00 & 5.00 & 5.00 & 5.00 & 9.13 & 9.13 & 9.13 & 9.13 & 1.00 & 1.00 & 1.00 & 1.00 \\
\hline
\end{tabular}
\end{table*}
\clearpage

\begin{table*}[h!]
\centering
\tiny
\caption{Entropy, Diversity, and Overlap results for dataset Au-PBE MSMSE 2021}
\begin{tabular}{l|cccc|cccc|cccc}
\hline
\multicolumn{1}{c}{\textbf{}} & \multicolumn{4}{c}{\textbf{Entropy (nats)}} & \multicolumn{4}{c}{\textbf{Diversity (nats)}} & \multicolumn{4}{c}{\textbf{Overlap (\%)}} \\
\hline
Size (\%) & Random & K-means & Mean FPS & MSC (ours) & Random & K-means & Mean FPS & MSC (ours) & Random & K-means & Mean FPS & MSC (ours) \\
\hline

10.0 & 6.75 & 7.90 & 8.39 & 8.75 & 2.15 & 2.17 & 2.18 & 2.21 & 0.90 & 0.91 & 0.91 & 0.89 \\
25.0 & 6.84 & 8.21 & 8.54 & 8.79 & 2.19 & 2.22 & 2.22 & 2.23 & 0.97 & 0.99 & 0.99 & 0.99 \\
50.0 & 6.87 & 6.90 & 6.91 & 8.76 & 2.21 & 2.23 & 2.23 & 2.23 & 0.99 & 1.00 & 1.00 & 1.00 \\
75.0 & 6.90 & 6.90 & 6.91 & 8.09 & 2.22 & 2.23 & 2.23 & 2.23 & 1.00 & 1.00 & 1.00 & 1.00 \\
100.0 & 6.92 & 6.92 & 6.92 & 6.92 & 2.23 & 2.23 & 2.23 & 2.23 & 1.00 & 1.00 & 1.00 & 1.00 \\
\hline
\end{tabular}
\end{table*}

\begin{table*}[h!]
\centering
\tiny
\caption{Entropy, Diversity, and Overlap results for dataset CA-9}
\begin{tabular}{l|cccc|cccc|cccc}
\hline
\multicolumn{1}{c}{\textbf{}} & \multicolumn{4}{c}{\textbf{Entropy (nats)}} & \multicolumn{4}{c}{\textbf{Diversity (nats)}} & \multicolumn{4}{c}{\textbf{Overlap (\%)}} \\
\hline
Size (\%) & Random & K-means & Mean FPS & MSC (ours) & Random & K-means & Mean FPS & MSC (ours) & Random & K-means & Mean FPS & MSC (ours) \\
\hline

10.0 & 8.40 & 9.94 & 10.16 & 10.97 & 10.86 & 11.45 & 11.53 & 11.71 & 0.80 & 0.84 & 0.85 & 0.87 \\
25.0 & 8.58 & 9.45 & 9.46 & 10.67 & 11.48 & 11.80 & 11.82 & 12.02 & 0.88 & 0.93 & 0.92 & 0.98 \\
50.0 & 8.66 & 9.01 & 8.99 & 10.48 & 11.84 & 11.93 & 11.96 & 12.09 & 0.96 & 0.97 & 0.97 & 1.00 \\
75.0 & 8.69 & 8.90 & 8.90 & 9.58 & 11.99 & 12.01 & 12.03 & 12.08 & 0.99 & 0.99 & 0.99 & 1.00 \\
100.0 & 8.71 & 8.71 & 8.71 & 8.71 & 12.08 & 12.08 & 12.08 & 12.08 & 1.00 & 1.00 & 1.00 & 1.00 \\
\hline
\end{tabular}
\end{table*}

\begin{table*}[h!]
\centering
\tiny
\caption{Entropy, Diversity, and Overlap results for dataset CA-9 BB}
\begin{tabular}{l|cccc|cccc|cccc}
\hline
\multicolumn{1}{c}{\textbf{}} & \multicolumn{4}{c}{\textbf{Entropy (nats)}} & \multicolumn{4}{c}{\textbf{Diversity (nats)}} & \multicolumn{4}{c}{\textbf{Overlap (\%)}} \\
\hline
Size (\%) & Random & K-means & Mean FPS & MSC (ours) & Random & K-means & Mean FPS & MSC (ours) & Random & K-means & Mean FPS & MSC (ours) \\
\hline

10.0 & 9.40 & 10.37 & 10.48 & 11.02 & 10.80 & 11.20 & 11.24 & 11.37 & 0.65 & 0.67 & 0.67 & 0.69 \\
25.0 & 9.66 & 10.17 & 10.12 & 11.04 & 11.43 & 11.69 & 11.70 & 11.88 & 0.78 & 0.83 & 0.82 & 0.90 \\
50.0 & 9.78 & 9.67 & 9.63 & 11.00 & 11.79 & 11.87 & 11.90 & 12.03 & 0.92 & 0.93 & 0.93 & 1.00 \\
75.0 & 9.83 & 9.57 & 9.56 & 10.64 & 11.95 & 11.96 & 11.98 & 12.04 & 0.98 & 0.98 & 0.98 & 1.00 \\
100.0 & 9.87 & 9.87 & 9.87 & 9.87 & 12.04 & 12.04 & 12.04 & 12.04 & 1.00 & 1.00 & 1.00 & 1.00 \\
\hline
\end{tabular}
\end{table*}

\begin{table*}[h!]
\centering
\tiny
\caption{Entropy, Diversity, and Overlap results for dataset CA-9 BR}
\begin{tabular}{l|cccc|cccc|cccc}
\hline
\multicolumn{1}{c}{\textbf{}} & \multicolumn{4}{c}{\textbf{Entropy (nats)}} & \multicolumn{4}{c}{\textbf{Diversity (nats)}} & \multicolumn{4}{c}{\textbf{Overlap (\%)}} \\
\hline
Size (\%) & Random & K-means & Mean FPS & MSC (ours) & Random & K-means & Mean FPS & MSC (ours) & Random & K-means & Mean FPS & MSC (ours) \\
\hline

10.0 & 9.37 & 10.35 & 10.48 & 10.99 & 10.78 & 11.19 & 11.24 & 11.37 & 0.65 & 0.67 & 0.67 & 0.69 \\
25.0 & 9.63 & 10.16 & 10.13 & 11.05 & 11.40 & 11.68 & 11.70 & 11.88 & 0.78 & 0.83 & 0.82 & 0.90 \\
50.0 & 9.77 & 9.66 & 9.63 & 10.99 & 11.78 & 11.87 & 11.90 & 12.03 & 0.92 & 0.93 & 0.93 & 1.00 \\
75.0 & 9.83 & 9.57 & 9.56 & 10.64 & 11.94 & 11.96 & 11.98 & 12.04 & 0.98 & 0.98 & 0.98 & 1.00 \\
100.0 & 9.87 & 9.87 & 9.87 & 9.87 & 12.04 & 12.04 & 12.04 & 12.04 & 1.00 & 1.00 & 1.00 & 1.00 \\
\hline
\end{tabular}
\end{table*}

\begin{table*}[h!]
\centering
\tiny
\caption{Entropy, Diversity, and Overlap results for dataset CA-9 RR}
\begin{tabular}{l|cccc|cccc|cccc}
\hline
\multicolumn{1}{c}{\textbf{}} & \multicolumn{4}{c}{\textbf{Entropy (nats)}} & \multicolumn{4}{c}{\textbf{Diversity (nats)}} & \multicolumn{4}{c}{\textbf{Overlap (\%)}} \\
\hline
Size (\%) & Random & K-means & Mean FPS & MSC (ours) & Random & K-means & Mean FPS & MSC (ours) & Random & K-means & Mean FPS & MSC (ours) \\
\hline

10.0 & 8.69 & 9.88 & 10.30 & 10.89 & 10.49 & 11.05 & 11.19 & 11.34 & 0.76 & 0.78 & 0.78 & 0.80 \\
25.0 & 8.87 & 9.69 & 9.81 & 10.76 & 11.15 & 11.57 & 11.62 & 11.82 & 0.83 & 0.89 & 0.89 & 0.95 \\
50.0 & 9.00 & 9.22 & 9.23 & 10.47 & 11.58 & 11.76 & 11.79 & 11.93 & 0.92 & 0.94 & 0.94 & 1.00 \\
75.0 & 9.05 & 8.97 & 8.96 & 9.71 & 11.80 & 11.85 & 11.86 & 11.92 & 0.97 & 0.97 & 0.98 & 1.00 \\
100.0 & 9.08 & 9.08 & 9.08 & 9.08 & 11.92 & 11.92 & 11.92 & 11.92 & 1.00 & 1.00 & 1.00 & 1.00 \\
\hline
\end{tabular}
\end{table*}

\begin{table*}[h!]
\centering
\tiny
\caption{Entropy, Diversity, and Overlap results for dataset CGM-MLP natcomm2023 screening amorphous carbon}
\begin{tabular}{l|cccc|cccc|cccc}
\hline
\multicolumn{1}{c}{\textbf{}} & \multicolumn{4}{c}{\textbf{Entropy (nats)}} & \multicolumn{4}{c}{\textbf{Diversity (nats)}} & \multicolumn{4}{c}{\textbf{Overlap (\%)}} \\
\hline
Size (\%) & Random & K-means & Mean FPS & MSC (ours) & Random & K-means & Mean FPS & MSC (ours) & Random & K-means & Mean FPS & MSC (ours) \\
\hline

10.0 & 9.92 & 9.97 & 9.66 & 10.05 & 2.29 & 2.30 & 2.27 & 2.31 & 0.06 & 0.06 & 0.05 & 0.06 \\
25.0 & 10.81 & 10.79 & 10.60 & 10.97 & 2.38 & 2.38 & 2.36 & 2.40 & 0.17 & 0.16 & 0.14 & 0.18 \\
50.0 & 11.49 & 11.44 & 11.37 & 11.63 & 2.44 & 2.44 & 2.43 & 2.45 & 0.39 & 0.35 & 0.33 & 0.40 \\
75.0 & 11.89 & 11.85 & 11.84 & 11.97 & 2.48 & 2.47 & 2.47 & 2.48 & 0.60 & 0.57 & 0.56 & 0.61 \\
100.0 & 12.18 & 12.18 & 12.18 & 12.18 & 2.50 & 2.50 & 2.50 & 2.50 & 1.00 & 1.00 & 1.00 & 1.00 \\
\hline
\end{tabular}
\end{table*}
\clearpage

\begin{table*}[h!]
\centering
\tiny
\caption{Entropy, Diversity, and Overlap results for dataset CGM-MLP natcomm2023 screening graphite train}
\begin{tabular}{l|cccc|cccc|cccc}
\hline
\multicolumn{1}{c}{\textbf{}} & \multicolumn{4}{c}{\textbf{Entropy (nats)}} & \multicolumn{4}{c}{\textbf{Diversity (nats)}} & \multicolumn{4}{c}{\textbf{Overlap (\%)}} \\
\hline
Size (\%) & Random & K-means & Mean FPS & MSC (ours) & Random & K-means & Mean FPS & MSC (ours) & Random & K-means & Mean FPS & MSC (ours) \\
\hline

10.0 & 2.29 & 2.31 & 2.30 & 2.31 & 2.29 & 2.32 & 2.30 & 2.31 & 0.28 & 0.23 & 0.23 & 0.28 \\
25.0 & 2.97 & 3.10 & 3.10 & 3.10 & 2.99 & 3.10 & 3.10 & 3.11 & 0.44 & 0.47 & 0.59 & 0.57 \\
50.0 & 3.55 & 3.74 & 3.74 & 3.75 & 3.57 & 3.74 & 3.74 & 3.75 & 0.70 & 0.91 & 0.90 & 0.95 \\
75.0 & 3.90 & 4.03 & 4.06 & 4.08 & 3.93 & 4.04 & 4.07 & 4.08 & 0.93 & 0.99 & 1.00 & 1.00 \\
100.0 & 4.17 & 4.17 & 4.17 & 4.17 & 4.19 & 4.19 & 4.19 & 4.19 & 1.00 & 1.00 & 1.00 & 1.00 \\
\hline
\end{tabular}
\end{table*}

\begin{table*}[h!]
\centering
\tiny
\caption{Entropy, Diversity, and Overlap results for dataset C NPJ2020}
\begin{tabular}{l|cccc|cccc|cccc}
\hline
\multicolumn{1}{c}{\textbf{}} & \multicolumn{4}{c}{\textbf{Entropy (nats)}} & \multicolumn{4}{c}{\textbf{Diversity (nats)}} & \multicolumn{4}{c}{\textbf{Overlap (\%)}} \\
\hline
Size (\%) & Random & K-means & Mean FPS & MSC (ours) & Random & K-means & Mean FPS & MSC (ours) & Random & K-means & Mean FPS & MSC (ours) \\
\hline

10.0 & 5.69 & 6.68 & 6.85 & 8.27 & 7.62 & 8.15 & 8.22 & 8.71 & 0.94 & 0.96 & 0.96 & 0.99 \\
25.0 & 5.74 & 6.61 & 6.66 & 7.85 & 8.00 & 8.45 & 8.45 & 8.66 & 0.97 & 0.99 & 0.99 & 1.00 \\
50.0 & 5.81 & 6.32 & 6.35 & 6.95 & 8.28 & 8.51 & 8.52 & 8.55 & 0.99 & 1.00 & 1.00 & 1.00 \\
75.0 & 5.86 & 6.04 & 6.03 & 6.26 & 8.46 & 8.53 & 8.53 & 8.53 & 1.00 & 1.00 & 1.00 & 1.00 \\
100.0 & 5.83 & 5.83 & 5.83 & 5.83 & 8.53 & 8.53 & 8.53 & 8.53 & 1.00 & 1.00 & 1.00 & 1.00 \\
\hline
\end{tabular}
\end{table*}

\begin{table*}[h!]
\centering
\tiny
\caption{Entropy, Diversity, and Overlap results for dataset DP-GEN Cu}
\begin{tabular}{l|cccc|cccc|cccc}
\hline
\multicolumn{1}{c}{\textbf{}} & \multicolumn{4}{c}{\textbf{Entropy (nats)}} & \multicolumn{4}{c}{\textbf{Diversity (nats)}} & \multicolumn{4}{c}{\textbf{Overlap (\%)}} \\
\hline
Size (\%) & Random & K-means & Mean FPS & MSC (ours) & Random & K-means & Mean FPS & MSC (ours) & Random & K-means & Mean FPS & MSC (ours) \\
\hline

10.0 & 6.75 & 8.32 & 9.37 & 9.95 & 9.05 & 9.48 & 9.66 & 9.99 & 0.73 & 0.76 & 0.75 & 0.51 \\
25.0 & 7.03 & 9.31 & 10.04 & 10.55 & 9.79 & 10.34 & 10.44 & 10.67 & 0.81 & 0.88 & 0.89 & 0.83 \\
50.0 & 7.14 & 9.61 & 9.84 & 9.90 & 10.31 & 10.78 & 10.80 & 10.80 & 0.89 & 0.99 & 1.00 & 1.00 \\
75.0 & 7.24 & 8.29 & 8.30 & 8.49 & 10.60 & 10.80 & 10.80 & 10.80 & 0.95 & 1.00 & 1.00 & 1.00 \\
100.0 & 7.30 & 7.30 & 7.30 & 7.30 & 10.80 & 10.80 & 10.80 & 10.80 & 1.00 & 1.00 & 1.00 & 1.00 \\
\hline
\end{tabular}
\end{table*}

\begin{table*}[h!]
\centering
\tiny
\caption{Entropy, Diversity, and Overlap results for dataset Fe nanoparticles PRB 2023}
\begin{tabular}{l|cccc|cccc|cccc}
\hline
\multicolumn{1}{c}{\textbf{}} & \multicolumn{4}{c}{\textbf{Entropy (nats)}} & \multicolumn{4}{c}{\textbf{Diversity (nats)}} & \multicolumn{4}{c}{\textbf{Overlap (\%)}} \\
\hline
Size (\%) & Random & K-means & Mean FPS & MSC (ours) & Random & K-means & Mean FPS & MSC (ours) & Random & K-means & Mean FPS & MSC (ours) \\
\hline

10.0 & 6.64 & 5.82 & 5.43 & 5.99 & 6.90 & 6.10 & 5.81 & 6.21 & 0.64 & 0.38 & 0.25 & 0.27 \\
25.0 & 6.89 & 6.33 & 6.11 & 7.17 & 7.36 & 6.82 & 6.62 & 7.51 & 0.80 & 0.61 & 0.55 & 0.82 \\
50.0 & 7.08 & 6.72 & 6.68 & 7.41 & 7.70 & 7.41 & 7.39 & 7.93 & 0.91 & 0.82 & 0.81 & 0.96 \\
75.0 & 7.23 & 7.02 & 7.05 & 7.38 & 7.95 & 7.79 & 7.82 & 8.06 & 0.97 & 0.93 & 0.94 & 1.00 \\
100.0 & 7.26 & 7.26 & 7.26 & 7.26 & 8.07 & 8.07 & 8.07 & 8.07 & 1.00 & 1.00 & 1.00 & 1.00 \\
\hline
\end{tabular}
\end{table*}

\begin{table*}[h!]
\centering
\tiny
\caption{Entropy, Diversity, and Overlap results for dataset FitSNAP Fe NPJ 2021}
\begin{tabular}{l|cccc|cccc|cccc}
\hline
\multicolumn{1}{c}{\textbf{}} & \multicolumn{4}{c}{\textbf{Entropy (nats)}} & \multicolumn{4}{c}{\textbf{Diversity (nats)}} & \multicolumn{4}{c}{\textbf{Overlap (\%)}} \\
\hline
Size (\%) & Random & K-means & Mean FPS & MSC (ours) & Random & K-means & Mean FPS & MSC (ours) & Random & K-means & Mean FPS & MSC (ours) \\
\hline

10.0 & 7.46 & 6.92 & 5.95 & 6.58 & 7.84 & 7.06 & 6.00 & 7.48 & 0.77 & 0.19 & 0.01 & 0.11 \\
25.0 & 7.72 & 7.33 & 6.96 & 8.03 & 8.17 & 7.51 & 7.07 & 8.42 & 0.92 & 0.38 & 0.09 & 0.77 \\
50.0 & 7.77 & 7.92 & 7.73 & 8.20 & 8.35 & 8.19 & 8.02 & 8.55 & 0.97 & 0.80 & 0.60 & 0.99 \\
75.0 & 7.85 & 8.14 & 8.14 & 7.96 & 8.49 & 8.56 & 8.59 & 8.55 & 0.99 & 0.99 & 0.98 & 1.00 \\
100.0 & 7.88 & 7.88 & 7.88 & 7.88 & 8.56 & 8.56 & 8.56 & 8.56 & 1.00 & 1.00 & 1.00 & 1.00 \\
\hline
\end{tabular}
\end{table*}

\begin{table*}[h!]
\centering
\tiny
\caption{Entropy, Diversity, and Overlap results for dataset Mg edmonds 2022}
\begin{tabular}{l|cccc|cccc|cccc}
\hline
\multicolumn{1}{c}{\textbf{}} & \multicolumn{4}{c}{\textbf{Entropy (nats)}} & \multicolumn{4}{c}{\textbf{Diversity (nats)}} & \multicolumn{4}{c}{\textbf{Overlap (\%)}} \\
\hline
Size (\%) & Random & K-means & Mean FPS & MSC (ours) & Random & K-means & Mean FPS & MSC (ours) & Random & K-means & Mean FPS & MSC (ours) \\
\hline

10.0 & 7.20 & 7.58 & 7.43 & 8.10 & 7.87 & 7.88 & 7.66 & 8.41 & 0.41 & 0.42 & 0.19 & 0.20 \\
25.0 & 7.77 & 8.44 & 8.42 & 9.10 & 8.62 & 8.75 & 8.67 & 9.35 & 0.59 & 0.65 & 0.50 & 0.48 \\
50.0 & 8.18 & 9.02 & 9.10 & 9.42 & 9.18 & 9.36 & 9.39 & 9.66 & 0.78 & 0.89 & 0.89 & 0.97 \\
75.0 & 8.41 & 9.19 & 9.21 & 9.33 & 9.48 & 9.68 & 9.69 & 9.70 & 0.90 & 0.99 & 1.00 & 1.00 \\
100.0 & 8.55 & 8.55 & 8.55 & 8.55 & 9.70 & 9.70 & 9.70 & 9.70 & 1.00 & 1.00 & 1.00 & 1.00 \\
\hline
\end{tabular}
\end{table*}
\clearpage

\begin{table*}[h!]
\centering
\tiny
\caption{Entropy, Diversity, and Overlap results for dataset Mo PRM2019}
\begin{tabular}{l|cccc|cccc|cccc}
\hline
\multicolumn{1}{c}{\textbf{}} & \multicolumn{4}{c}{\textbf{Entropy (nats)}} & \multicolumn{4}{c}{\textbf{Diversity (nats)}} & \multicolumn{4}{c}{\textbf{Overlap (\%)}} \\
\hline
Size (\%) & Random & K-means & Mean FPS & MSC (ours) & Random & K-means & Mean FPS & MSC (ours) & Random & K-means & Mean FPS & MSC (ours) \\
\hline

10.0 & 4.46 & 6.57 & 7.36 & 6.59 & 6.85 & 7.48 & 7.59 & 8.74 & 0.79 & 0.81 & 0.77 & 0.97 \\
25.0 & 4.91 & 6.67 & 7.30 & 6.04 & 7.73 & 7.92 & 8.01 & 8.77 & 0.86 & 0.87 & 0.86 & 1.00 \\
50.0 & 5.03 & 6.76 & 6.95 & 5.28 & 8.25 & 8.44 & 8.43 & 8.77 & 0.92 & 0.93 & 0.93 & 1.00 \\
75.0 & 5.16 & 6.44 & 6.52 & 5.20 & 8.57 & 8.71 & 8.72 & 8.77 & 0.96 & 0.98 & 0.98 & 1.00 \\
100.0 & 5.14 & 5.14 & 5.14 & 5.14 & 8.77 & 8.77 & 8.77 & 8.77 & 1.00 & 1.00 & 1.00 & 1.00 \\
\hline
\end{tabular}
\end{table*}

\begin{table*}[h!]
\centering
\tiny
\caption{Entropy, Diversity, and Overlap results for dataset NDSC TUT 2022}
\begin{tabular}{l|cccc|cccc|cccc}
\hline
\multicolumn{1}{c}{\textbf{}} & \multicolumn{4}{c}{\textbf{Entropy (nats)}} & \multicolumn{4}{c}{\textbf{Diversity (nats)}} & \multicolumn{4}{c}{\textbf{Overlap (\%)}} \\
\hline
Size (\%) & Random & K-means & Mean FPS & MSC (ours) & Random & K-means & Mean FPS & MSC (ours) & Random & K-means & Mean FPS & MSC (ours) \\
\hline

10.0 & 0.16 & 0.20 & 0.29 & 0.51 & 0.16 & 0.20 & 0.30 & 0.51 & 1.00 & 1.00 & 1.00 & 1.00 \\
25.0 & 0.16 & 0.23 & 0.26 & 0.41 & 0.17 & 0.23 & 0.27 & 0.41 & 1.00 & 1.00 & 1.00 & 1.00 \\
50.0 & 0.18 & 0.22 & 0.22 & 0.31 & 0.19 & 0.22 & 0.22 & 0.31 & 1.00 & 1.00 & 1.00 & 1.00 \\
75.0 & 0.20 & 0.20 & 0.20 & 0.24 & 0.21 & 0.21 & 0.21 & 0.25 & 1.00 & 1.00 & 1.00 & 1.00 \\
100.0 & 0.20 & 0.20 & 0.20 & 0.20 & 0.20 & 0.20 & 0.20 & 0.20 & 1.00 & 1.00 & 1.00 & 1.00 \\
\hline
\end{tabular}
\end{table*}

\begin{table*}[h!]
\centering
\tiny
\caption{Entropy, Diversity, and Overlap results for dataset NEP qHPF}
%
\end{table*}

\begin{table*}[h!]
\centering
\tiny
\caption{Entropy, Diversity, and Overlap results for dataset Nb PRM2019}
%
\end{table*}

\begin{table*}[h!]
\centering
\tiny
\caption{Entropy, Diversity, and Overlap results for dataset Si JCP 2017}
%
\end{table*}

\begin{table*}[h!]
\centering
\tiny
\caption{Entropy, Diversity, and Overlap results for dataset Si PRX GAP}
%
\end{table*}
\clearpage

\begin{table*}[h!]
\centering
\tiny
\caption{Entropy, Diversity, and Overlap results for dataset Sn-SCAN PRM 2023}
%
\end{table*}

\begin{table*}[h!]
\centering
\tiny
\caption{Entropy, Diversity, and Overlap results for dataset Ta Linear JCP2014}
%
\end{table*}

\begin{table*}[h!]
\centering
\tiny
\caption{Entropy, Diversity, and Overlap results for dataset Ta PINN 2021}
%
\end{table*}

\begin{table*}[h!]
\centering
\tiny
\caption{Entropy, Diversity, and Overlap results for dataset Ta PRM2019}
%
\end{table*}

\begin{table*}[h!]
\centering
\tiny
\caption{Entropy, Diversity, and Overlap results for dataset Ti NPJCM 2021}
%
\end{table*}

\begin{table*}[h!]
\centering
\tiny
\caption{Entropy, Diversity, and Overlap results for dataset V PRM2019}
%
\end{table*}
\clearpage

\begin{table*}[h!]
\centering
\tiny
\caption{Entropy, Diversity, and Overlap results for dataset W-14}
%
\end{table*}

\begin{table*}[h!]
\centering
\tiny
\caption{Entropy, Diversity, and Overlap results for dataset W LML-retrain bulk MD test}
%
\end{table*}

\begin{table*}[h!]
\centering
\tiny
\caption{Entropy, Diversity, and Overlap results for dataset W PRB2019}
%
\end{table*}

\begin{table*}[h!]
\centering
\tiny
\caption{Entropy, Diversity, and Overlap results for dataset Zn MTP CMS2023}
%
\end{table*}

\begin{table*}[h!]
\centering
\tiny
\caption{Entropy, Diversity, and Overlap results for dataset aC JCP 2023}
%
\end{table*}

\begin{table*}[h!]
\centering
\tiny
\caption{Entropy, Diversity, and Overlap results for dataset defected phosphorene ACS 2023}
%
\end{table*}
\clearpage

\begin{table*}[h!]
\centering
\tiny
\caption{Entropy, Diversity, and Overlap results for dataset discrepencies and error metrics NPJ 2023 enhanced validation set}
%
\end{table*}

\begin{table*}[h!]
\centering
\tiny
\caption{Entropy, Diversity, and Overlap results for dataset discrepencies and error metrics NPJ 2023 interstitial enhanced training set}
%
\end{table*}

\begin{table*}[h!]
\centering
\tiny
\caption{Entropy, Diversity, and Overlap results for dataset discrepencies and error metrics NPJ 2023 vacancy enhanced training set}
%
\end{table*}

\begin{table*}[h!]
\centering
\tiny
\caption{Entropy, Diversity, and Overlap results for dataset mlearn Cu}
%
\end{table*}

\begin{table*}[h!]
\centering
\tiny
\caption{Entropy, Diversity, and Overlap results for dataset mlearn Ge}
%
\end{table*}

\begin{table*}[h!]
\centering
\tiny
\caption{Entropy, Diversity, and Overlap results for dataset mlearn Li}
%
\end{table*}
\clearpage

\begin{table*}[h!]
\centering
\tiny
\caption{Entropy, Diversity, and Overlap results for dataset mlearn Mo}
%
\end{table*}

\begin{table*}[h!]
\centering
\tiny
\caption{Entropy, Diversity, and Overlap results for dataset mlearn Ni}
%
\end{table*}

\begin{table*}[h!]
\centering
\tiny
\caption{Entropy, Diversity, and Overlap results for dataset mlearn Si}
%
\end{table*}

\begin{table*}[h!]
\centering
\tiny
\caption{Entropy, Diversity, and Overlap results for dataset pure magnesium DFT PRM2020}
\label{tab:si:misc-63}
%
\end{table*}

\end{document}